\documentclass[final]{siamltex}

\usepackage[]{graphicx}
\usepackage{url}
\usepackage{ifpdf}
\usepackage{hyperref}
\usepackage{mathtools}
\hypersetup{breaklinks,colorlinks,citecolor=blue}
\usepackage[noadjust]{cite}
\usepackage[T1]{fontenc}

\usepackage{amsmath}
\usepackage{amssymb}
\usepackage{graphicx}
\usepackage{epstopdf}
\usepackage{color}
\usepackage[linesnumbered, ruled]{algorithm2e}
\usepackage{multirow}

\usepackage{tikz} 
\usepackage{pgfplots} 
\usetikzlibrary{arrows} 
\usepackage{subfigure}
\usepackage{pstool}
\usepackage{soul}

\newcommand{\red}[1]{\textcolor{red}{#1}}

\newcommand{\fot}{{\frac{1}{2}}}

\newtheorem{remark}[theorem]{Remark}

\SetKwRepeat{Do}{do}{while}

\title{Wavelet-Based Segmentation on the Sphere}

\author{Xiaohao Cai$^{\ast}$, Christopher G. R. Wallis$^{\ast}$, Jennifer Y. H. Chan$^{\ast}$, and
Jason D. McEwen\thanks{Mullard Space Science Laboratory (MSSL), University College London (UCL), UK}
  }

\begin{document}

\maketitle

\begin{abstract}
Segmentation, a useful/powerful technique in pattern recognition, is the process of identifying object outlines within images. 
There are a number of efficient algorithms for segmentation 
in Euclidean space that depend on the variational approach and partial differential equation modelling. 
Wavelets have been used successfully in various problems in image processing, 
including segmentation, inpainting, noise removal, super-resolution image restoration, and many others. 
Wavelets on the sphere have been developed to solve such problems for data defined on the sphere, 
which arise in numerous fields such as cosmology and geophysics. 
In this work, we propose a wavelet-based method to segment images on the sphere, accounting for the 
underlying geometry of spherical data. Our method is a direct extension of the tight-frame based segmentation method
used to automatically identify tube-like structures such as blood vessels in medical imaging.
It  is compatible with any arbitrary type of wavelet frame defined on the sphere, 
such as axisymmetric wavelets, directional wavelets, curvelets, and hybrid wavelet constructions.
Such an approach allows the desirable properties of wavelets to be naturally inherited in the segmentation process. 
In particular, directional wavelets and curvelets, which were designed to efficiently capture directional signal content, provide additional  
advantages in segmenting images containing prominent directional and curvilinear features.
We present several numerical experiments, applying our wavelet-based segmentation method, as well as the common K-means method, 
on real-world spherical images, including an Earth topographic map, a light probe image, solar data-sets,
and spherical retina images. These experiments demonstrate the superiority of our method and show that it is 
capable of segmenting different kinds of spherical images, including those with prominent directional features.
Moreover, our algorithm is efficient with convergence usually within a few iterations.

\end{abstract}



\section{Introduction}\label{sec:intro}
Spherical images are common in nature, for example, in cosmology \cite{MVWBCHLMS07},
astrophysics \cite{SSCFG12}, planetary science \cite{A14}, geophysics \cite{SLNDVJVCV11},
and neuro-science \cite{RMSBSW11}, where images are naturally defined on the sphere. 
Clearly, images defined on the sphere are different to Euclidean images in 2D and 3D 
in terms of symmetries, coordinate systems and metrics constructed (see for example \cite{LH10}).
Image segmentation aims to separate a given image into different components, where each part
shares similar characteristics in terms of, e.g., edges, intensities, colours, and textures.
It generally serves as a preliminary step for object recognition and interpretation, and is a 
fundamental yet challenging task in image processing. 
In this paper, we present an effective segmentation method that uses spherical wavelets to segment spherical images.

In the literature, many different approaches have been proposed for image segmentation for 2D, 3D and 
vector-valued images, e.g., 
{\cite{CCMS13,CCNZ15,LF01,MS89,RSK09,SM00,SW14}}.
In particular, in \cite{MS89} the well-known Mumford-Shah model was proposed,
which formulates the image segmentation problem by minimising an energy function and
finding optimal piecewise smooth approximations of the given image. 
More details about these kind of methods can be found in \cite{MS_Handbook_15,C15,CCZ13,CV01}.
These types of methods generally give good segmentation results. 
However, their applicabilities and performance heavily depend on the models used; in some cases 
(e.g.\ segmenting images containing complex textures) the models are difficult or expensive to compute due to the non-convex nature of the problem. 
In \cite{SM00}, a graph-cut based method was proposed to segment point clouds into different groups. 
The more pixels in the image, the larger the size of the eigenvalue problem needs to be solved, which makes 
the method inefficient in terms of speed and accuracy.
Methods based on deformable models \cite{ESF01,LF01} segment via evolving geodesic active contours
that are built from a partial differential equation, with the ability to detect twisted, convoluted and occluded structures,  
but are sensitive to noises and blurs in images. 

Recently, segmentation methods \cite{C15,CCZ13,CS13} designed utilising techniques 
in image restoration (e.g.\ \cite{M04,ROF92}), were proposed for gray-scale images. 
In \cite{C15}, a segmentation model that combines the image segmentation model of \cite{MS89} and the data fidelity terms from 
image restoration models \cite{M04,ROF92} was considered to deal with images contaminated by different types of noise 
(e.g.\ Gaussian, Poisson or impulsive noise). 
In \cite{CCZ13}, the methodology of two-stage methods, solving image restoration 
models first followed by a thresholding second stage, was proposed. These methods were later named SaT (smoothing and thresholding) 
segmentation methods. One advantage of the SaT methods is the fast speed of implementation. 
Akin to the SaT methods, the T-ROF method (thresholding the Rudin-Osher-Fatemi model) in \cite{C15} concluded that the thresholding approach for segmentation was 
equivalent to solving the Chan-Vese segmentation model \cite{CV01}; more detailed theoretical proofs can be found in \cite{CCSSZ18}.
Based on the SaT methodology, a method named SLaT \cite{CCNZ15} was proposed for degraded colour image segmentation. 

In additional to the aforementioned methods, approaches based on wavelets and tight frames \cite{AG03,CCMS11,CCMS13}  have been 
proposed for segmentation. In \cite{CCMS11,CCMS13}, 
a tight-frame based segmentation method was designed 
for a vessel segmentation problem in medical imaging. 
The major advantage of this method is the ability to segment
twisted, convoluted and occluded structures without user interactions. Moreover, the ability of the method to follow the branching of different layers, from thinner to larger structures, makes the method a good 
candidate for a tubular-structured segmentation problem in medical imaging.
However, all the tight-frame systems discussed and used in \cite{CCMS13} (e.g.\ framelets \cite{RS97}, contourlets \cite{DV04},
curvelets \cite{CD05-1,CD05-2}, and dual-tree complex wavelet\footnote{\url{http://taco.poly.edu/WaveletSoftware/}}) 
are designed for 2D or 3D data on a Euclidean manifold. 
Consequently, these approaches cannot be applied to problems where data-sets live 
natively on the sphere. 

Wavelets have become a powerful analysis tool for spherical images,
due to their ability to simultaneously extract both spectral and spatial information. 
A variety of wavelet frameworks have been constructed on the sphere in recent years, e.g.\ \cite{ADJV02,BKMP09,MDW16,MLBVW15,MVW13}, 
and have led to many insightful scientific studies in the fields mentioned above (see \cite{MVWBCHLMS07,SSCFG12,A14,SLNDVJVCV11,RMSBSW11}). 
Different types of wavelets on the sphere have been designed to probe different structure in spherical images, for example isotropic or directional and geometrical features, such as linear or curvilinear structures, to mention a few. Axisymmetric wavelets \cite{NPW06,BKMP09,LMVW13,SMAN06} are useful for probing spherical images with isotropic structure, directional wavelets \cite{MVW13,WMVB08,MDW16,MLBVW15} for probing directional structure, ridgelets \cite{MR10,SMAN06} for analysing antipodal signals on the sphere, and curvelets \cite{SMAN06,CLKM15} for studying highly anisotropic image content such as curve-like features (we refer to \cite{CD05-1,CD05-2} for the general definition of Euclidean ridgelets and curvelets).
Fast algorithms have been developed to compute exact forward and inverse wavelet transforms on the sphere for very large spherical images containing millions of pixels \cite{MVW13,MHML07,MLBVW15} (leveraging novel sampling theorems on
the sphere \cite{MW11} and the rotation group \cite{MBLPW15}).
Localisation properties of wavelet constructions have also been studied in detail \cite{BKMP09,NPW06,MDW16}, showing important quasi-exponential localisation and asymptotic uncorrelation properties for certain wavelet constructions.
An investigation into the use of axisymmetric and directional wavelets for sparse image reconstruction was performed recently in \cite{Wallis2016}, showing excellent performance.
The spherical wavelets adopted in the experimental section of this paper
are reviewed briefly in Section \ref{sec:review}. 

In this paper, we devise an iterative framework for segmenting spherical images
using wavelets defined on the sphere, extending the method proposed in \cite{CCMS11,CCMS13}.  
The first stage of the method, as a preprocessing step, suppresses noises in the given data by soft thresholding wavelet coefficients. 
Then, potential boundary pixels are classified gradually by the iterative procedure.
The framework is compatible with any arbitrary type of spherical wavelet, 
such as the axisymmetric wavelets, directional wavelets, or curvelets mentioned above.  
The iterative strategy in the proposed framework is effective, particularly for
images containing anisotropic textures. There is also flexibility regarding the implementation of iterations.
Motivated by the SaT methodology in \cite{CS13,CCZ13}, when segmenting images containing many (or mostly)
isotropic structures, the iterative strategy in our method can be replaced by a simple thresholding to reduce 
the computation time significantly without sacrificing segmentation quality considerably.
We test the proposed framework on a variety of types of spherical images, including 
an Earth topographic map, a light probe (spherical) image, two sets of solar data, and 
two retina images projected on the sphere.

To the best of our knowledge, this is the first segmentation method that works directly on the {\it whole} sphere and is practical for any type of spherical images, benefiting from the compatibility of the method with 
any type of spherical wavelets. 
A method \cite{RWB08} was proposed for segmenting spherical particles in volumetric data sets based on an extension of the generalised Hough transform and an active contour approach. However, 
the data considered in \cite{RWB08} were 3D data containing spherical-like particles, not data defined on the sphere directly.
{Another method \cite{NHBT07} was proposed for 3D shape segmentation based on an active contour formulation with a shape prior which was formed by using spherical wavelets. However, the data considered in \cite{NHBT07} were 3D data containing 3D shapes, not data that are defined natively on the sphere. }

{Very recently, spherical deep neural networks, an important tool in artificial intelligence, have emerged, e.g., SphereNet \cite{CCG18}, 
Spherical CNNs \cite{CGKW18, EBMD18}, 
Spherical CNNs on Unstructured Grids \cite{JHKPMN19}, 
and DeepSphere \cite{PDKS19}. 
The success of deep learning generally depends on the exploitation of data properties by the network architectures for efficient and principled learning, as well as on the quality of the training sets. 
Our proposed method tackles the task of segmentation with a completely different approach and does not require any training sets. On the other hand, an extension of our method with deep neural networks is possible (see a brief discussion in Section \ref{sec:alg}). 
}

The main contributions in this paper are: (1) a segmentation framework for spherical images is devised, for the first time;
(2) the framework uses an iterative strategy with the flexibility to tailor the iterative procedure according to data types and features; 
(3) spherical wavelets, including axisymmetric wavelets, directional wavelets and the newly-constructed 
hybrid directional-curvelet wavelets, are implemented and tested in the framework; (4) a series of applications are presented, illustrating the 
performance of our proposed segmentation method.

The remainder of this paper is organised as follows. In Section \ref{sec:review}, 
we review related work about spherical wavelets and segmentation methods, 
and present our new hybrid directional-curvelet wavelet construction.
In Section \ref{sec:alg}, we introduce our spherical segmentation method.
In Section \ref{sec:results}, the proposed method and methods for comparison are tested on a variety of spherical
images such as an Earth map, light probe images, and two solar maps. To further demonstrate
the ability of our method on segmenting highly directional and elongated structures, in Section \ref{sec:results} we also 
apply it to retinal images, which contain a complex network of blood vessels.
Conclusions are given in Section \ref{sec:con}.

\section{Background}\label{sec:review}

Let $f\in \textrm{L}^2(\mathbb{S}^2)$ be the given image defined on the sphere $\mathbb{S}^2$.
Without loss of generality, we assume $f$ in [0, 1]. Let  $\omega = (\theta, \phi) \in \mathbb{S}^2$ denote spherical
coordinates with colatitude $\theta \in [0,\pi]$ and longitude $\phi \in [0,2\pi)$. Let $\bar{\mathbb{S}}^2$
be the discretised sphere of $\mathbb{S}^2$.  We review sampling, wavelets and discrete gradient operators on the sphere subsequently, before recalling the tight-frame based segmentation method of \cite{CCMS11,CCMS13}.  In addition, we present a new hybrid directional-curvelet wavelet construction.

\subsection{Sampling on the sphere}\label{sec:sampling}

We adopt the equiangular sampling theorem on the sphere of \cite{MW11}, which defines how to capture the information content in a signal band-limited at $L$ in $\sim 2 L^2$ samples.  This sampling theorem requires the fewest number of samples to capture all information content on band-limited spherical images.  In additional, fast algorithms to perform the associated spherical harmonic transform are presented \cite{MW11}.
Typically, we consider band-limited spherical images whose spherical harmonic coefficients $f_{\ell m} = 0, \forall \ell \ge L$,
where $f_{\ell m} = \langle f, Y_{\ell m} \rangle$ and $Y_{\ell m} \in \textrm{L}^2(\mathbb{S}^2)$ are the spherical harmonics, with $\ell \in \mathbb{N}$ and $m \in \mathbb{Z}$ satisfies $|m| \le \ell$. In practice many real-world signals can be approximated accurately by a band-limited signal).
The equiangular sample positions of the sphere associated with this sampling theorem are given by
\[
\theta_t = \frac{\pi(2t+1)}{2L-1}, \quad \phi_p = \frac{2\pi p}{2L-1}
\]
where  $t\in \{0, 1, \ldots, L-1\}$ and $p\in \{0, 1, \ldots, 2L-2\}$
index the equiangular samples in $\theta$ and $\phi$, respectively. 
For example, when $L=512$, the sphere is discretised with $512\times 1023=523776$ samples.
Please refer to \cite{MW11} and references therein for more information about sampling on the sphere.

\subsection{Wavelets on the sphere}\label{sec:wavelet}
In many real-life problems, data to be processed are usually in a discretised form, as described previously. Also, exact reconstruction of the signal is commonly desired. 
Scale-discretised wavelets \cite{CLKM15,LMVW13,MLBVW15,MDW16,MVW13,WMVB08} on the sphere allow the exact synthesis of discrete 
spherical images from their wavelet coefficients.  We adopt scale-discretised wavelet constructions in this work, which we review concisely in this section.  In addition, we present a new hybrid directional-curvelet scale-discretised wavelet construction.

{\it Wavelet transforms.}
 Let $\Psi^{(j)}\in \textrm{L}^2(\mathbb{S}^2)$ be the wavelet with wavelet scales $j\in \mathbb{N}$ and $0 \le J_{\rm min} \le j \le J_{\rm max}$,
 which encode the angular localisation of $\Psi^{(j)}$, where $J_{\rm min}$ and $J_{\rm max}$ are the
 minimum and maximum wavelet scales considered, respectively; see \cite{MLBVW15} for more details about $j$. 
 For directional wavelet transforms, wavelet coefficients are defined on the rotation group SO(3), parameterised by
 Euler angles $\rho = (\alpha, \beta, \gamma) \in \textrm{SO(3)}$ with $\alpha \in [0,2\pi)$, $\beta \in [0,\pi]$ and
 $\gamma \in [0,2\pi)$. Wavelet coefficients $W^{\Psi^{(j)}} \in \textrm{L}^2\textrm{(SO(3))}$ are computed 
 by the wavelet forward transform (analysis) defined by
\begin{equation}
  \label{eqn:analysis}
W^{\Psi^{(j)}}(\rho) \equiv (f\circledast \Psi^{(j)})(\rho) \equiv \langle f, {\cal R}_{\rho} \Psi^{(j)}  \rangle 
	= \int_{\mathbb{S}^2} \textrm{d}\Omega(\omega) f(\omega) ({\cal R}_{\rho} \Psi^{(j)})^{\ast}(\omega),
\end{equation}
where ${\cal R}_{\rho} $ is a rotation operator related to a 3D rotation matrix $R_{\rho}$ by   
$({\cal R}_{\rho} \Psi^{(j)}) (\omega) \equiv \Psi^{(j)} (R_{\rho}^{\textrm{-1}} \hat{\omega})$ 
($\hat{\omega}$ is the Cartesian vector of $\omega$), 
$\textrm{d}\Omega(\omega) = \sin\theta \textrm{d} \theta \textrm{d} \phi$ is the usual rotation invariant
measure on the sphere; the symbol $\langle \cdot, \cdot \rangle$, the operator $\circledast$, and $\cdot^*$ denote 
the inner product of functions, directional convolution on the sphere and complex conjugation, respectively.
Low-frequency content of the signal not probed by wavelets are probed by the scaling function $\Phi\in \textrm{L}^2(\mathbb{S}^2)$, which is generally axisymmetric.  
The scaling coefficients $W^{\Phi} \in \textrm{L}^2(\mathbb{S}^2)$ are given by
\begin{equation}
  \label{eqn:analysis_scaling}
W^{\Phi}(\omega) \equiv (f\odot \Phi)(\omega) \equiv \langle f, {\cal R}_{\omega} \Phi  \rangle 
	= \int_{\mathbb{S}^2} \textrm{d}\Omega(\omega') f(\omega') ({\cal R}_{\omega} \Phi)^{\ast}(\omega'), 
\end{equation}
where ${\cal R}_{\omega} = {\cal R}_{(\phi, \theta, 0)}$, and the operator $\odot$ denotes axisymmetric convolution on the sphere.

The spherical image $f$  can be synthesised perfectly from its wavelet 
and scaling coefficients (under the wavelet admissibility condition \cite{MLBVW15}) by the wavelet backward transform (synthesis) by 
\begin{equation}  \label{eqn:synthesis}
f(\omega) = \int_{\mathbb{S}^2} \textrm{d}\Omega(\omega') W^{\Phi}(\omega') ({\cal R}_{\omega'} \Phi)(\omega)  
		+ \sum_{j=J_{\rm min}}^{J_{\rm max}} \int_{\textrm{SO(3)}} \textrm{d}\varrho(\rho) W^{\Psi^{(j)}}(\rho) ({\cal R}_{\rho} \Psi^{(j)})(\omega),
\end{equation}
where $\textrm{d}\varrho(\rho) = \sin\beta \textrm{d} \alpha \textrm{d} \beta \textrm{d} \gamma$ 
is the usual invariant measure on SO(3).

{\it Construction of different types of wavelets.}
Spherical wavelets, constructed to ensure the admissibility condition is satisfied, are defined in harmonic space in the factorised form by
\begin{equation}\label{eqn:wav_con}
\Psi^{(j)}_{\ell m} \equiv \sqrt{\frac{2\ell+1}{8\pi^2}} \kappa^{(j)}(\ell) \zeta_{\ell m},
\end{equation}
where kernel $\kappa^{(j)} \in \textrm{L}^2(\mathbb{R}^+)$, a positive real function, is constructed to be a smooth function 
with compact support to control the angular localisation properties of wavelet $\Psi^{(j)}$, 
with harmonic coefficients $\Psi^{(j)}_{\ell m} = \langle \Psi^{(j)},  Y_{\ell m} \rangle$;
see \cite{MDW16} for the detailed definition. The directionality component $\zeta\in \textrm{L}^2(\mathbb{S}^2)$,
with harmonic coefficients $\zeta_{lm} = \langle \zeta, Y_{\ell m} \rangle$, is designed to control the directional localisation properties of $\Psi^{(j)}$. 
The wavelets recovered are steerable when imposing an azimuthal band-limit $N$ on the directionality component such that 
$\zeta_{\ell m} = 0$ for $|m| \ge N, \forall \ell, m$. While steerability is achieved, the directional localisation of the wavelet is
controlled by imposing a specific form for the directional auto-correlation of the wavelet.
The detailed construction of $\zeta$ and $\zeta_{\ell m}$ for directional wavelets can be found in \cite{MDW16} and those 
for curvelets can be found in \cite{CLKM15}.
In particular, the spherical curvelets proposed in \cite{CLKM15} exhibits the parabolic scaling
relation. Such a geometric feature is unique to curvelets,
making it highly anisotropic and directionally sensitive, and  thus suitable for extracting local curvilinear structures effectively.
Moreover, scale-discretised wavelets support the exact analysis and synthesis of both scalar and spin signals, although only the former are considered herein.

\begin{figure}
\centering
	\begin{tabular}{ccc}
  		\includegraphics[trim={0 0 0 0}, clip, width=0.3\linewidth]{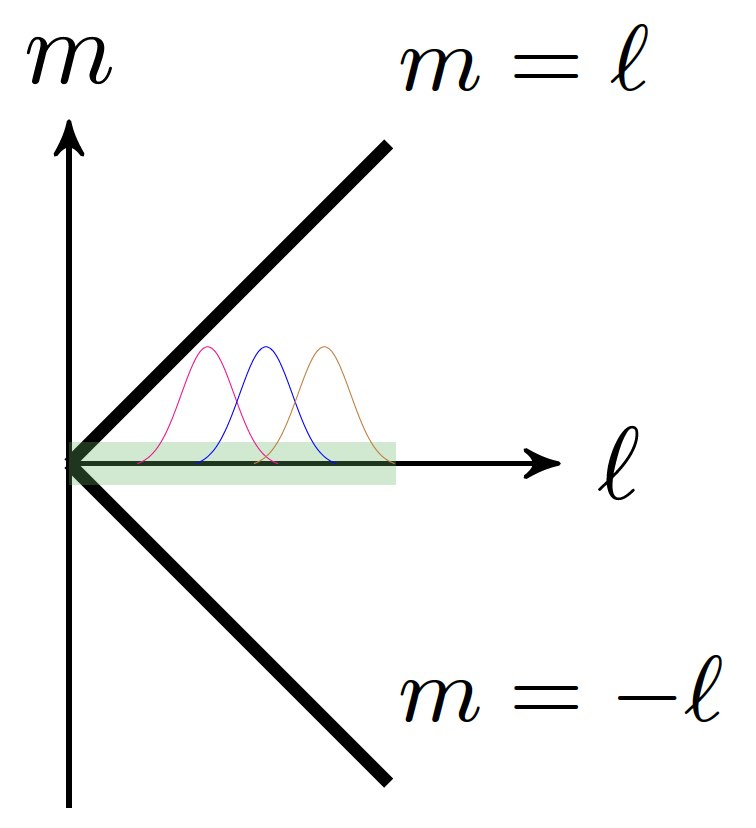} &
		\includegraphics[trim={0px 0 0px 0}, clip, width=0.3\linewidth]{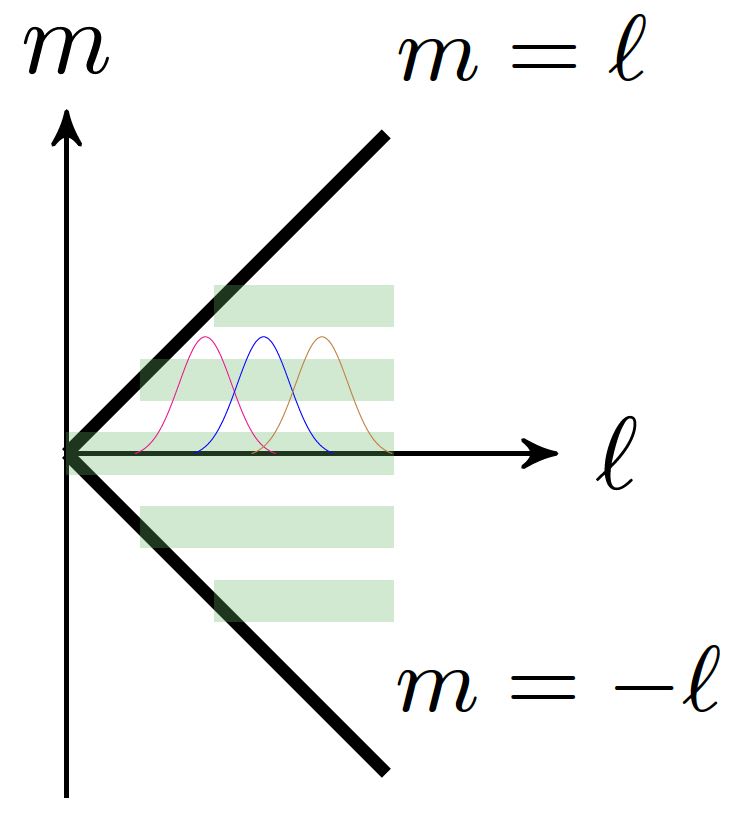} &
	        	\includegraphics[trim={0px 0 0px 0}, clip, width=0.3\linewidth]{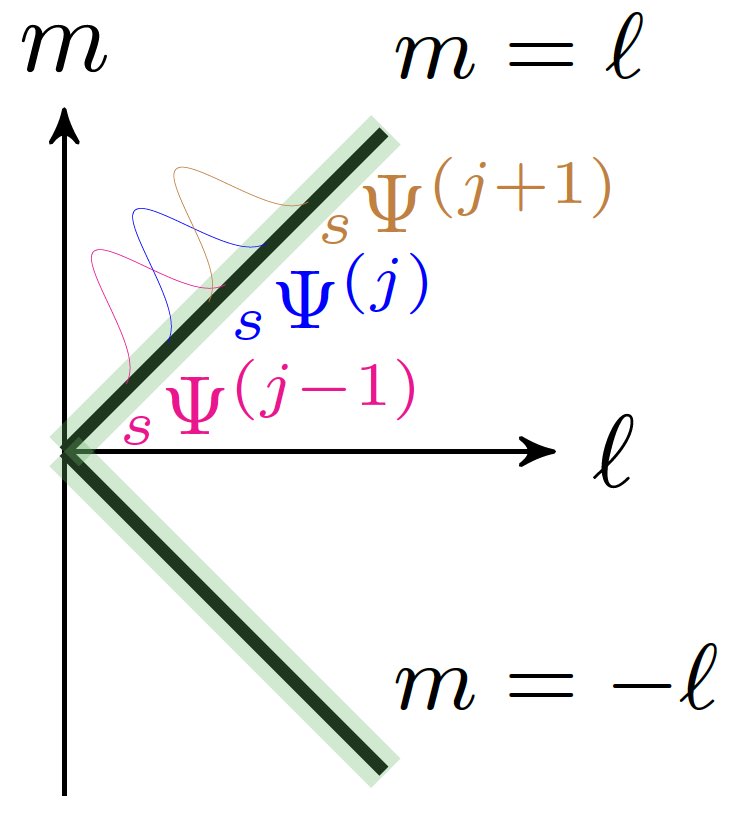}
	\end{tabular}
\caption[]{Harmonic tilings of different types of wavelets, including axisymmetric wavelets, directional wavelets, and curvelets, 
	respectively, from left to right (refer to \cite{CLKM15}).}
\label{fig:ml_tiling}
\end{figure}

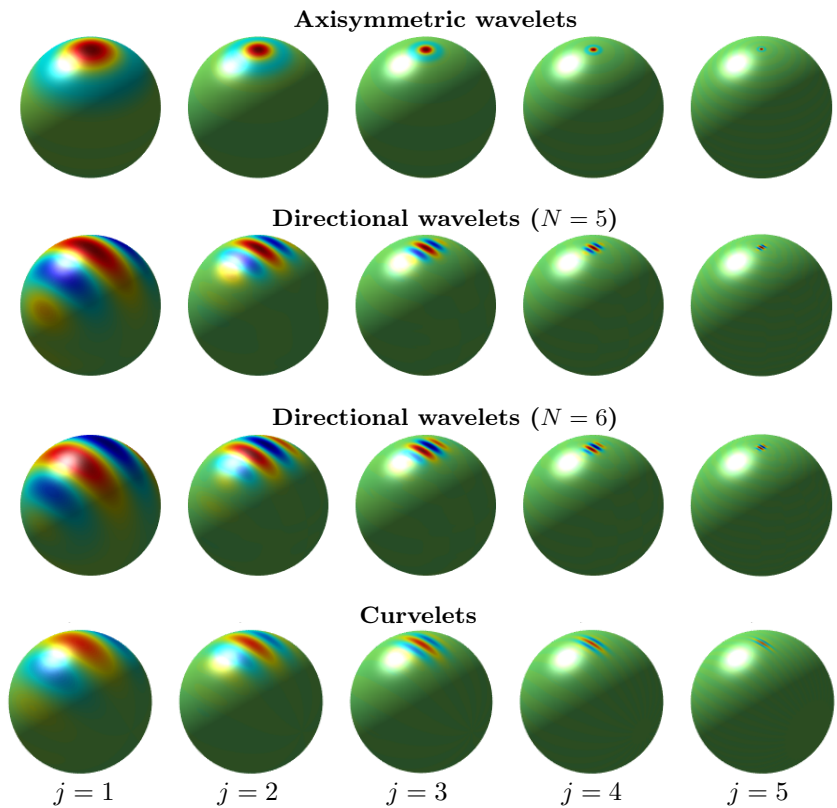
\begin{figure}
\centering
	\begin{tabular}{c}
  		\includegraphics[trim={{.33\linewidth} {.1\linewidth} {.1\linewidth} {.13\linewidth}}, clip, width=0.9\linewidth]
		{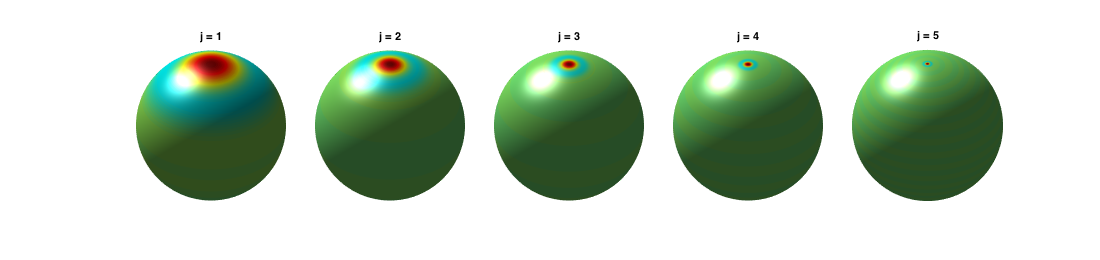}   \put(-225,65){\small \bf Axisymmetric wavelets} 
		\\
		\includegraphics[trim={{.33\linewidth} {.1\linewidth} {.1\linewidth} {.13\linewidth}}, clip, width=0.9\linewidth]
		{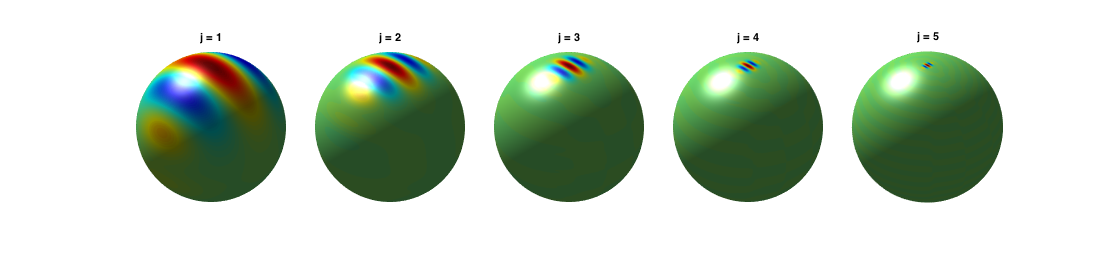} \put(-233,65){\small \bf Directional wavelets ($N=5$)} 
		\\
	        	\includegraphics[trim={{.33\linewidth} {.1\linewidth} {.1\linewidth} {.13\linewidth}}, clip, width=0.9\linewidth]
		{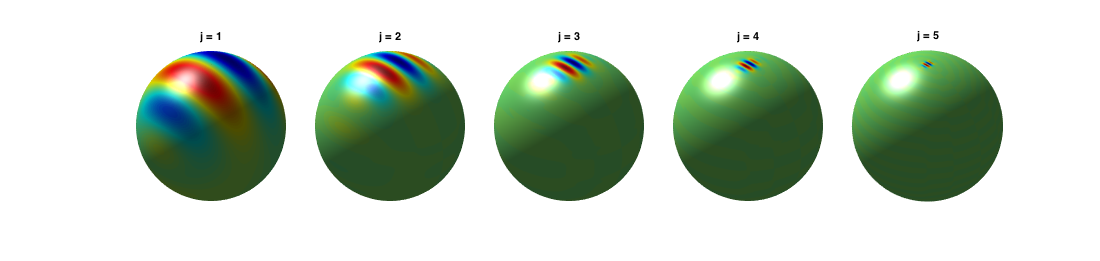}  \put(-233,65){\small \bf Directional wavelets ($N=6$)} 
		\\
		\includegraphics[trim={{.31\linewidth} {.1\linewidth} {.1\linewidth} {.138\linewidth}}, clip, width=0.9\linewidth]
		{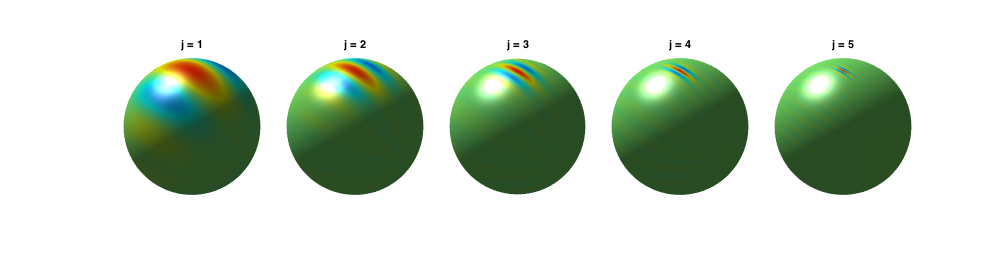}  \put(-200,69){\small \bf Curvelets}
		\put(-317,2){ $j=1$}  \put(-255,2){ $j=2$}  \put(-191,2){ $j=3$} \put(-125,2){ $j=4$} \put(-62,2){ $j=5$} 
	\end{tabular}
\caption[]{Scalar scale-discretised axisymmetric wavelets, directional wavelets ($N = 5$ and 6), and curvelets on the sphere for $L = 512$,
	from the first row to the fourth row.}
\label{fig:ml_show}
\end{figure}

Fig.\ \ref{fig:ml_tiling} and Fig.\ \ref{fig:ml_show} show the harmonic tilings of different types of scale-discretised 
wavelets and the corresponding wavelets plotted on the sphere, respectively.
We refer the reader to \cite{LMVW13,MDW16,MVW13,MLBVW15,WMVB08}, and \cite{CLKM15}
for details about the construction of scale-discretised axisymmetric and directional wavelets, and curvelets, respectively.
Code to compute these wavelet transforms is public and available in the existing {\tt S2LET}\footnote{\url{http://www.s2let.org}}  package, which relies on the {\tt SSHT}\footnote{\url{http://www.spinsht.org}} \cite{MW11} and {\tt SO3}\footnote{\url{http://www.sothree.org}} \cite{MBLPW15} packages.

{\it Hybrid wavelets.} Different wavelet transforms have differing computational requirements.
Generally, computing axisymmetric wavelet transforms
are fastest with computational time scaling as $\mathcal{O}(L^3)$ \cite{LMVW13}, 
directional wavelet transforms are slower with computational time scaling as $\mathcal{O}(N\,L^3)$ \cite{MLBVW15}, 
while curvelet transforms are the slowest with computational time scaling as $\mathcal{O}(L^3 \log_{2}{L})$ \cite{CLKM15}.
As an example, a complete round-trip of a forward and backward wavelet transform, with band-limit $L = 512$, takes a few seconds on 
a Macbook with i5 processor for axisymmetric wavelets, several minutes for directional wavelets and 
several hours for curvelets (see Tables \ref{tab:earth}, \ref{tab:lightprobe}, \ref{tab:solar} and \ref{tab:retina}). 
However, the increase in computational cost is offset by an improved ability to represent directional and curvilinear 
structure for directional wavelets and curvelets, respectively.

With the aim to exploit the advantages of the curvelet transform \cite{CLKM15} while shortening the computational time needed, here we construct 
a hybrid form of wavelet transform on the sphere using both curvelets and directional wavelets. 
The idea (proposed as a future work in our paper \cite{CLKM15}) is to
describe small-scale features with directional wavelets and remaining features with curvelets,
thereby inheriting the excellent directional localisation of curvelets and computational advantages of directional wavelets.

The separation between the two wavelet types is performed in harmonic space, at a defined transition band-limit $L_{\rm trans}$.
 The curvelet transform is performed up to the band-limit $L_{\rm trans}$, ignoring the final wavelet scale. This provides  
 the large-scale curvelet coefficients. In order to calculate the component of the image represented by curvelets,  
 the inverse transform is performed, yielding $f^{\rm curv}$.
 The directional wavelet coefficients are found by first subtracting this image from the original, 
 $f^{\rm dir} = f - f^{\rm curv}$, before performing the directional wavelet transform on the difference image $f^{\rm dir}$. 
The balancing between curvelets and directional wavelets, $L_{\rm trans}$
is also flexible and can be tuned in our hybrid construction, depending on the importance of directional structure in the 
image or the computational time available. 

This hybrid wavelet transform is implemented in the {\tt S2LET} package.\footnote{Support for hybrid wavelets in the {\tt S2LET} package will be made public following the publication of this article.}
The current implementation is not optimised
as it performs a full backward wavelet transform when only one scale needs to be transformed.
This optimisation is left for future work.

\subsection{Gradient operators on the sphere}\label{sec:gradient}
The segmentation method developed herein requires the computation of gradients on the sphere 
$\nabla f = (\frac{\partial f}{\partial \theta}, \frac{\partial f}{\partial \phi})$, with the continuous magnitude of the gradient given by
\begin{equation*}
\| \nabla f \| \equiv
\sqrt{
\Biggl ( \frac{\partial f}{\partial \theta} \Biggr)^2
+
\frac{1}{\sin^2\theta}\Biggl (  \frac{\partial f}{\partial \phi} \Biggr )^2
}.
\end{equation*}
Discrete gradient operators for the equiangular sampling scheme adopted \cite{MW11} are defined in \cite{MPTVVW13}.  
The discrete magnitude of the gradient is simply defined by
\begin{equation} \label{eqn-grad}
\| \nabla f \|
\equiv
\sqrt{
\bigl ( \delta_{\theta} f \bigr )^2
+
\frac{1}{\sin^2{\theta_t}}\bigl (  \delta_{\phi}  f \bigr )^2
}.
\end{equation}
where $\delta_{\theta}$ and $\delta_{\phi}$ are finite difference operators.  For more details of the discrete gradient operator please refer to \cite{MPTVVW13}.

\subsection{Tight-frame based segmentation method}\label{sec:tf-med}
The following presents the generic tight-frame  algorithm used in e.g.\ \cite{JRZ08}: 
\begin{eqnarray}
{f}^{(i+\fot)} &=& \mathcal{U}({f}^{(i)}), \label{tf-q1}  \\
{f}^{(i+1)} &=& {\cal A}^{\textrm T}\mathcal{T}_\lambda({\cal A} {f}^{(i+\fot)}),\quad
i=1, 2,\ldots. \label{tf-q2}
\end{eqnarray}
Here ${\cal A}$ and ${\cal A}^{\textrm T}$ are the tight-frame (wavelets in our case) forward and backward transforms respectively,
$f^{(i)}$ is an approximate solution at the $i$-th iteration, $\mathcal{U}$ is a problem-dependent operator (e.g.\ $\mathcal{U}$ is the 
identity operator for a denoising problem), and
$\mathcal{T}_\lambda(\cdot)$ is the soft-thresholding operator defined by 
\[
\mathcal{T}_\lambda(\vec{v}) \equiv
[t_{\lambda}(v_1), \cdots, t_{\lambda}(v_n)]^{\textrm T},
\] 
where $\vec{v}=[v_1, \cdots, v_n]^T \in \mathbb{R}^n$ and $ \lambda \in \mathbb{R}^+$
are a given vector and constant respectively, and
\begin{equation}\label{ltk}
t_{\lambda}(v_k) \equiv
\left\{
\begin{array}{lcl}
\textrm{sign}(v_k)(|v_k|-\lambda), & &  \text{if}\  |v_k| > \lambda, \\
0, \qquad \qquad \qquad  & &  \text{if}\  |v_k| \leq \lambda.
\end{array}
\right.
\end{equation}

To obtain a binary result, where values 1 and 0 represent the object of interest and the background respectively, 
an iterative procedure was proposed in \cite{CCMS11,CCMS13} to
gradually update an interval that contains pixel values of potential boundary pixels until the interval is empty.
Note that the test image discussed in \cite{CCMS11,CCMS13} is assumed to have a low noise level and scaled to the range [0, 1]. 

The main segmentation procedures of \cite{CCMS13} are as follows. Firstly, separate the given image to three parts by thresholding, 
i.e., area of background, area of object of interest, and the uncertainty-area which needs to be labelled as background or
object in future steps. Secondly, denoise and smooth the uncertainty-area by the tight-frame
algorithm \cite{JRZ08} to get a new uncertainty-area which is smaller than the previous one.
Thirdly, stop the algorithm when the uncertainty-area is empty (a binary result is then obtained), otherwise continue.

\section{Spherical segmentation method}\label{sec:alg}
In \cite{CCMS11,CCMS13}, the tight-frame based segmentation method is applied to Euclidean images but it is in principle 
extendable to a spherical domain and is compatible with different types of wavelets transforms.  
In this paper, based on the idea in the method \cite{CCMS11,CCMS13} on Euclidean images, we propose the wavelet-based segmentation 
framework on the sphere for spherical images. 

The idea behind the method is to detect the candidates of possible pixels on (near) the boundary first, then
gradually purify these boundary-like pixels via an iterative procedure until all pixels on the sphere are classified as
inside or outside of a boundary. With the aid of the fact that possible pixels on the boundary have particular 
properties in terms of pixel values and gradients, boundary-like pixels are detected and represented by
a range $[a_0, b_0]$. Then, an iterative strategy shrinking this range is applied, so to keep removing 
pixels from it until the range itself is empty. All pixels are eventually classified either as in the foreground
(the objects of interest) or in the background. Note that pixels in the foreground and in the background are 
represented respectively by values 1 and 0. When a binary result is obtained the algorithm stops. 

The greater the anisotropic structure in the image, the more complicated the boundary-like pixels in $[a_0, b_0]$. 
Therefore, using an iterative procedure is particularly useful for images containing anisotropic structures.
Otherwise, replacing the iterative procedure by thresholding is more economical (as demonstrated very effective 
in \cite{CCNZ15,CCSSZ18,CCZ13,CS13}). In the following, we discuss each of the iterative steps of the
proposed method in more detail. 

{\it Preprocessing.} If $f$ is contaminated with significant noise, a preprocessing step to suppress the noise is necessary.
We use one iteration step of the tight-frame algorithm \eqref{tf-q2} to deal with the noise by soft thresholding, i.e.
\begin{equation} \label{f_bar}
\bar{f} = {\cal A}^\top\mathcal{T}_{\bar{\lambda}}({\cal A} {f}).
\end{equation}
Note that ${\cal A}$ here is a wavelet transform on sphere.

{\it Initialisation.} Let $\Lambda^{(0)}$ be the initial set of potential boundary pixels, which is identified by
using the gradient of $\bar{f}$, i.e.\ pixels with gradient larger than a given threshold $\epsilon$ are 
in $\Lambda^{(0)}$, therefore
\begin{equation} \label{lambda0}
\Lambda^{(0)} \equiv \{k \in \bar{\mathbb{S}}^2  \ | \ \| [\nabla \bar{f}]_k\|_1 > \epsilon\}.
\end{equation}
Here $[\nabla \bar{f}]_k$ (cf. \eqref{eqn-grad}) is the discrete gradient of $\bar{f}$ at the $k$-th pixel on the sphere.
Set $f^{(0)} = \bar{f}$, with $\Lambda^{(0)}$ defined in \eqref{lambda0}. We start the iterative process from $i=0$.
The $i$-th iteration is described in detail below.

{\it Step 1: computing the range $[a_{i}$, $b_{i}]$.}
Given $\Lambda^{(i)}$, define $[a_{i}$, $b_{i}]$ by 
\begin{equation}
a_i \equiv \max\left\{\frac{\mu^{(i)} + \mu^{(i)}_{-}}{2},0\right\}, \quad
b_i \equiv  \min\left\{\frac{\mu^{(i)}+ \mu^{(i)}_{+}}{2},1\right\},
\label{int_i}
\end{equation}
where 
\begin{equation} \label{muplus}
\mu^{(i)} = \frac{1}{|\Lambda^{(i)}|} \sum_{k\in \Lambda^{(i)}} f^{(i)}_k
\end{equation}
is the mean pixel value on $\Lambda^{(i)}$, $|\cdot |$ denotes the cardinality of the set, $f^{(i)}_k$ is the pixel value of pixel $k$ in spherical image $f^{(i)}$,
and $\mu^{(i)}_{-}$ and $\mu^{(i)}_{+}$ are defined by
\begin{equation} \label{muminus}
\mu^{(i)}_{-} = \frac{\sum_{\{ k\in \Lambda^{(i)}: f ^{(i)}_k \leq \mu^{(i)}\}}f ^{(i)}_k}{|\{ k\in \Lambda^{(i)}: f ^{(i)}_k \leq \mu^{(i)}\}|},
\quad 
\mu^{(i)}_{+} =
\frac{\sum_{\{ k\in \Lambda^{(i)}: f ^{(i)}_k \geq \mu^{(i)}\}}f ^{(i)}_k}{|\{ k\in \Lambda^{(i)}: f ^{(i)}_k \geq \mu^{(i)}\}|}.
\end{equation}
Note that $\mu^{(i)}_{-}$ and $\mu^{(i)}_{+}$, the mean pixel values of the two sets separated by $\mu^{(i)}$, reflect
the mean energies of the pixels on the boundary closer to the background and closer to the foreground respectively. 
The definition of $[a_{i}$, $b_{i}]$ in \eqref{int_i} is approximately half length of $[a_{i-1}$, $b_{i-1}]$,  
ensuring the shrinkage property of these ranges.

{\it Step 2: thresholding the image into three parts.}
Using $[a_i, b_i] \subseteq [0, 1]$, we separate image $f^{(i)}$ into three parts ---
those below (set those pixel values that are smaller than $a_i$ to 0), 
inside (stretch those pixel values between 0 and 1 using a simple linear contrast stretch), 
and above (set those pixel values that are larger than $b_i$ to 1) the range, i.e.,
\begin{equation}
f^{(i+\fot)}_k =
\left\{
\begin{array}{lclll}
0, & & {\rm if} \  f^{(i)}_k  \le a_i, \\
\frac{f^{(i)}_k - m_i}{M_i-m_i}, & &
 a_i \le f^{(i)}_k \le b_i, & & \quad {\rm for\ all} \ k \in {\bar{\mathbb{S}}^2}.
\label{eq5} \\
1, & & {\rm if} \  b_i \le  f^{(i)}_k,
\end{array}
\right.
\end{equation}
where
\begin{eqnarray}
\begin{split}
M_i &= {\rm max}\{ f^{(i)}_k \ | \      a_i \le f^{(i)}_k \le b_i, k \in {\Lambda^{(i)}} \}, \\
m_i &={\rm min}\{ f^{(i)}_k \ | \      a_i \le f^{(i)}_k \le b_i, k \in {\Lambda^{(i)}} \}. \label{Mm}
\end{split}
\end{eqnarray}
The set of the remaining pixels that wait to be labelled is represented by
\begin{equation}\label{Lambdai}
\Lambda^{(i+1)}=\{ k \ | \ 0 < f^{(i+\fot )}_k < 1 , k \in {{\bar{\mathbb{S}}^2}} \}.
\end{equation}
Note that if $\Lambda^{(i+1)} = \emptyset$, the threshold image $f^{(i+\fot)}_k $ is binary and the algorithm stops.

\begin{remark}
After obtaining $\Lambda^{(i+1)}$ from formula \eqref{Lambdai}, the segmentation accuracy could be 
improved by making a correction of $\Lambda^{(i+1)}$, such as by adding the labelled but isolated 
(or wrongly labelled) pixels back to $\Lambda^{(i+1)}$ before moving to step \eqref{tfi} (or step \eqref{eqn-th}).
We leave this to future work.
\end{remark}

When the pixels in range $[a_i, b_i]$ can be classified easily (e.g.\ the number of pixels left to be classified is few
therefore they are no longer critical to the final segmentation result) as 
the background or the objects of interest, the thresholding step for segmentation can be invoked:
\begin{equation} \label{eqn-th}
f^{(i+\frac{3}{2})}_k =
\begin{cases}
0, & {\rm if} \  f^{(i+\fot)}_k  < \mu, \\
1, & {\rm if} \  f^{(i+\fot)}_k \ge  \mu,
\end{cases}
\quad 
\mu= \frac{1}{|\Lambda^{(i+1)}|} \sum_{k\in \Lambda^{(i+1)}} f^{(i)}_k
\end{equation}
and the iteration terminates.

{\it Step 3: spherical wavelets iteration.}
Let ${\cal I}$ be the identity operator and ${\cal P}^{(i+1)}$ be the operator where the entry 
is 1 if the corresponding index is in $\Lambda^{(i+1)}$, and 0 otherwise. Then
\begin{equation} \label{tfi}
f^{(i+1)} \equiv ({\cal I}-{\cal P}^{(i+1)})f^{(i+\fot)} + {\cal P}^{(i+1)}
{\cal A}^{\textrm{T}}\mathcal{T}_{\lambda}({\cal A}f^{(i+\fot)}).
\end{equation}
Recall ${\cal A}$ here represents a wavelet transform on sphere 
(e.g.\ axisymmetric wavelets, directional wavelets, curvelets, or hybrid wavelets).
Note that the values of all pixels outside $\Lambda^{(i+1)}$
are either $0$ or $1$, hence the cost of \eqref{tfi} can be reduced significantly
by applying the forward and backward wavelet transforms on pixels around $\Lambda^{(i+1)}$ only.
This optimisation is left for future work.
 
 {\it Stopping criterion.}
As soon as all the pixels of $f^{(i+\fot)}$ are either of value $0$ or $1$, or equivalently when
$\Lambda^{(i)}=\emptyset$, the iteration is terminated, then all the pixels with value 1 constitute the objects
of interest otherwise they are considered as background. 

Algorithm \ref{alg:seg} below summarises the steps required to segment a spherical image $f$ by our segmentation method.
Its convergence proof follows the proof given in \cite{CCMS13}. In subsequent sections, algorithm \ref{alg:seg} is referred to as WSSA for simplicity.
 
\begin{algorithm} 
\caption{Wavelet-based Spherical Segmentation Algorithm (WSSA)}
\label{alg:seg}
 \textbf{Input:} given image $f \in \textrm{L}^2(\mathbb{S}^2)$ \\
  Preprocessing by \eqref{f_bar} \\
 Set $f^{(0)}= \bar{f}$  and $\Lambda^{(0)}$ by \eqref{lambda0}\\
\Do{Stopping criterion is not reached}{
   	compute $[a_i, b_i]$ by \eqref{int_i} \\
   	compute $f^{(i+\fot)}$ by \eqref{eq5} \\
   	stop if $f^{(i+\fot)}$ is a binary image \\
   	compute $\Lambda^{(i+1)}$ by \eqref{Lambdai} \\
   	compute $f^{(i+1)}$ by \eqref{tfi} (or compute $f^{(i+\frac{3}{2})}$ by \eqref{eqn-th} then stop) \\
        $i=i+1$
}
\end{algorithm}

{A potential extension of our proposed spherical segmentation method is to interface with spherical neural networks, which have recently shown to be highly effective for analysing spherical images \cite{CCG18,CGKW18,EBMD18,PDKS19,JHKPMN19}. 
The first vital steps of it involve using a spherical neural network to effectively denoise and smooth the data on the sphere, i.e. to replace the procedures shown in lines 2 and 9 of Algorithm \ref{alg:seg}. We leave the detailed implementation as our future work. }

\section{Experiments}\label{sec:results}
We apply our method, namely WSSA, to various kinds of real-life images, including an Earth topographic map, light probe image, solar data-sets,
and retina images projected on the sphere. Axisymmetric wavelets, directional wavelets, and hybrid wavelets constructed by combining the 
directional wavelets and curvelets are tested and their performances are compared. 
Algorithm \ref{alg:seg} (WSSA) equipped with axisymmetric wavelets, directional wavelets and hybrid wavelets are referred to as 
WSSA-A, WSSA-D, and WSSA-H, respectively.
The code to perform these spherical wavelets transforms used are available in the software package {\tt S2LET} \cite{LMVW13,MLBVW15}. 
The popular K-means method (e.g.\ \cite{HW79,KMNPSW02}) is implemented for comparison purposes.
Here, the K-means method is applied to data on the sphere according to the pixels intensities, using the {\sc Matlab} built-in function {\tt kmeans}.
All the experiments are executed on a MacBook with 2.2 GHz Intel Core i7 processor and 16GB RAM.

{\it Parameters}.
We set the spherical wavelet band-limit $L = 512$, the minimum angular scale $J_{\rm min} =2$, 
and the number of directions probed by directional wavelets to be $N=5$ and 6.
We discretise the sphere $\mathbb{S}^2$ with size $512\times 1023$ (refer to section \ref{sec:sampling}), 
therefore $|\bar{\mathbb{S}}^2| = 523776$. For the hybrid wavelets $L_{\rm trans}\in \{32, 64\}$, 
which means that curvelets are used for bands up to $\ell \lesssim \{32, 64\}$ and directional wavelets for the remaining bands. 
Gaussian noise with standard deviation $\sigma = \|f\|_{\infty} 10^{-\textrm{SNR}/20}$ and 0 mean is added to the test data,
where $\textrm{SNR} = 30$ dB and $\|\cdot \|_{\infty}$ is the infinity norm (referring to the maximum value). 
We fix the thresholding parameter $\bar{\lambda} = \sigma/4$ in \eqref{f_bar} for denoising, 
and $\lambda = \sigma/100$ in \eqref{tfi} for segmentation during the spherical wavelet iterations.

\subsection{Earth topographic map, light probe image and solar data-sets}
Our first example is segmenting an Earth topographic map. The original Earth topography data are taken from the 
Earth Gravitational Model (EGM2008) publicly released by the U.S.\ National Geospatial-Intelligence Agency 
(NGA) EGM Development Team.\footnote{These data were downloaded and extracted using the tools available 
from Frederik Simons' webpage: \url{http://www.frederik.net}.} The signal is band limited to $L=512$ by
performing a forward spherical harmonic transform, band limiting in harmonic space and transforming the signal back from its coefficients. 

\begin{figure}
	\centering
	\begin{tabular}{cccc}
	        \multicolumn{4}{c}{\bf Test data} \vspace{-0.0in}
	        \\
		\includegraphics[trim={{.9\linewidth} {.32\linewidth} {.8\linewidth} {.6\linewidth}}, clip, width=0.2\linewidth]
		{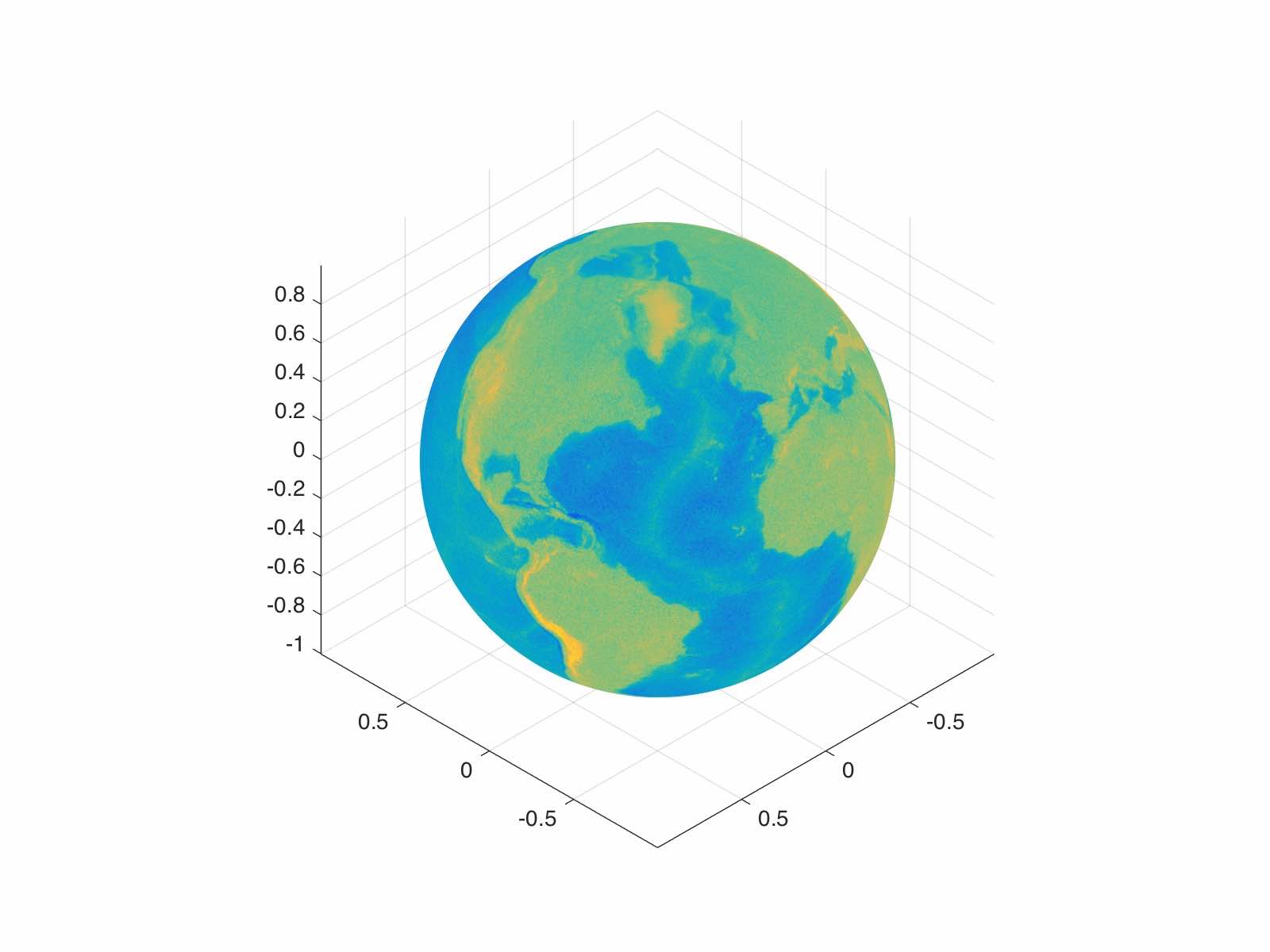} &
		\includegraphics[trim={{.28\linewidth} {.12\linewidth} {.19\linewidth} {.1\linewidth}}, clip, width=0.2\linewidth]
		{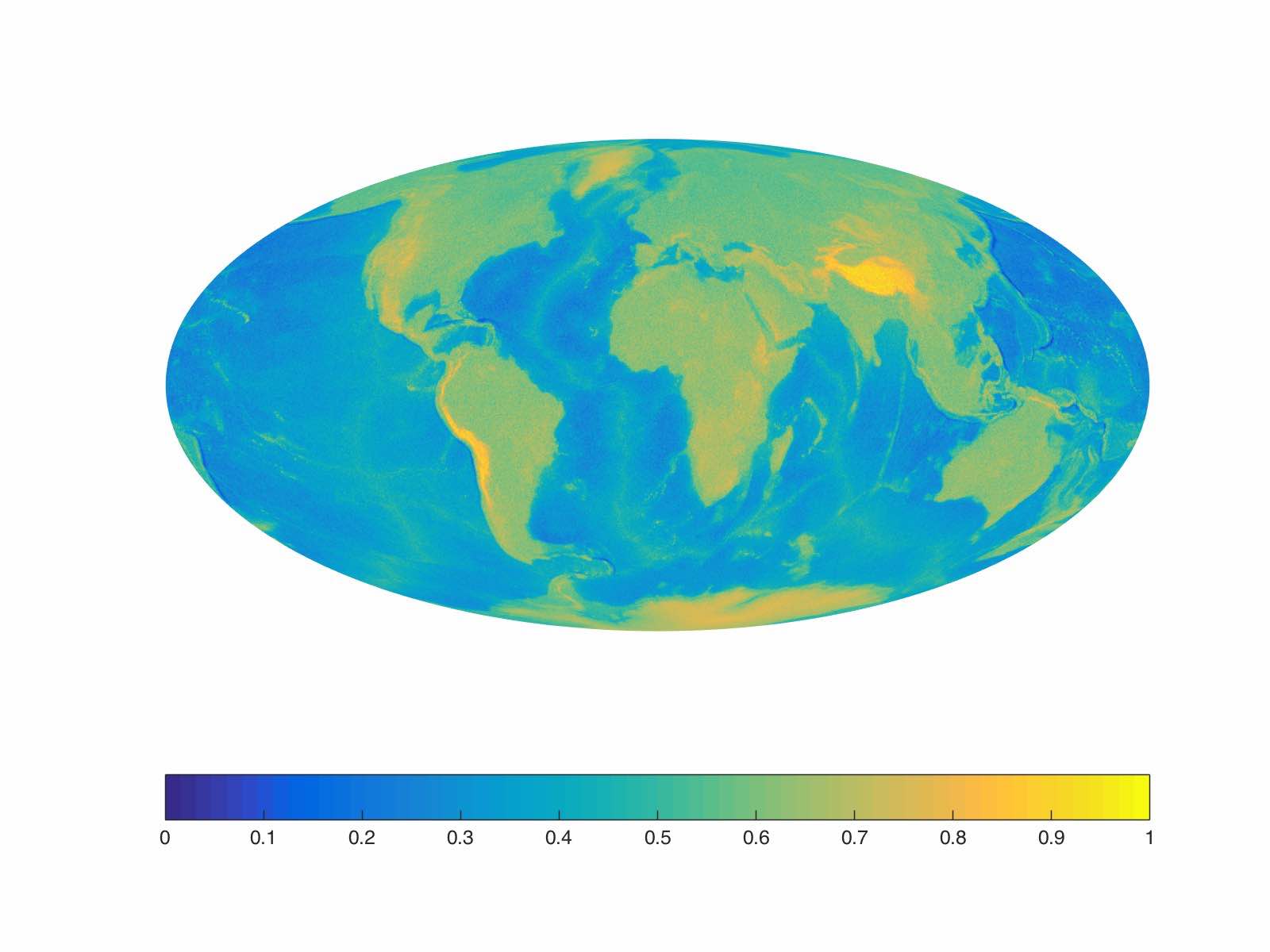}   \put(-40,24){\red{\framebox(14,10){ }}} & 
		\includegraphics[trim={{2.35\linewidth} {1.4\linewidth} {1.5\linewidth} {1.4\linewidth}}, clip, width=0.2\linewidth]
		{./fig/earth_map_AxisymWav_L_512_Noisy_image.jpg} &
		\includegraphics[trim={{2.35\linewidth} {1.4\linewidth} {1.5\linewidth} {1.4\linewidth}}, clip, width=0.2\linewidth]
		{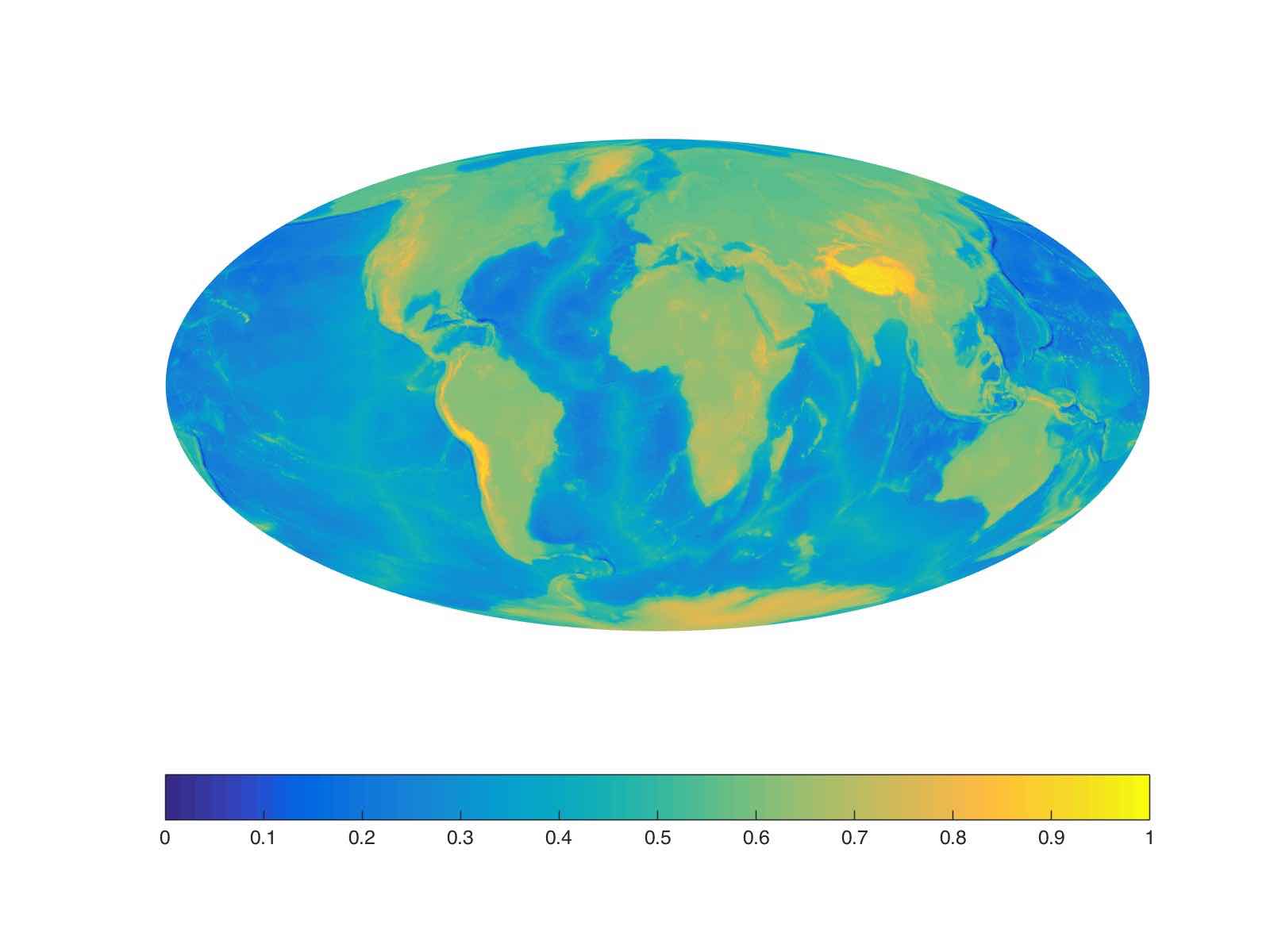}  
		\\
		{\small (a) noisy image} & {\small (b) noisy image} & {\small (c) noisy image} & {\small (d) original image}  \vspace{0.15in}
		\\
		 \multicolumn{4}{c}{\bf Segmentation results}  \vspace{-0.02in}
	        \\
	         \includegraphics[trim={{.9\linewidth} {.32\linewidth} {.8\linewidth} {.6\linewidth}}, clip, width=0.2\linewidth]
		{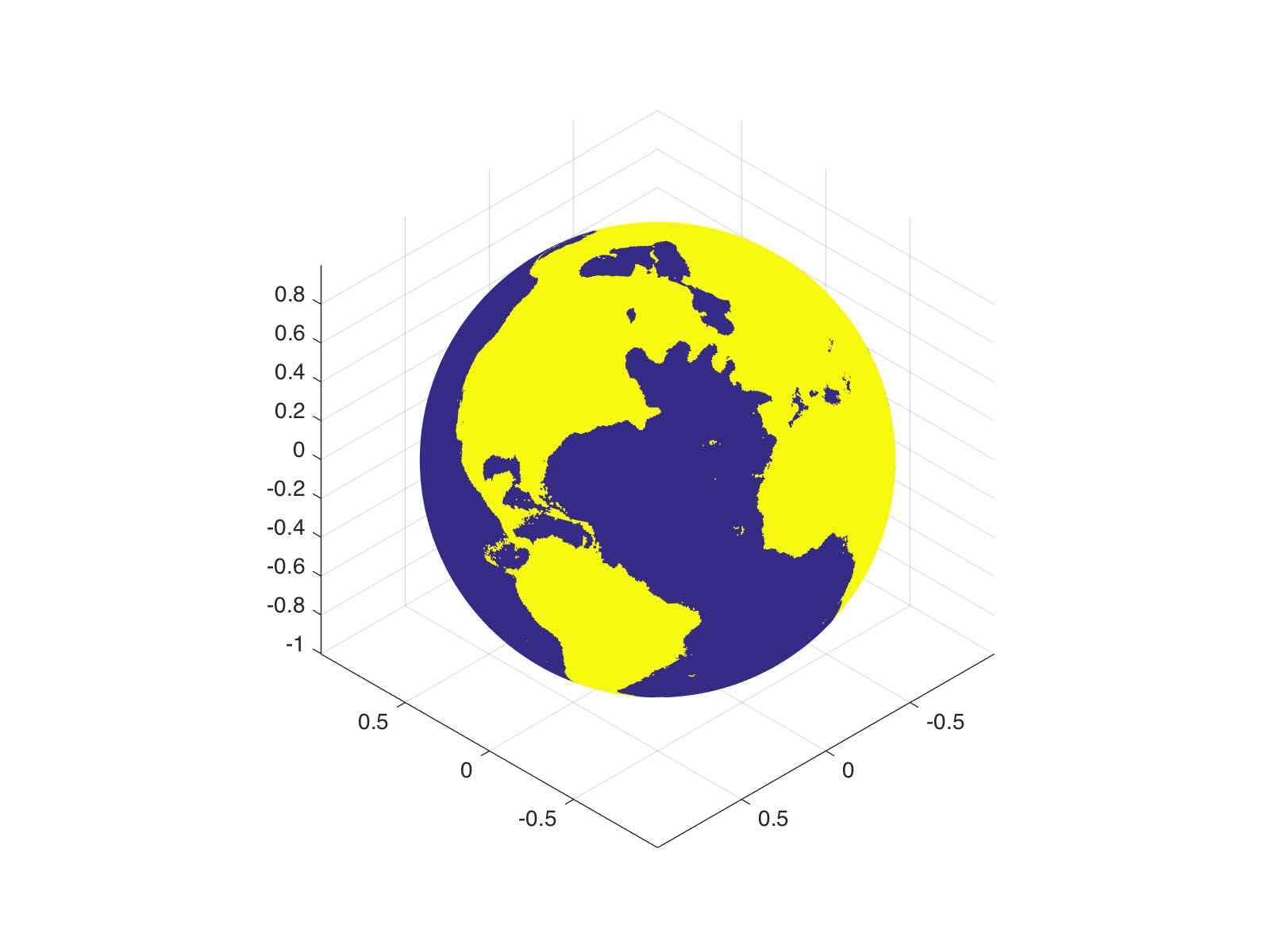} &
        		\includegraphics[trim={{.9\linewidth} {.32\linewidth} {.8\linewidth} {.6\linewidth}}, clip, width=0.2\linewidth]
		{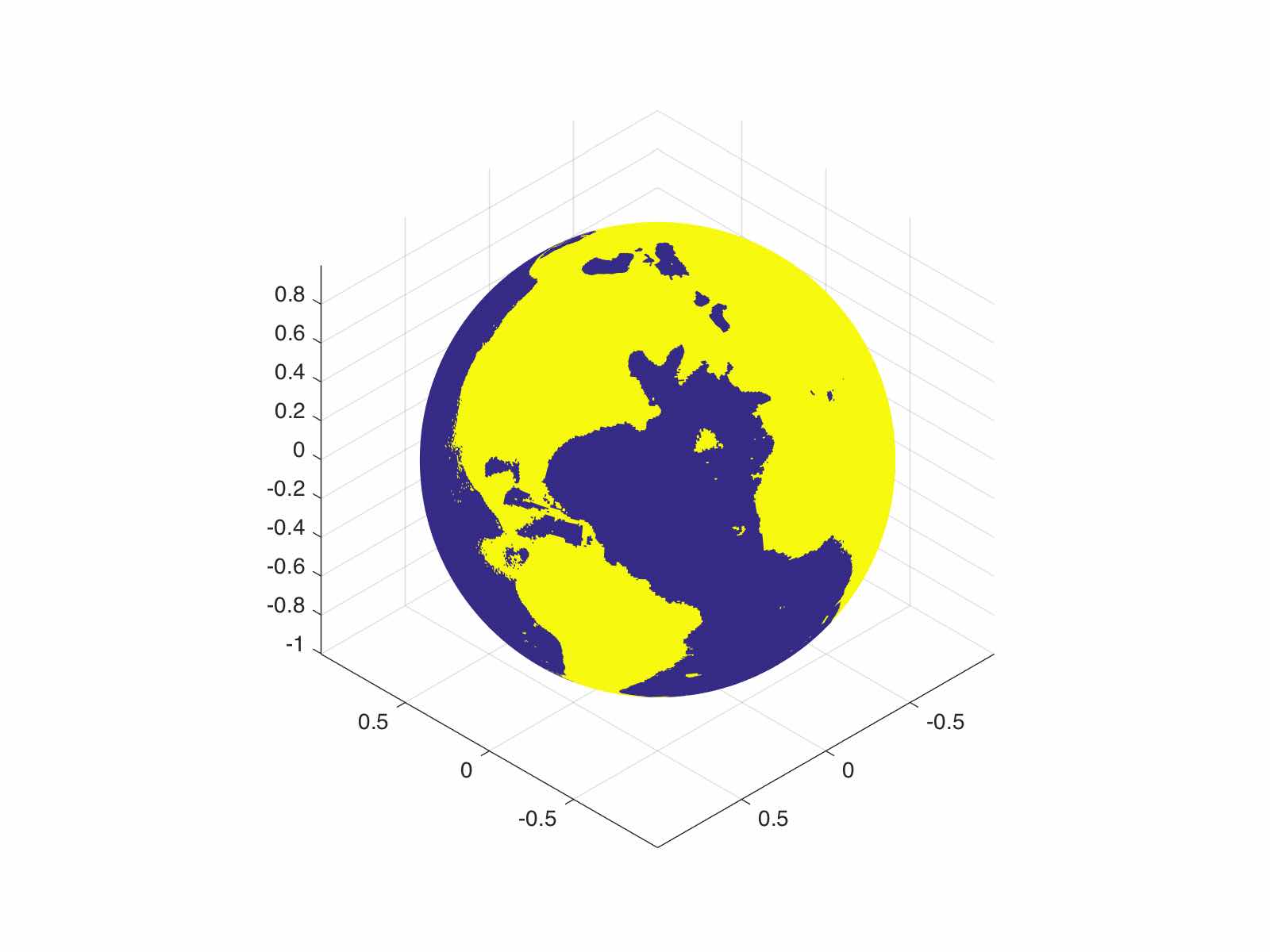} &
        		\includegraphics[trim={{.9\linewidth} {.32\linewidth} {.8\linewidth} {.6\linewidth}}, clip, width=0.2\linewidth]
		{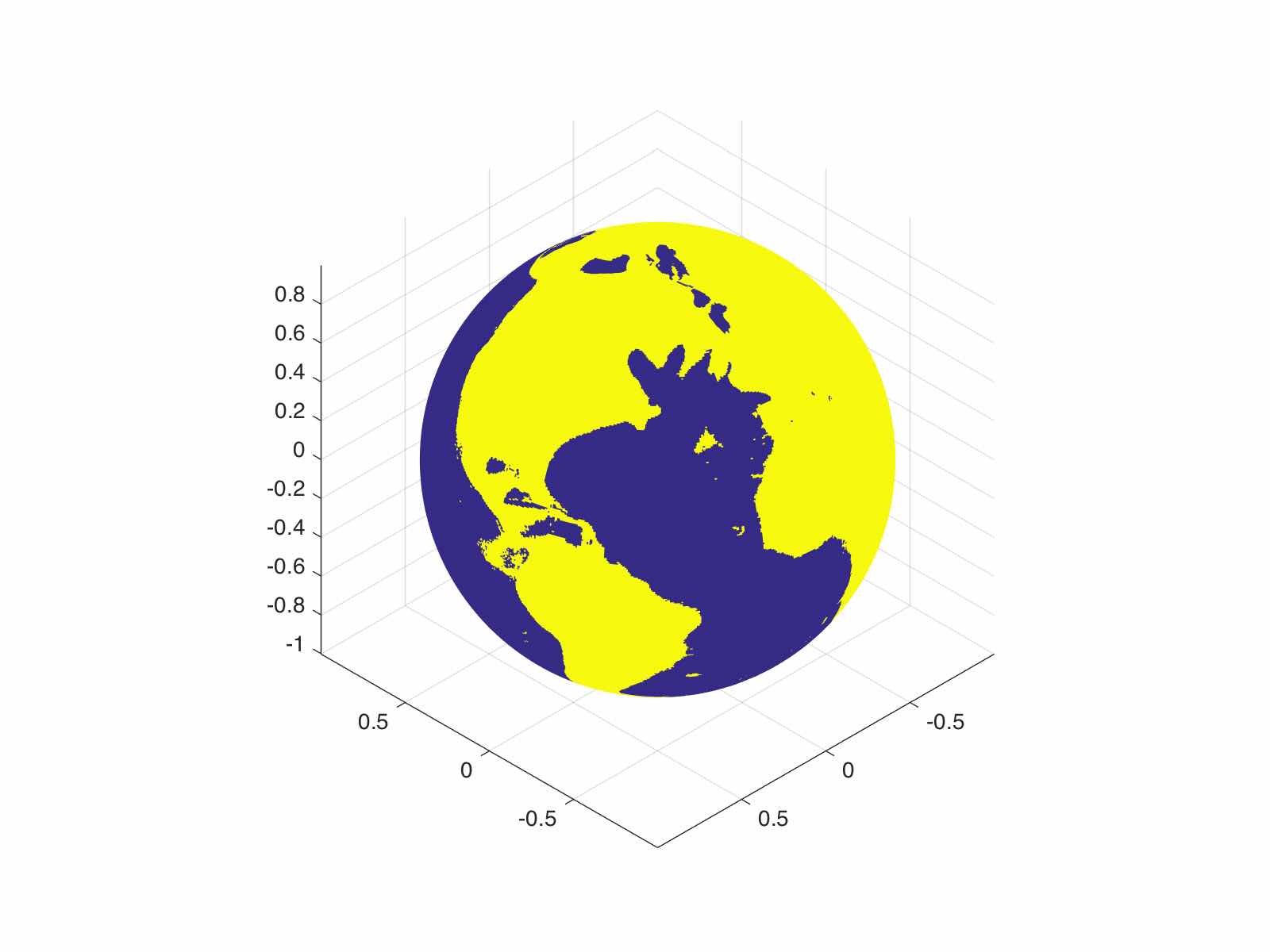} &
        		\includegraphics[trim={{.9\linewidth} {.32\linewidth} {.8\linewidth} {.6\linewidth}}, clip, width=0.2\linewidth]
		{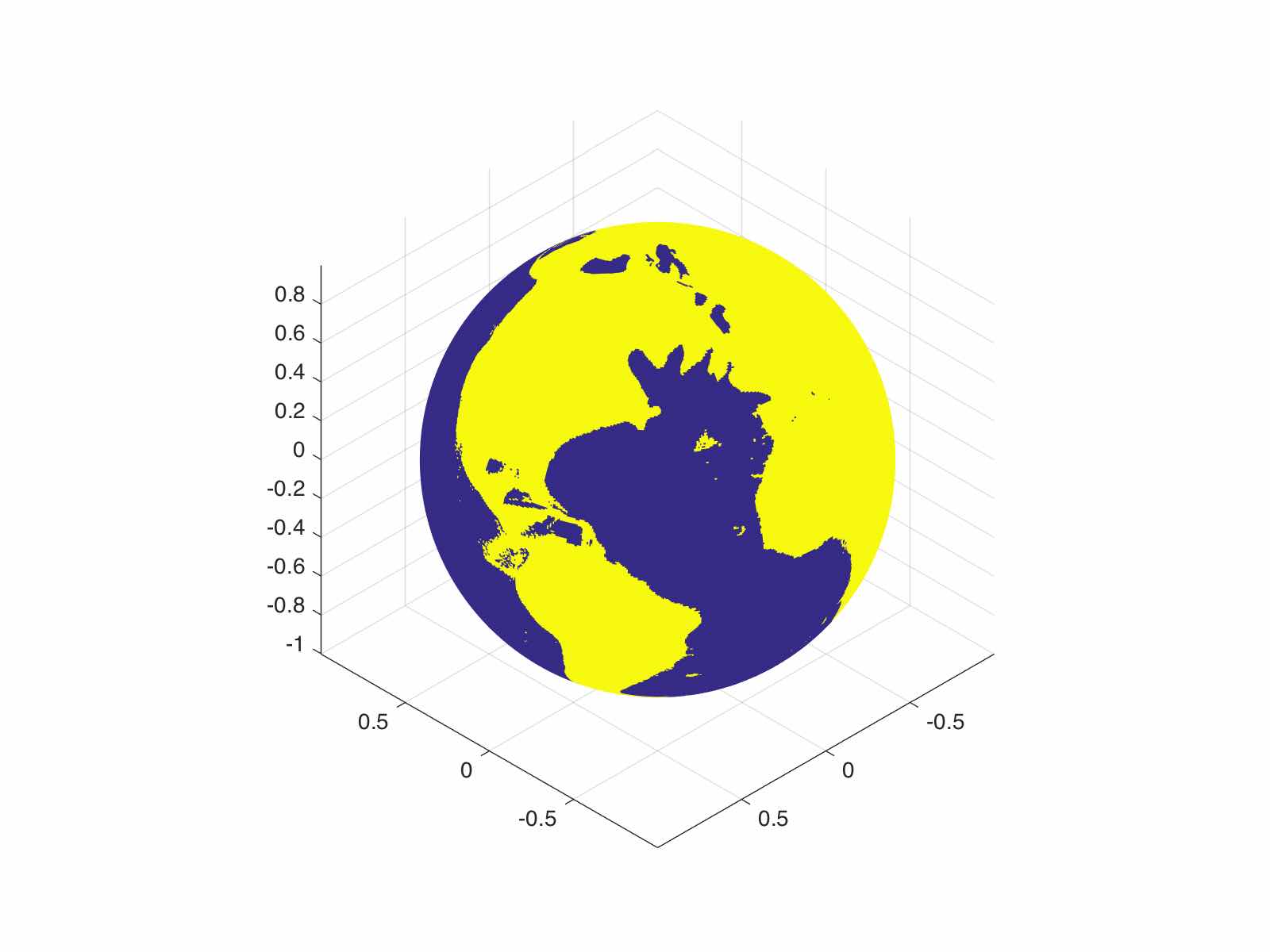} 
		\\
		\includegraphics[trim={{.5\linewidth} {.2\linewidth} {.3\linewidth} {.2\linewidth}}, clip, width=0.2\linewidth]
		{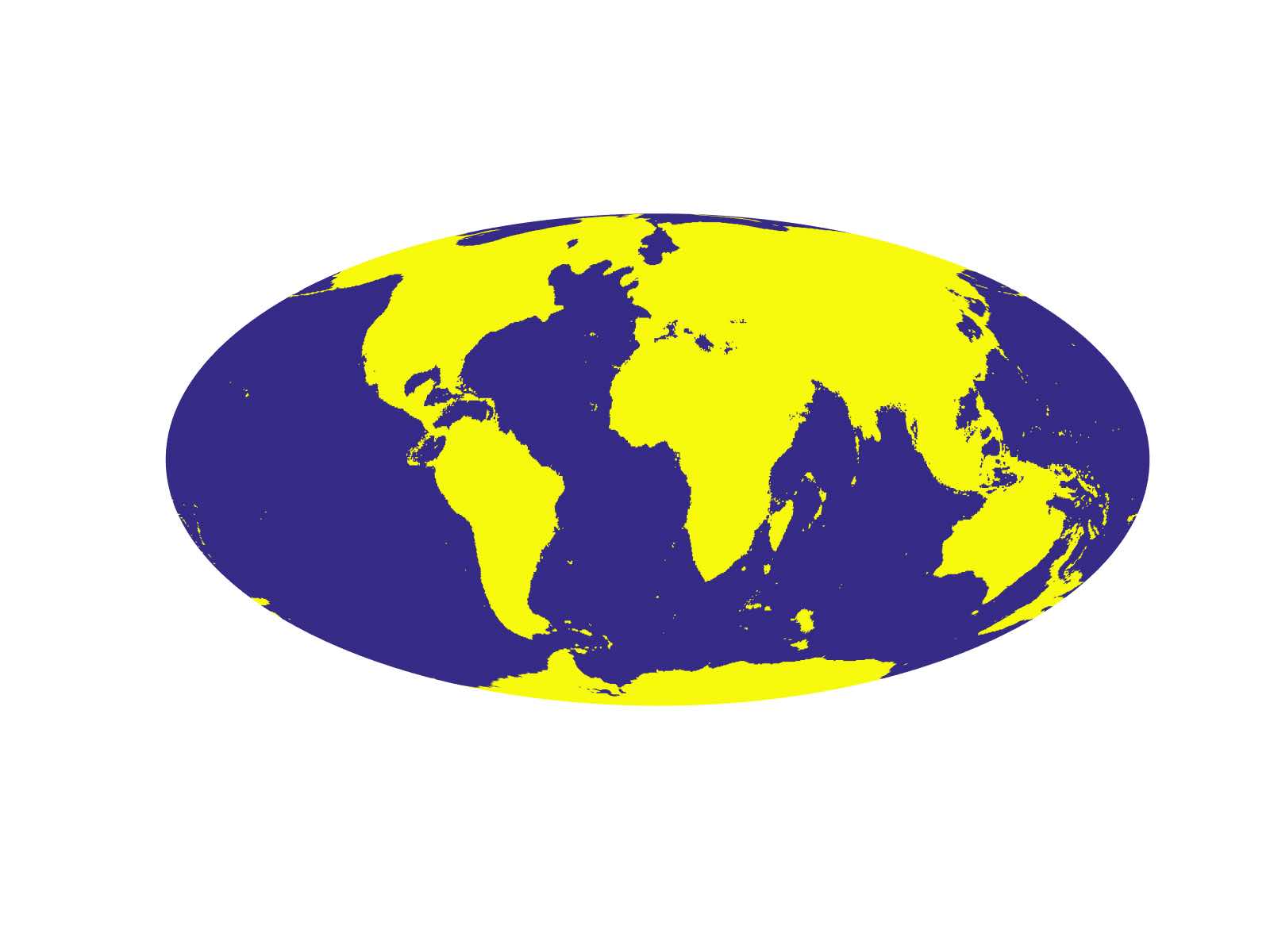}   \put(-35,19){\red{\framebox(12,12){ }}}  &
        		\includegraphics[trim={{.5\linewidth} {.2\linewidth} {.3\linewidth} {.2\linewidth}}, clip, width=0.2\linewidth]
		{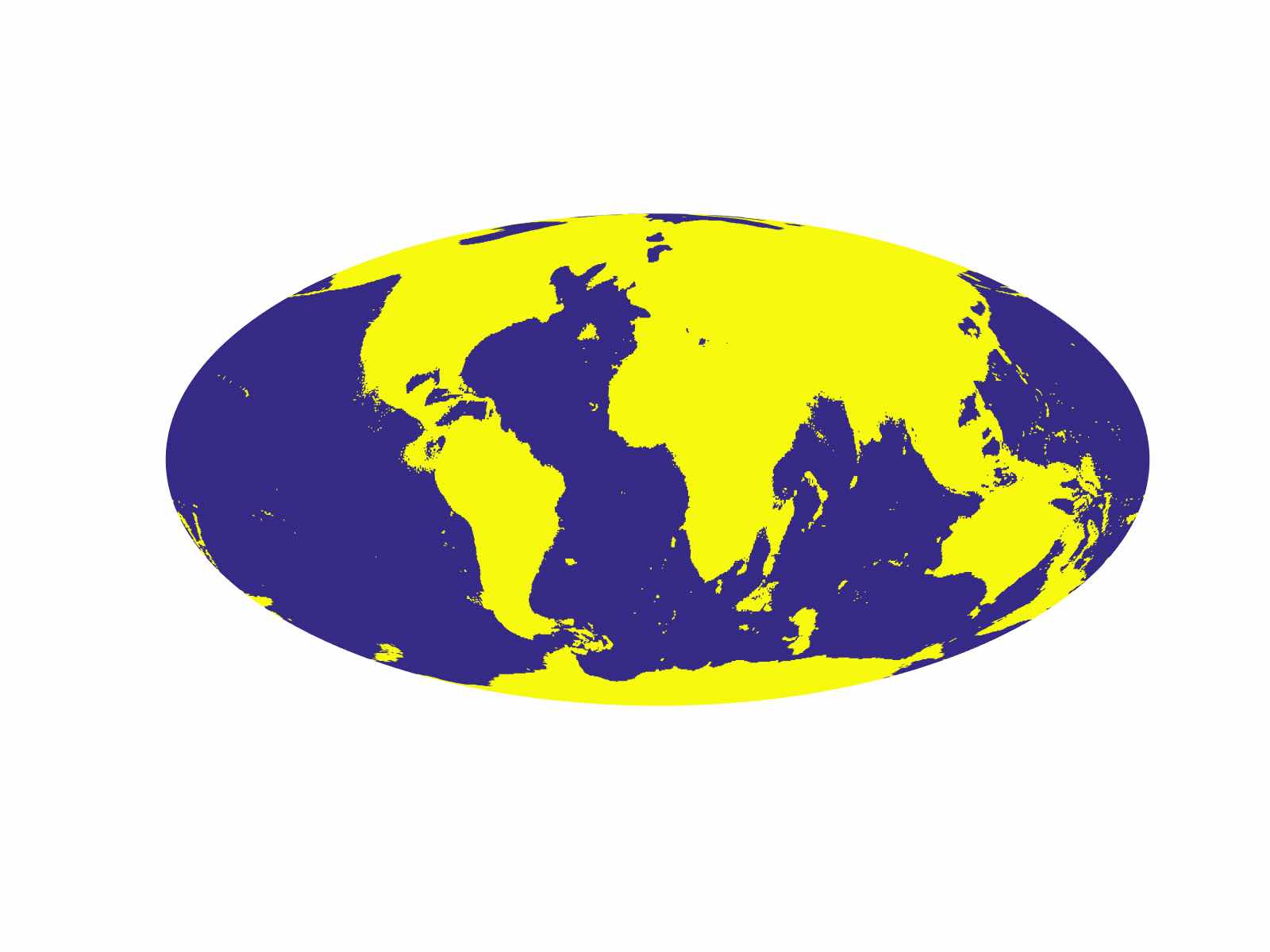}   \put(-35,19){\red{\framebox(12,12){ }}} &
                 \includegraphics[trim={{.5\linewidth} {.2\linewidth} {.3\linewidth} {.2\linewidth}}, clip, width=0.2\linewidth]
                 {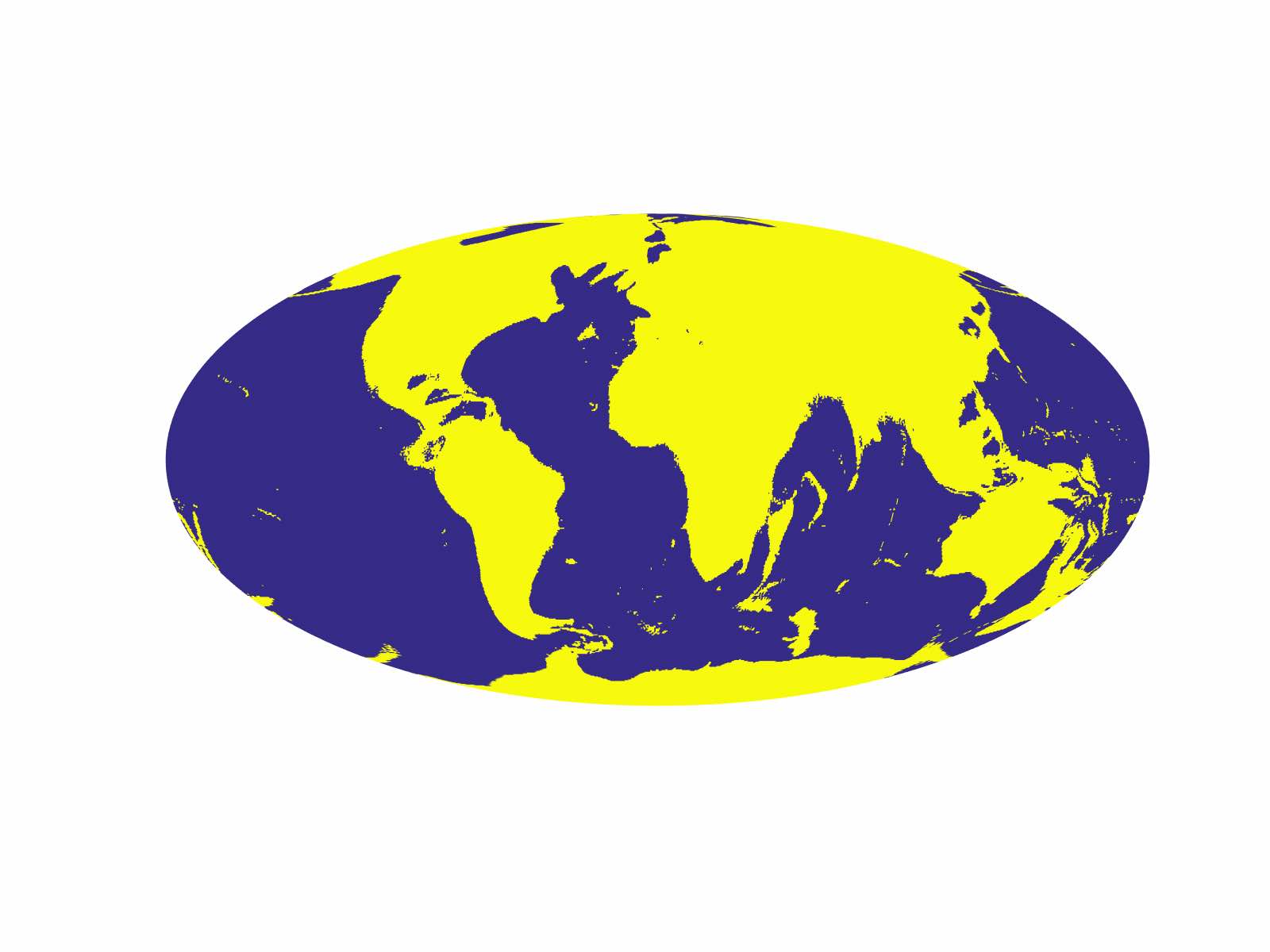}   \put(-35,19){\red{\framebox(12,12){ }}} &
                 \includegraphics[trim={{.5\linewidth} {.2\linewidth} {.3\linewidth} {.2\linewidth}}, clip, width=0.2\linewidth]
                 {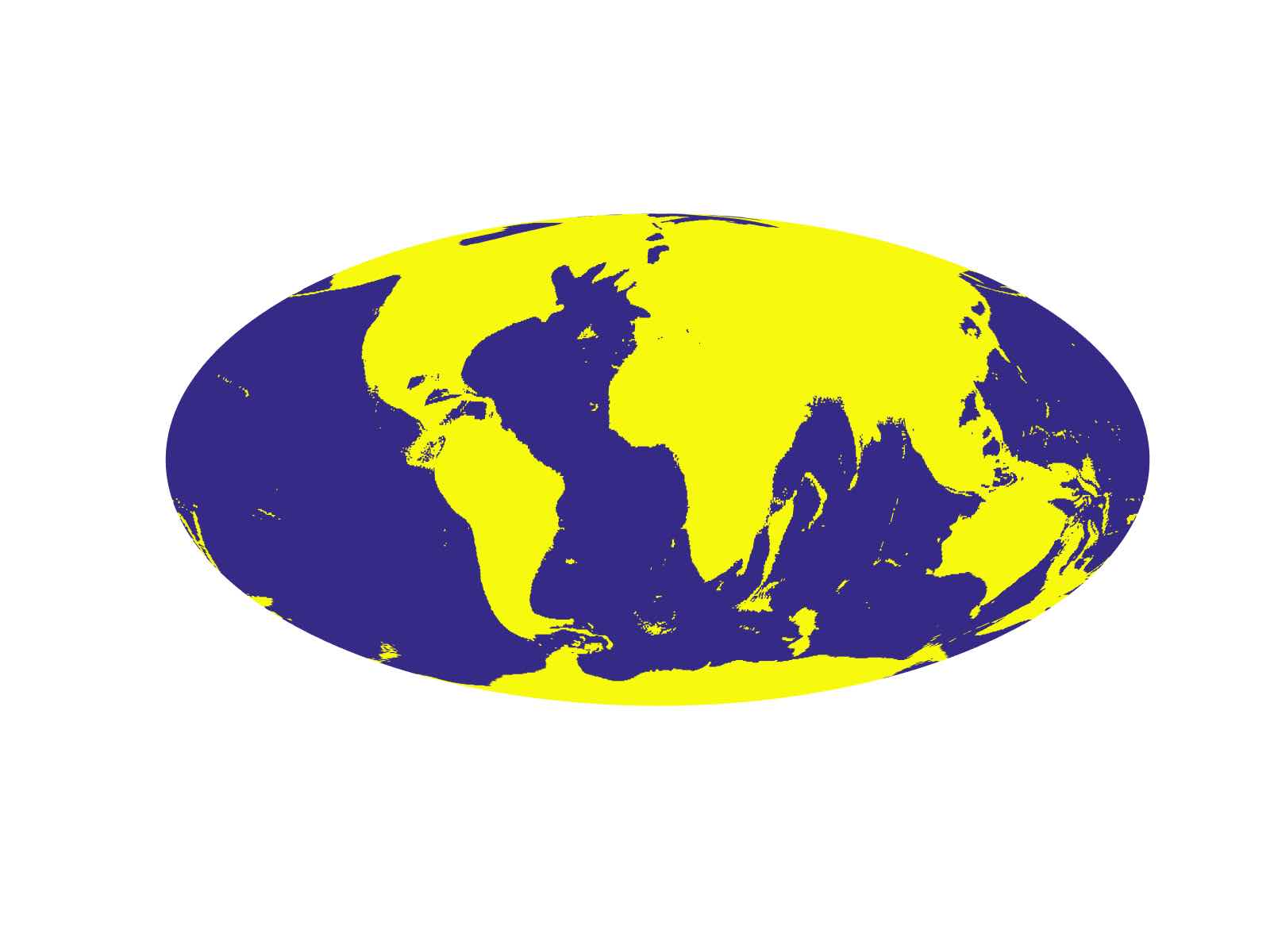}   \put(-35,19){\red{\framebox(12,12){ }}}
                 \\
                 \includegraphics[trim={{2.35\linewidth} {1.1\linewidth} {1.4\linewidth} {1.6\linewidth}}, clip, width=0.2\linewidth]
		{./fig/earth_map_AxisymWav_L_512_Resutls_Th.jpg}  &
                 \includegraphics[trim={{2.35\linewidth} {1.1\linewidth} {1.4\linewidth} {1.6\linewidth}}, clip, width=0.2\linewidth]
		{./fig/earth_map_AxisymWav_L_512_Resutls.jpg}   &
                 \includegraphics[trim={{2.35\linewidth} {1.1\linewidth} {1.4\linewidth} {1.6\linewidth}}, clip, width=0.2\linewidth]
                 {./fig/earth_map_DirectionWav_L_512_N_5_Resutls.jpg}  &
                 \includegraphics[trim={{2.35\linewidth} {1.1\linewidth} {1.4\linewidth} {1.6\linewidth}}, clip, width=0.2\linewidth]
                 {./fig/earth_map_HybridWav_L_512_N_6_Resutls.jpg}  
                 \\
		{\small (e) K-means} & {\small (f) WSSA-A} & {\small (g) WSSA-D} & {\small (h) WSSA-H}
        \end{tabular}
	\caption{Results of the Earth topographic map.  
	First row: noisy image shown on the sphere (a) and in 2D using a mollweide projection (b), and the zoomed-in red rectangle area 
	of the noisy (c) and original images (d), respectively; 
	Second to fourth rows from left to right: results of methods K-means (e), WSSA-A (f), WSSA-D (g) with $N=5$ (odd $N$), 
	and WSSA-H (h), respectively.}
	\label{fig-earthmap}
\end{figure}

\begin{table}[h] 
\begin{center}
\caption{Earth map in Fig.\ \ref{fig-earthmap}: Number of unclassified points at each iteration 
{(i.e. $|\Lambda^{(i)}|$, where $\Lambda^{(i)}$ is the set of unclassified points and $|\cdot |$ denotes the cardinality of a set)} 
and computation time in seconds.
	$^\ast$The fourth and fifth columns represent the results of WSSA-D with $N = 5$ and 6, respectively.} \label{tab:earth}
\begin{tabular}{|c||c|c|c|c|c|}
\hline 
  & {K-means} & {WSSA-A} & WSSA-D$^\ast$ & WSSA-D$^\ast$ & WSSA-H
\\ \hline \hline
\raisebox{-.15ex}{ $|\bar{\mathbb{S}}^2|$ }&  $523776$ & $523776$ & $523776$  & $523776$  & $523776$ 
\\ \hline
 $|\Lambda^{(0)}|$ & -  & 111371 & 110373   & 111854   & 111184
\\ \hline
$|\Lambda^{(1)}|$ &   -  & 106977   &  104381  &  105222  &  106946
\\ \hline
$|\Lambda^{(2)}|$ &  -  & 25880 &  25938  &   26387  &  27681
\\ \hline
$|\Lambda^{(3)}|$ &  -  & 6352   & 6750   & 6645    &  6940
\\ \hline
$|\Lambda^{(4)}|$ &  -  & 1824   &  1995  &  1937  &  1972
\\ \hline
$|\Lambda^{(5)}|$ &  -  & 615    & 680  &  668  &  664
\\ \hline
$|\Lambda^{(6)}|$ &  -  & 229   & 247  & 269 & 254
\\ \hline
$|\Lambda^{(7)}|$ &  -   &  96  & 89  & 97  & 83
\\ \hline
$|\Lambda^{(8)}|$ &  -   &  28  & 26  &  38 & 27
\\ \hline
$|\Lambda^{(9)}|$ &  -   &  5  & 7  & 12  & 9
\\ \hline
$|\Lambda^{(10)}|$ &  -   &  0  & 2  &  2 & 0
\\ \hline
$|\Lambda^{(11)}|$ &  -   &  -  & 0  &  0 & -
\\ \hline \hline
Time & $<$ 1 s   &  51.9 s  & 200.5 s &  217.2 s & 883.5 s
\\ \hline
\end{tabular}
\end{center}
\end{table}

Fig.\ \ref{fig-earthmap} shows the results of the K-means and our WSSA (-A, -D, and -H) method with $\epsilon= 0.02$ used in \eqref{lambda0} 
to obtain the initial set $\Lambda^{(0)}$. Fig.\ \ref{fig-earthmap} (a) and (b) are the test noisy image corrupted by Gaussian noise  
shown on the sphere and in 2D using a mollweide projection, respectively. Fig.\ \ref{fig-earthmap} (c) shows the zoomed-in details of the red 
rectangle in Fig.\ \ref{fig-earthmap} (b).
For easy of comparison, Fig.\ \ref{fig-earthmap} (d) shows the same zoomed-in area of Fig.\ \ref{fig-earthmap} (c) with no noise added.  
The second to the fourth  rows present the segmentation results shown on the sphere, in 2D using a mollweide projection, and with the zoomed-in details 
of the red rectangle area, respectively. From the results, we see that all the methods give reasonable segmentation results, 
i.e., the land and oceans are separated quite well (it should be noted that the separation is not necessarily strictly into land and sea as there is no requirement  
for the shore line to be the segmentation boundary). From the zoomed-in details, we see the WSSA method
(Fig.\ \ref{fig-earthmap} (f)--(h)) produces a better segmentation than the K-means method (Fig.\ \ref{fig-earthmap} (e)).
We also see that the WSSA method equipped with directional wavelets (Fig.\ \ref{fig-earthmap} (g))) and hybrid wavelets (Fig.\ \ref{fig-earthmap} (h)) 
are marginally better than using the axisymmetric wavelets (Fig.\ \ref{fig-earthmap} (f)) in terms of preserving directional features in these data. 
WSSA-D and WSSA-H methods give very similar results (note that the hybrid method comprises the directional wavelets as a 
major component in this test).

Table \ref{tab:earth} gives the number of unclassified points at each iteration ($|\Lambda^{(i)}|$) of the WSSA method and 
the computation time  in seconds,
from which we can see the WSSA method takes about 10 iterations to converge in general,
where each iteration takes roughly the same amount of computation time since in each iteration the computation time is 
dominated by a complete round-trip of wavelet transforms (which has yet to be optimised utilising the data structure in the proposed segmentation method). 
Note that after the third iteration, the number of unclassified 
pixels is already very low compared with the whole number of pixels on the sphere $|\bar{\mathbb{S}}^2|$. 
Moreover, from Table \ref{tab:earth}, we see the WSSA-H method needs the longest computation time, while the K-means needs the shortest;
for the WSSA-D method, the greater $N$, the longer the computation time required.
 
\begin{figure}
	\centering
	\begin{tabular}{cccc}
	        \multicolumn{4}{c}{\bf Test data} \vspace{-0.0in}
	        \\
		\includegraphics[trim={{.9\linewidth} {.32\linewidth} {.8\linewidth} {.6\linewidth}}, clip, width=0.2\linewidth]
		{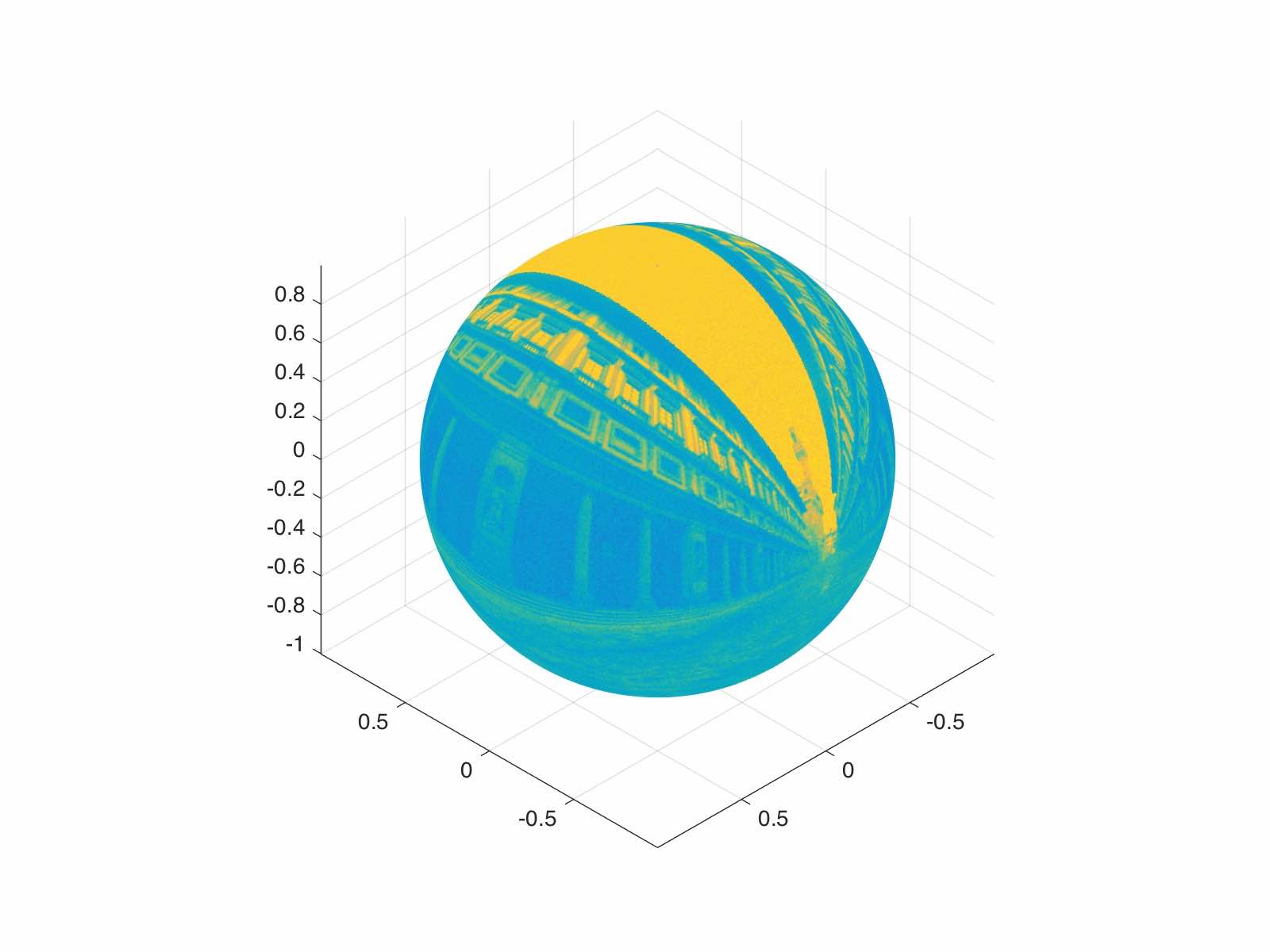} &
		\includegraphics[trim={{.35\linewidth} {.22\linewidth} {.29\linewidth} {.2\linewidth}}, clip, width=0.2\linewidth]
		{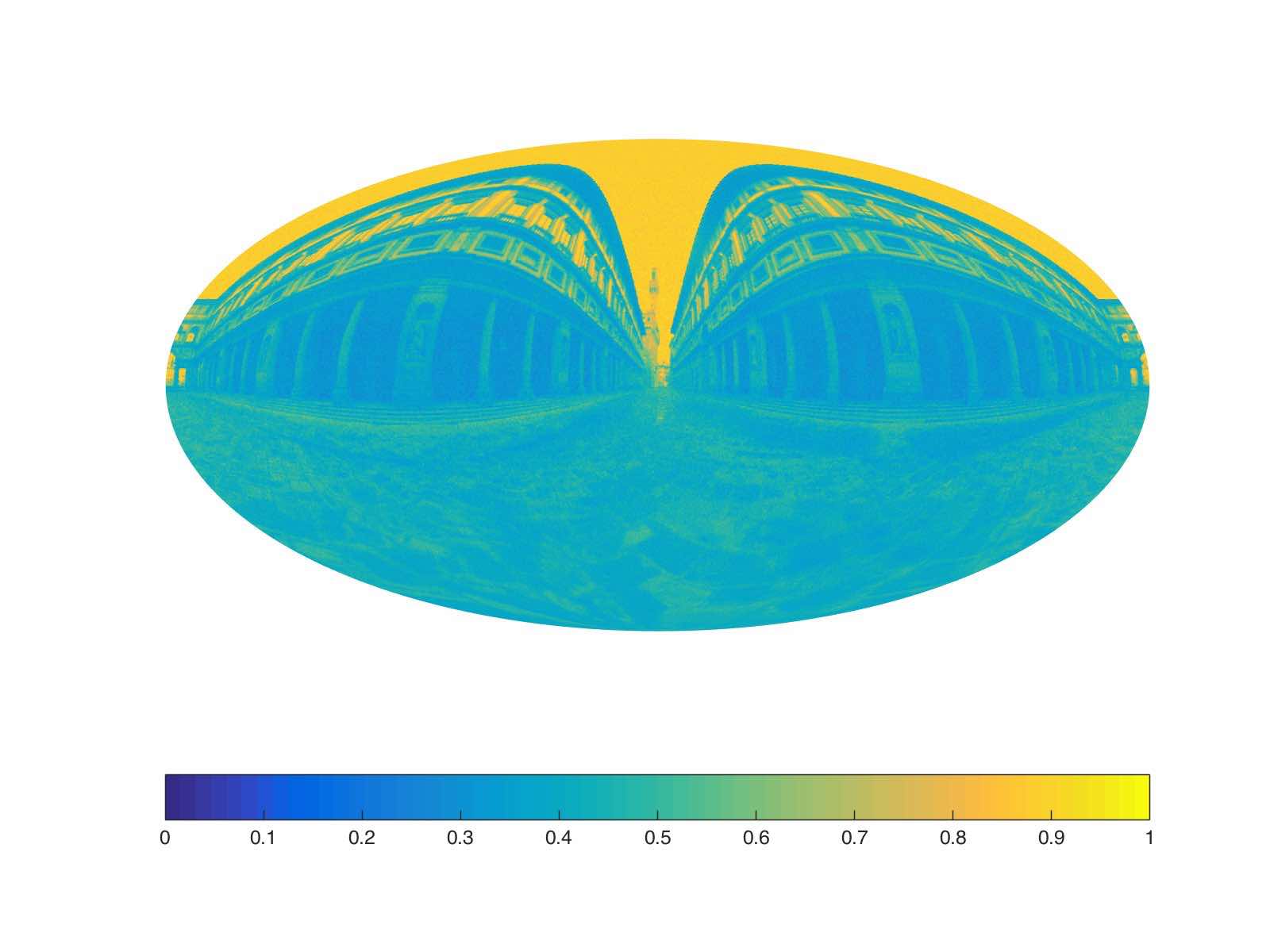}   \put(-51.5,37){\red{\framebox(14,10){ }}} & 
		\includegraphics[trim={{1.5\linewidth} {2.0\linewidth} {2.2\linewidth} {0.7\linewidth}}, clip, width=0.2\linewidth]
		{./fig/light_probe_uffizi_AxisymWav_L_512_Noisy_image.jpg} &
		\includegraphics[trim={{1.5\linewidth} {2.0\linewidth} {2.2\linewidth} {0.7\linewidth}}, clip, width=0.2\linewidth]
		{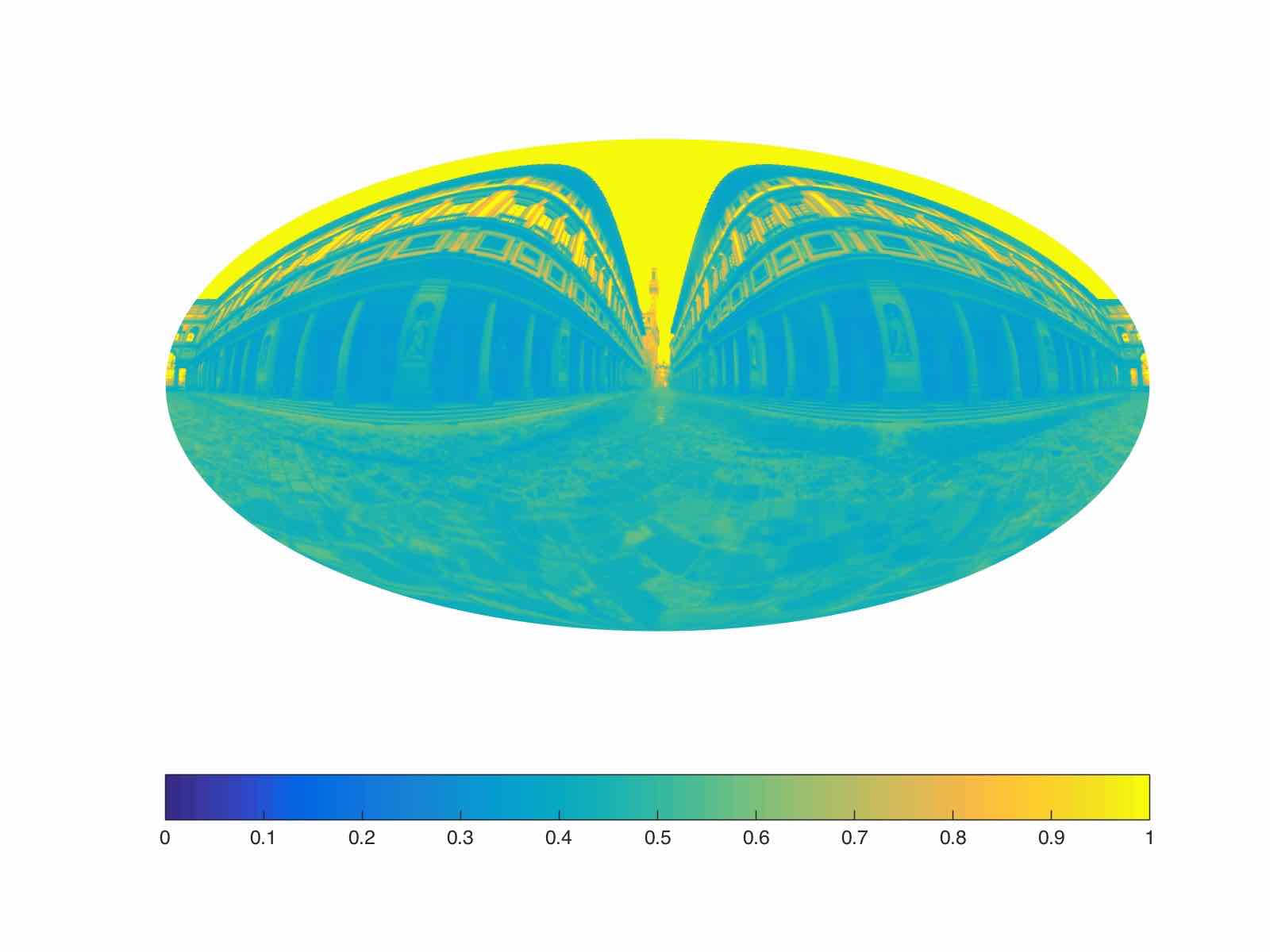}  
		\\
		{\small (a) noisy image} & {\small (b) noisy image} & {\small (c) noisy image} & {\small (d) original image}  \vspace{0.15in}
		\\
		 \multicolumn{4}{c}{\bf Segmentation results}  \vspace{-0.02in}
	        \\
	         \includegraphics[trim={{.9\linewidth} {.32\linewidth} {.8\linewidth} {.6\linewidth}}, clip, width=0.2\linewidth]
		{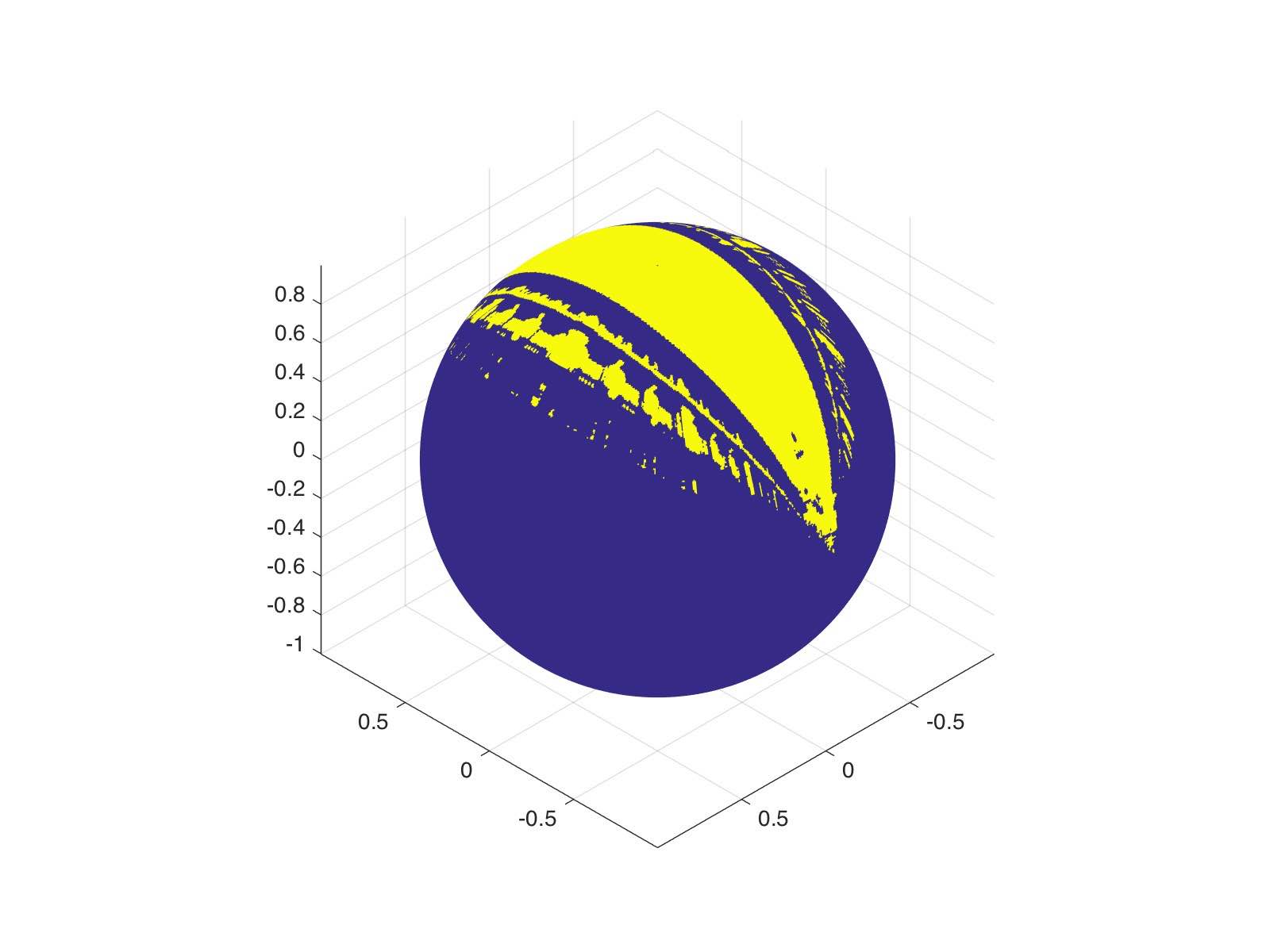} &
        		\includegraphics[trim={{.9\linewidth} {.32\linewidth} {.8\linewidth} {.6\linewidth}}, clip, width=0.2\linewidth]
		{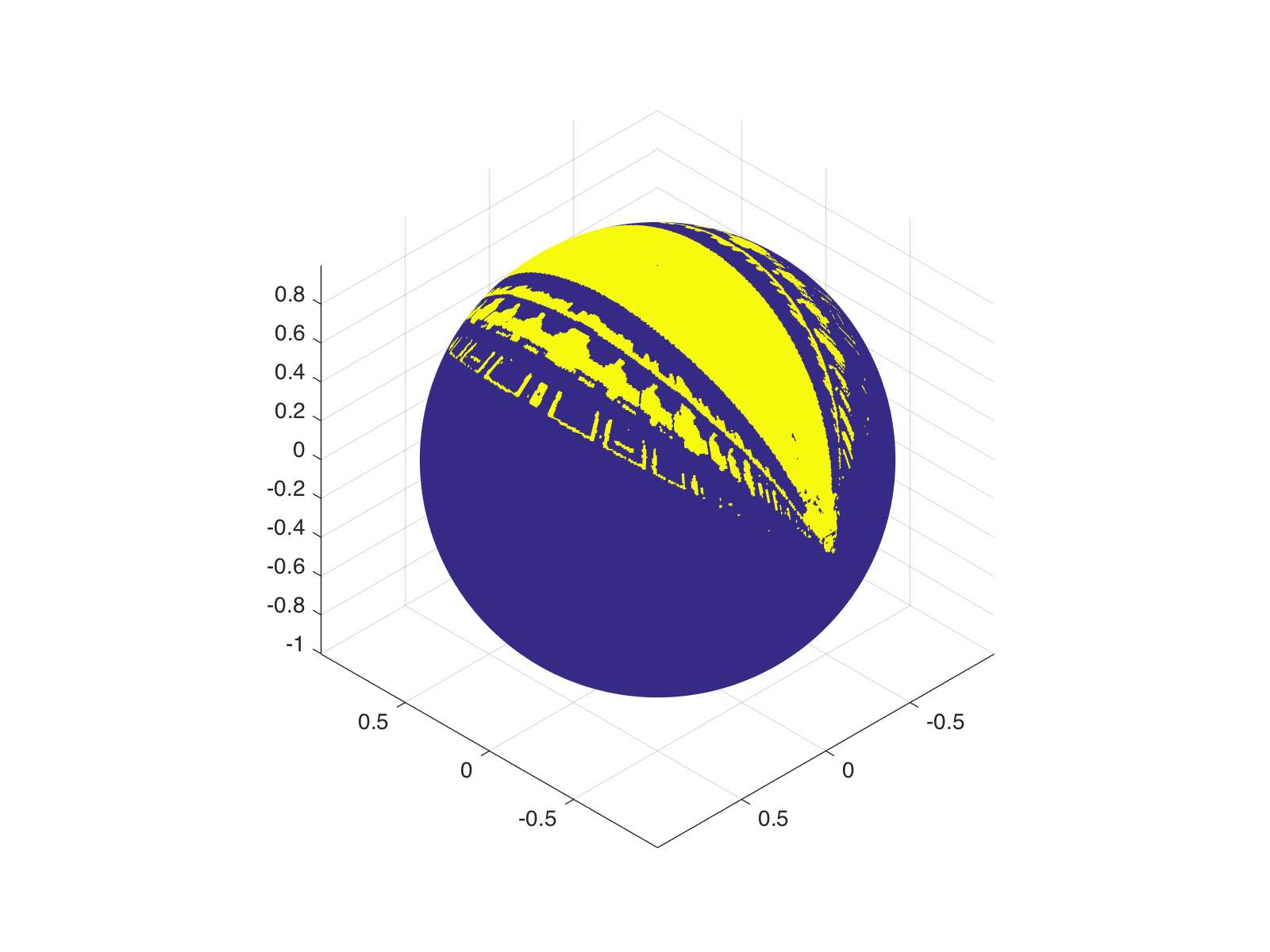} &
        		\includegraphics[trim={{.9\linewidth} {.32\linewidth} {.8\linewidth} {.6\linewidth}}, clip, width=0.2\linewidth]
		{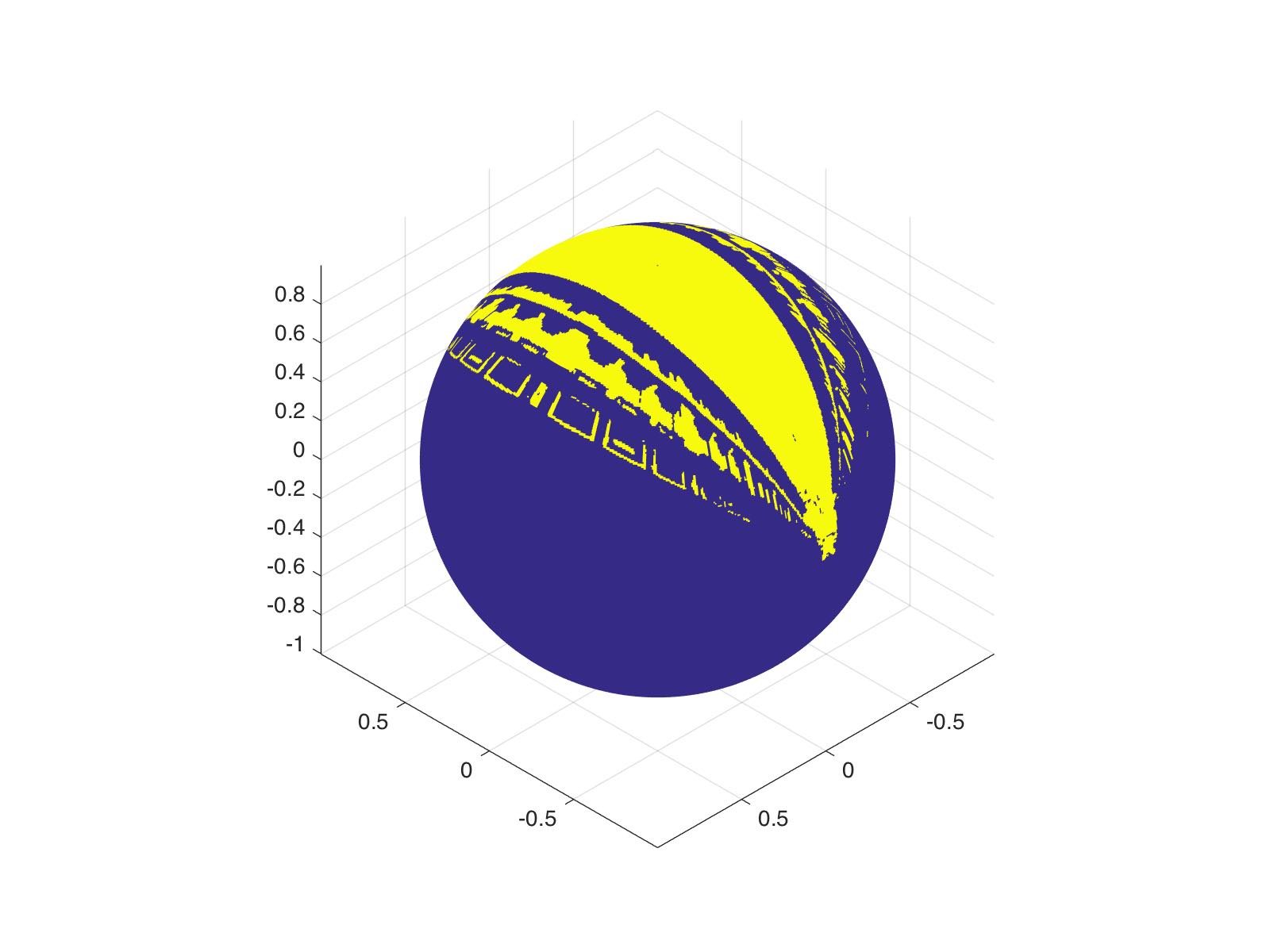} &
        		\includegraphics[trim={{.9\linewidth} {.32\linewidth} {.8\linewidth} {.6\linewidth}}, clip, width=0.2\linewidth]
		{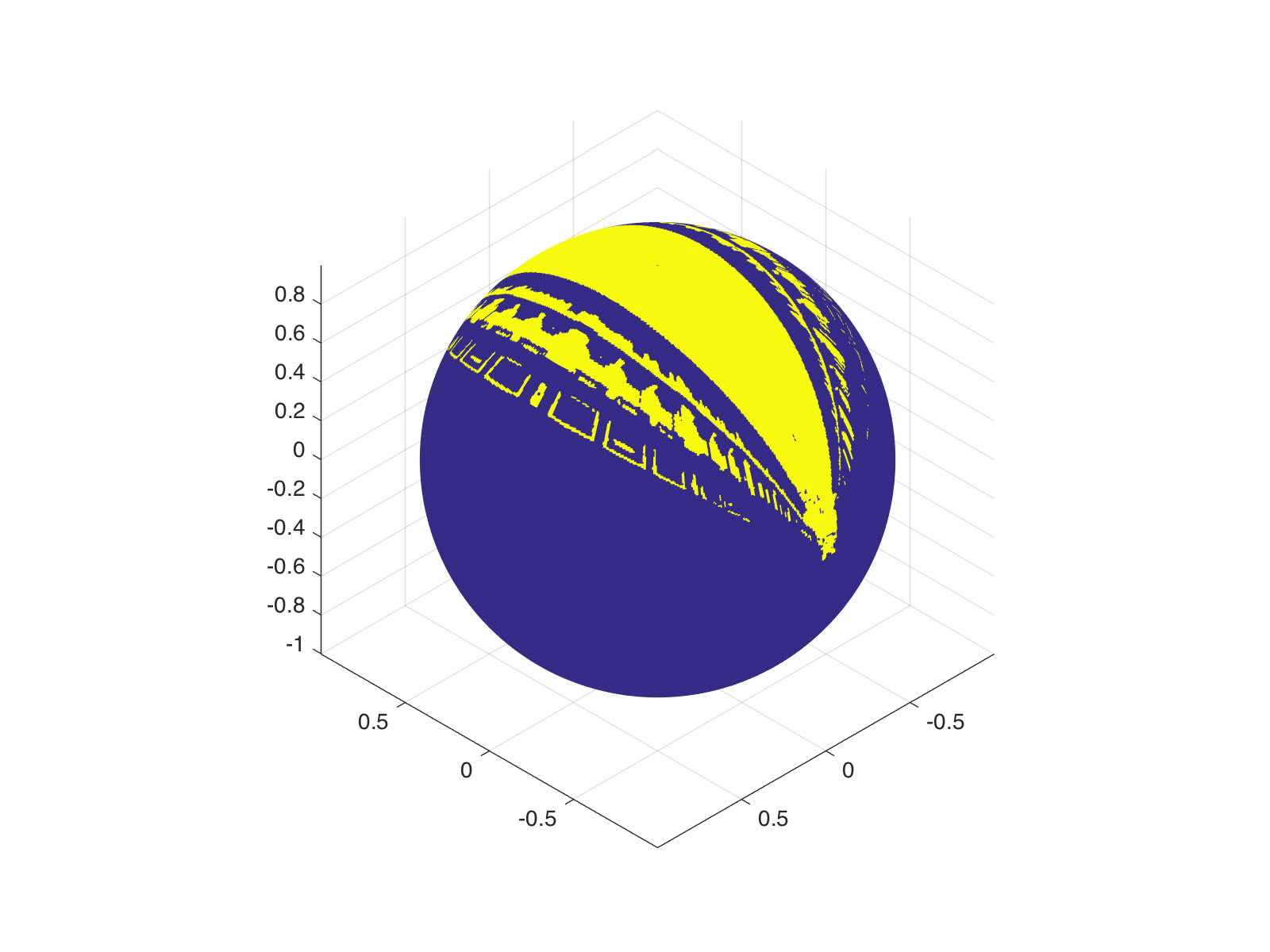} 
		\\
		\includegraphics[trim={{.5\linewidth} {.2\linewidth} {.3\linewidth} {.2\linewidth}}, clip, width=0.2\linewidth]
		{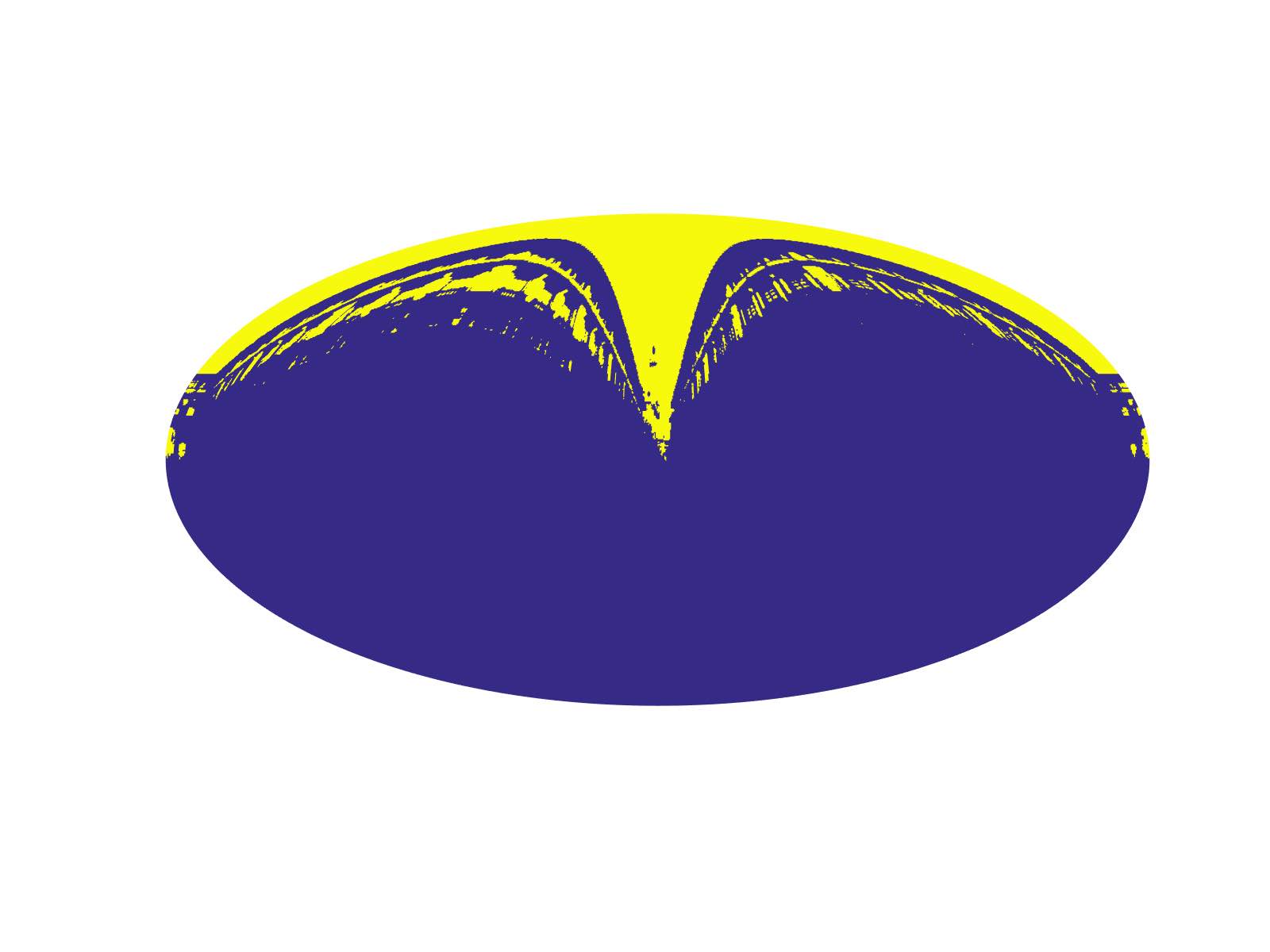}  \put(-53,35){\red{\framebox(14,10){ }}}  &
        		\includegraphics[trim={{.5\linewidth} {.2\linewidth} {.3\linewidth} {.2\linewidth}}, clip, width=0.2\linewidth]
		{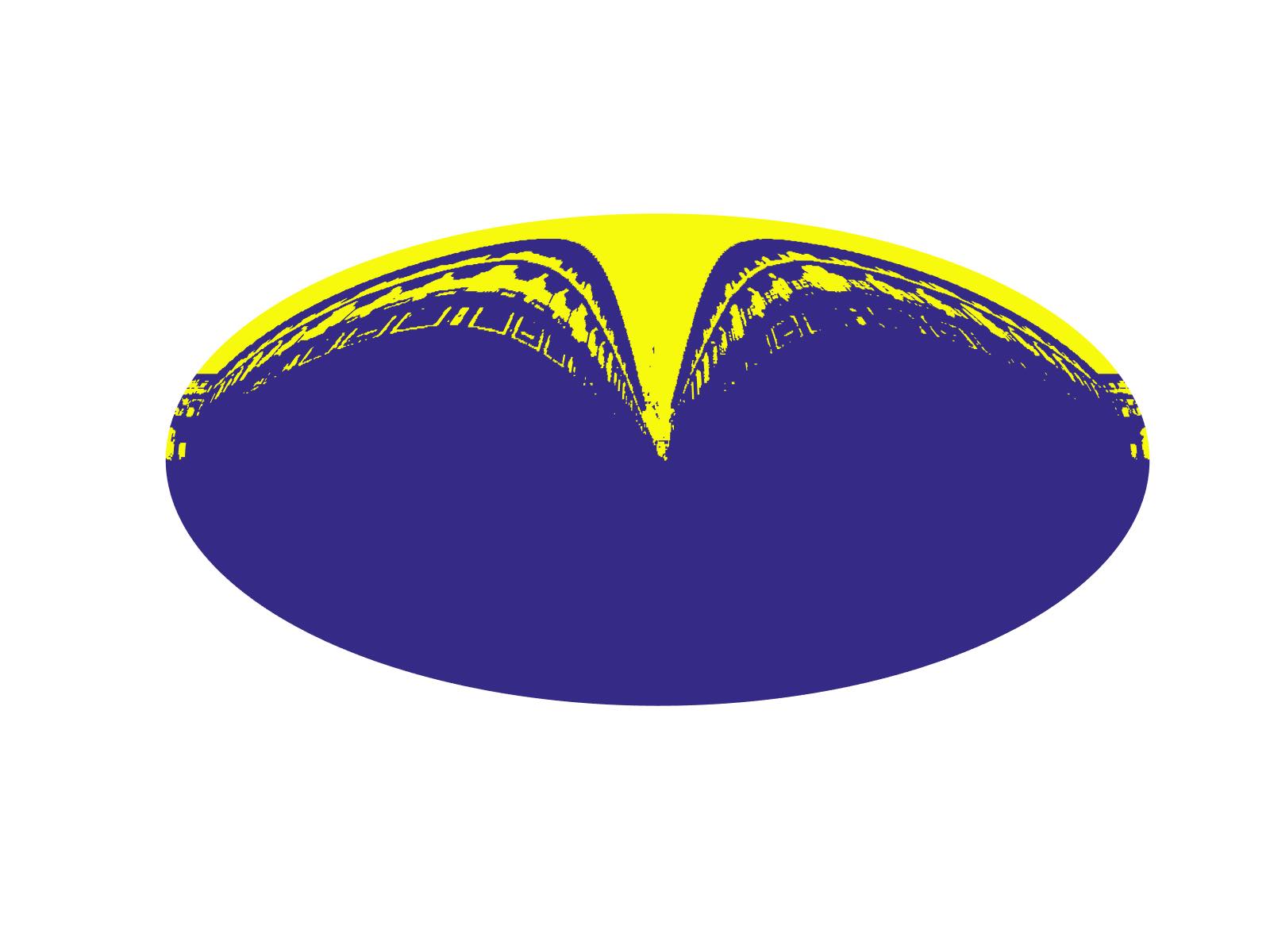}   \put(-53,35){\red{\framebox(14,10){ }}}  &
                 \includegraphics[trim={{.5\linewidth} {.2\linewidth} {.3\linewidth} {.2\linewidth}}, clip, width=0.2\linewidth]
                 {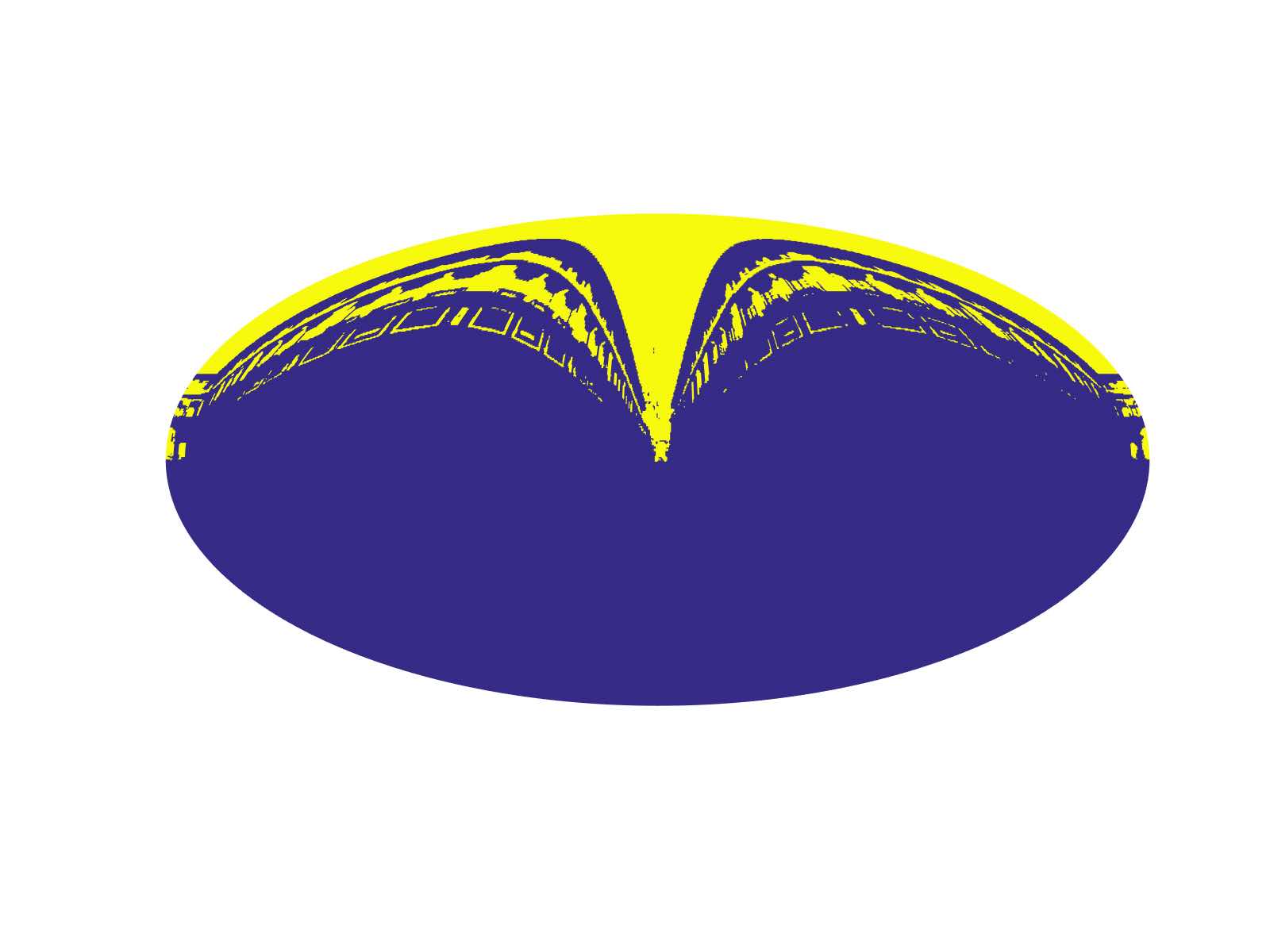}   \put(-53,35){\red{\framebox(14,10){ }}}   &
                 \includegraphics[trim={{.5\linewidth} {.2\linewidth} {.3\linewidth} {.2\linewidth}}, clip, width=0.2\linewidth]
                 {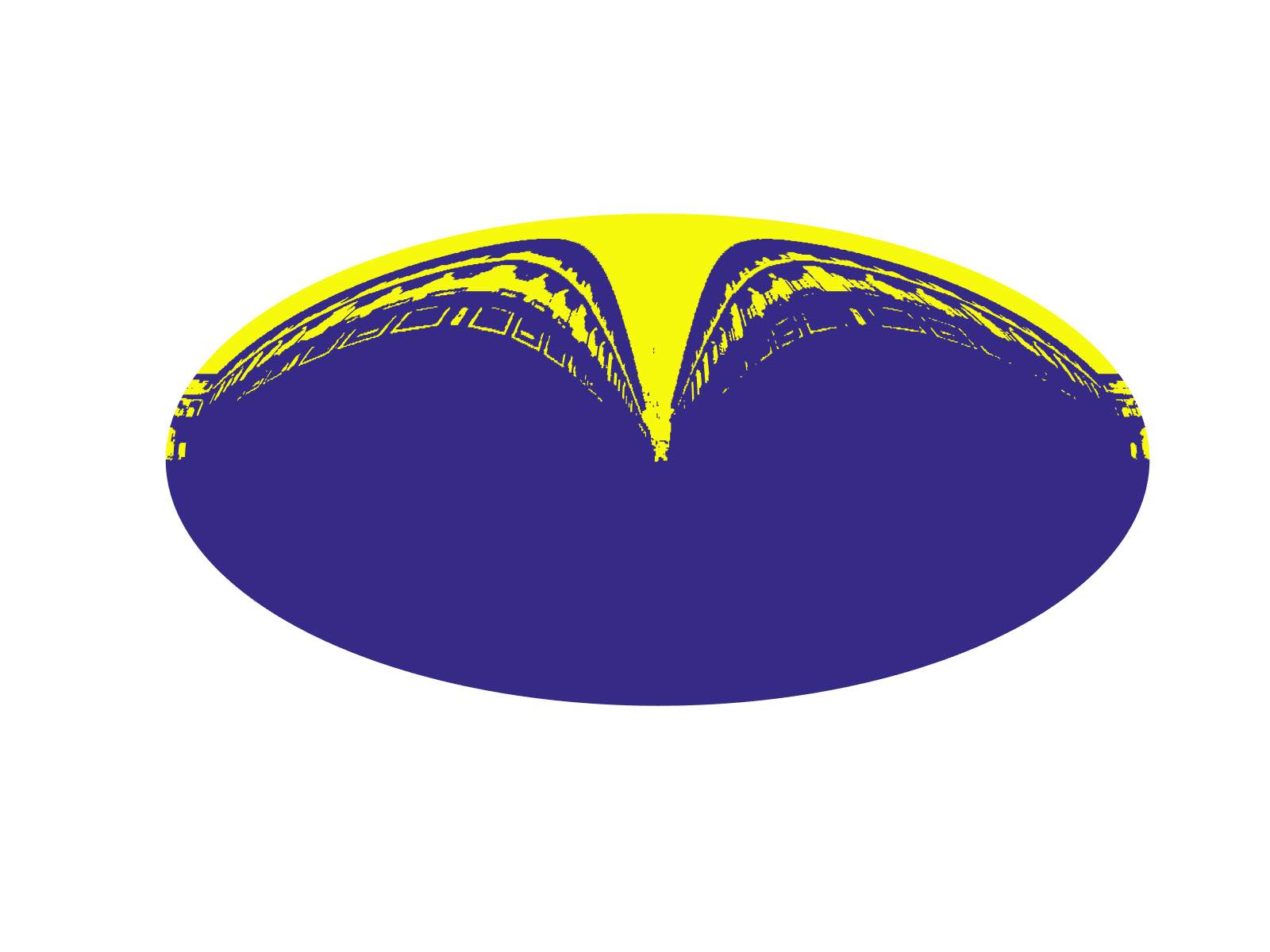}  \put(-53,35){\red{\framebox(14,10){ }}} 
                 \\
                 \includegraphics[trim={{1.5\linewidth} {1.76\linewidth} {2.2\linewidth} {0.94\linewidth}}, clip, width=0.2\linewidth]
		{./fig/light_probe_uffizi_AxisymWav_L_512_Resutls_Th.jpg}  &
                 \includegraphics[trim={{1.5\linewidth} {1.76\linewidth} {2.2\linewidth} {0.94\linewidth}}, clip, width=0.2\linewidth]
		{./fig/light_probe_uffizi_AxisymWav_L_512_Resutls.jpg}   &
                 \includegraphics[trim={{1.5\linewidth} {1.76\linewidth} {2.2\linewidth} {0.94\linewidth}}, clip, width=0.2\linewidth]
                 {./fig/light_probe_uffizi_DirectionWav_L_512_N_6_Resutls.jpg}  &
                 \includegraphics[trim={{1.5\linewidth} {1.76\linewidth} {2.2\linewidth} {0.94\linewidth}}, clip, width=0.2\linewidth]
                 {./fig/light_probe_uffizi_HybridWav_L_512_N_6_Resutls.jpg}  
                 \\
		{\small (e) K-means} & {\small (f) WSSA-A} & {\small (g) WSSA-D} & {\small (h) WSSA-H}
        \end{tabular}
	\caption{Results of light probe image - the Uffizi Gallery.  
	First row: noisy image shown on the sphere (a) and in 2D using a mollweide projection (b), and the zoomed-in red rectangle area 
	of the noisy (c) and original images (d), respectively; 
	Second to fourth rows from left to right: results of methods K-means (e), WSSA-A (f), WSSA-D (g) with $N=6$ (even $N$), 
	and WSSA-H (h), respectively.}
	\label{fig-lightprobe-uffizi}
\end{figure}

\begin{table}[h] 
\begin{center}
\caption{Light probe image - the Uffizi Gallery in Fig.\ \ref{fig-lightprobe-uffizi}: Number of unclassified points at each iteration and computation time in seconds.
	$^\ast$The fourth and fifth columns represent the results of WSSA-D with $N = 5$ and $6$, respectively.}  \label{tab:lightprobe}
 \vspace{-0.05in}
\begin{tabular}{|c||c|c|c|c|c|}
\hline 
  & {K-means} & {WSSA-A} & WSSA-D$^\ast$ & WSSA-D$^\ast$ & WSSA-H
\\ \hline \hline
\raisebox{-.15ex}{$|\bar{\mathbb{S}}^2|$} &  $523776$ & $523776$ & $523776$  & $523776$  & $523776$ 
\\ \hline
 $|\Lambda^{(0)}|$ & -  & 35242  & 36372 & 36711 & 36534
\\ \hline
$|\Lambda^{(1)}|$ &   -  & 21246 & 21176 & 21430 & 21337
\\ \hline
$|\Lambda^{(2)}|$ &  -  & 5350 & 5371 & 5456 & 5437
\\ \hline 
$|\Lambda^{(3)}|$ &  -  & 1453 & 1516 & 1491 & 1519
\\ \hline
$|\Lambda^{(4)}|$ &  -  & 402 & 434 & 404 & 431
\\ \hline
$|\Lambda^{(5)}|$ &  -  & 111 & 116 & 108 & 138
\\ \hline
$|\Lambda^{(6)}|$ &  -  & 30 & 29 & 29 & 30
\\ \hline
$|\Lambda^{(7)}|$ &  -   & 5 & 6 & 6 & 7
\\ \hline
$|\Lambda^{(8)}|$ &  -   & 0 & 0 & 0 & 0
\\ \hline \hline
Time & $<$ 1 s   & 41.9 s  & 145.7 s & 152.2 s & 702.7 s
\\ \hline
\end{tabular}
\end{center}
\end{table}

The second example is on segmenting a light probe image of a natural scene: the Uffizi Gallery in Florence\footnote{
The data were downloaded from the webpage: \url{http://www.pauldebevec.com/Probes/}.}.
A light probe image was created by taking two pictures of a mirrored ball ninety degrees apart and assembling 
the two radiance maps into a full sphere. 

Fig.\ \ref{fig-lightprobe-uffizi} shows the results of segmenting the light probe image of the Uffizi Gallery, with $\epsilon= 0.05$ used in the WSSA method.
The same conclusions as those of the Earth map segmentation are obtained. Nonetheless, specific to the example here, 
in separating the sky and the bright parts of the windows within the test data 
the WSSA method is better at detecting detailed structures than the K-means method, for example, in the window frames shown in the zoomed-in figures in 
Fig.\ \ref{fig-lightprobe-uffizi} (e)--(h). Again, the WSSA-D and WSSA-H methods are slightly better 
than the WSSA-A method which uses axisymmetric wavelets,
but WSSA-A is faster (see Table \ref{tab:lightprobe}).

\begin{figure}
	\centering
	\begin{tabular}{cccc}
	        \multicolumn{4}{c}{\bf Test data} \vspace{-0.1in}
	        \\
	        	\includegraphics[trim={{.9\linewidth} {.32\linewidth} {.8\linewidth} {.6\linewidth}}, clip, width=0.2\linewidth]
		{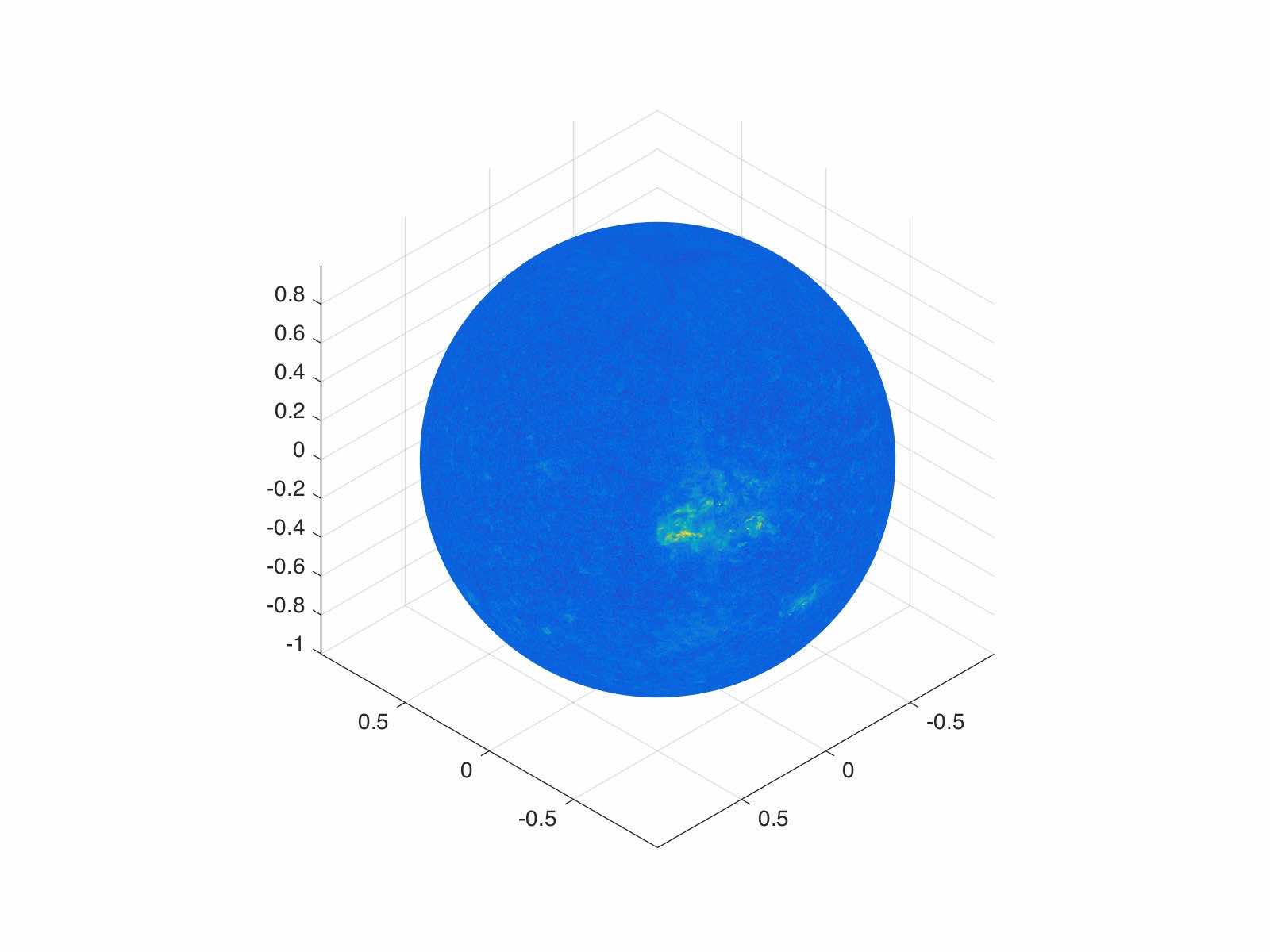} &
		\includegraphics[trim={{.18\linewidth} {.12\linewidth} {.09\linewidth} {.1\linewidth}}, clip, width=0.2\linewidth]
		{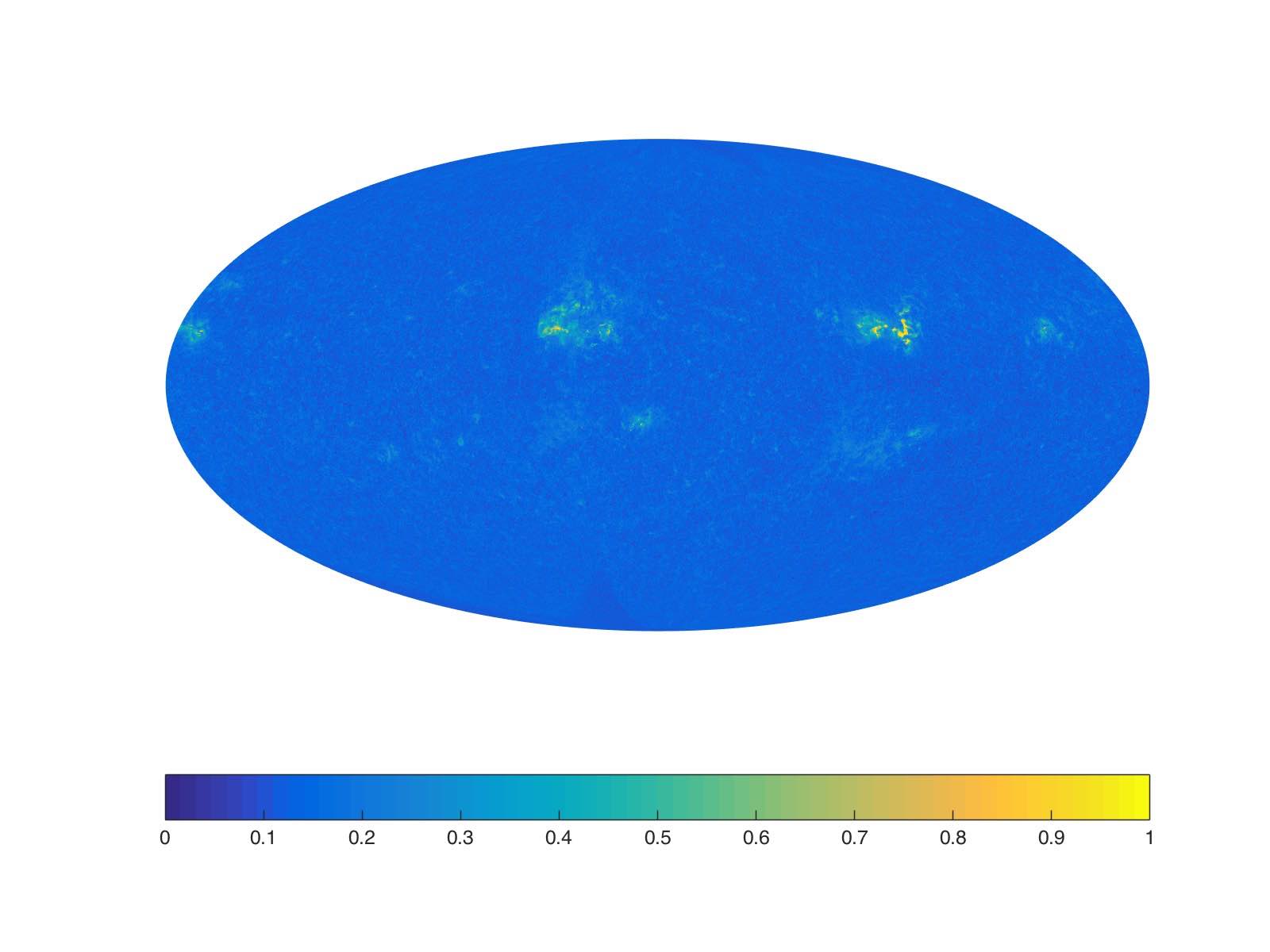}  \put(-27,32){\red{\framebox(10,10){ }}} & 
		\includegraphics[trim={{2.65\linewidth} {1.9\linewidth} {1.05\linewidth} {0.85\linewidth}}, clip, width=0.2\linewidth]
		{./fig/solar_map_AxisymWav_L_512_Noisy_image.jpg} &
		\includegraphics[trim={{2.65\linewidth} {1.9\linewidth} {1.05\linewidth} {0.85\linewidth}}, clip, width=0.2\linewidth]
		{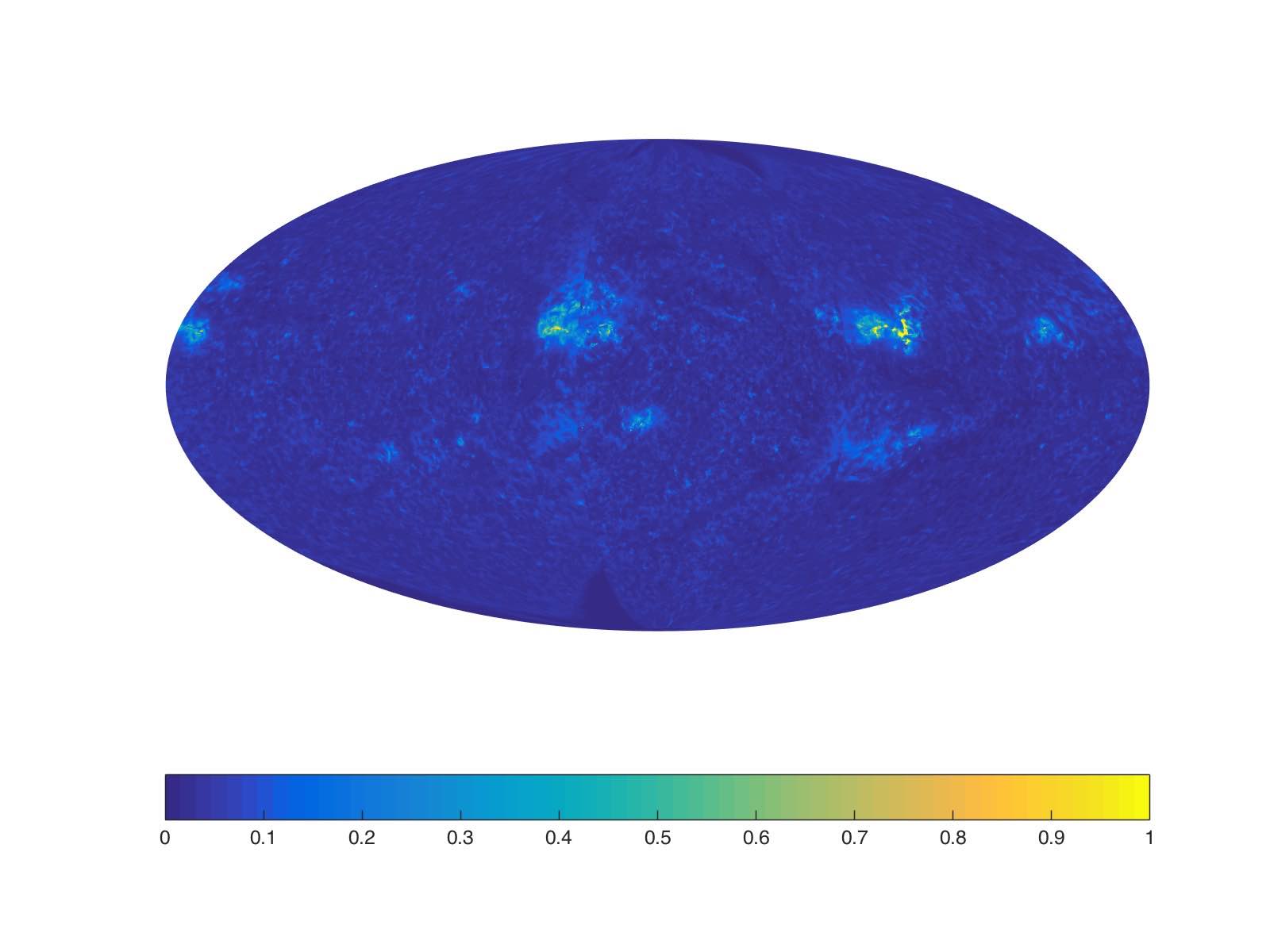}
		\\
		{\small (a) noisy image} & {\small (b) noisy image} & {\small (c) noisy image} & {\small (d) original image}  \vspace{0.15in}
		\\
		 \multicolumn{4}{c}{\bf Segmentation results}  \vspace{-0.02in}
	        \\
       		\includegraphics[trim={{.9\linewidth} {.32\linewidth} {.8\linewidth} {.6\linewidth}}, clip, width=0.2\linewidth]
		{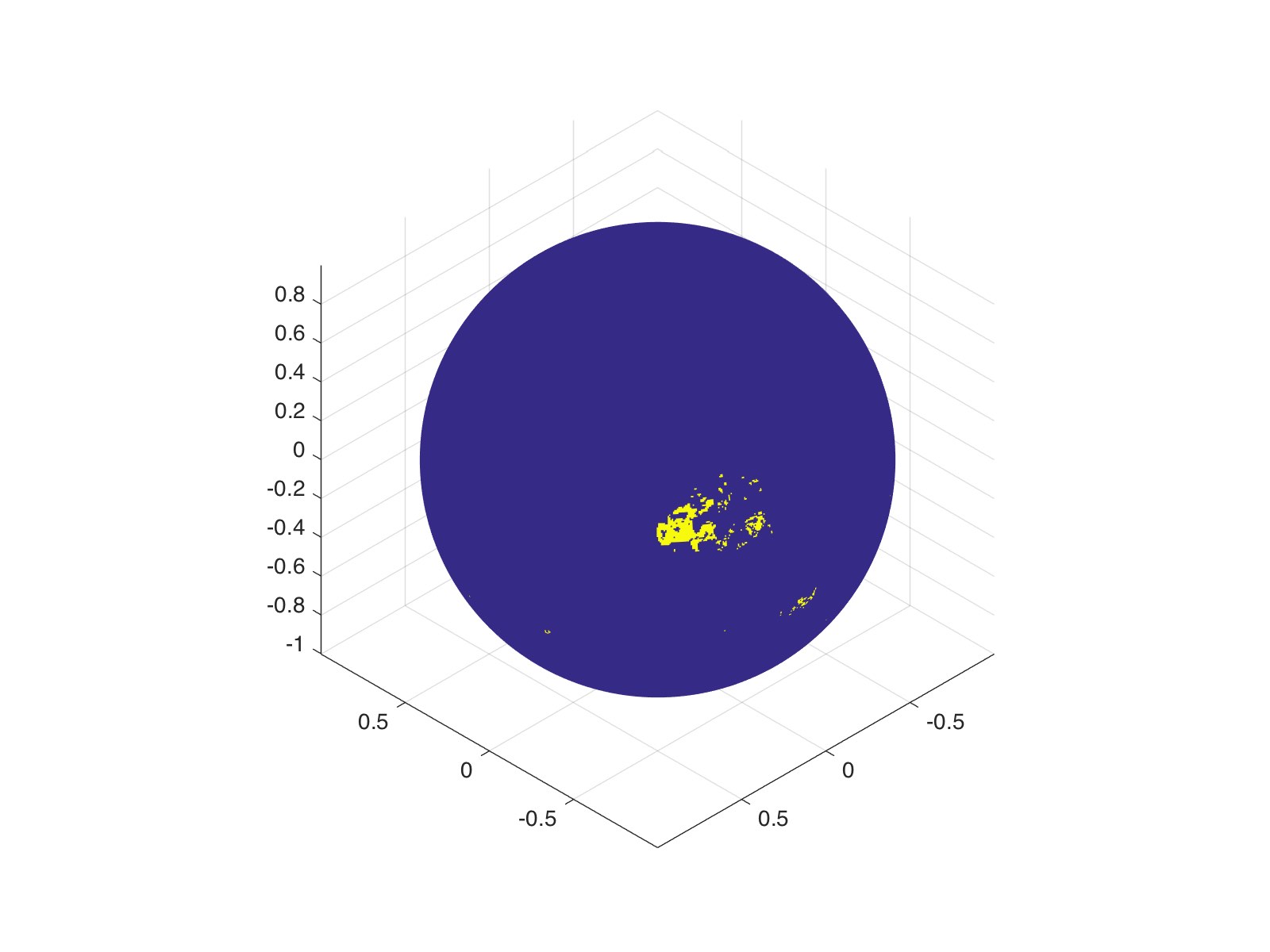} &
        		\includegraphics[trim={{.9\linewidth} {.32\linewidth} {.8\linewidth} {.6\linewidth}}, clip, width=0.2\linewidth]
		{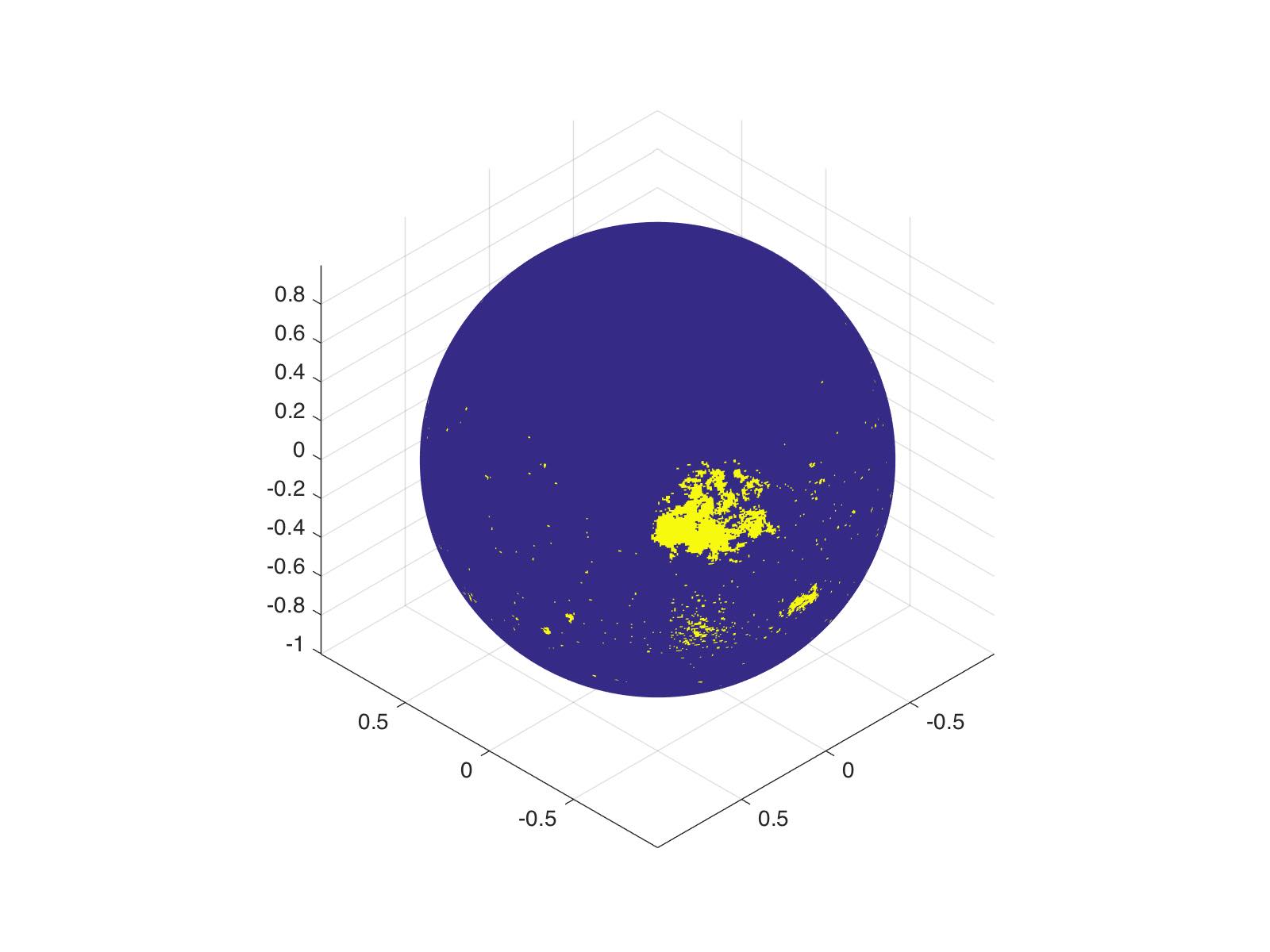} &
        		\includegraphics[trim={{.9\linewidth} {.32\linewidth} {.8\linewidth} {.6\linewidth}}, clip, width=0.2\linewidth]
		{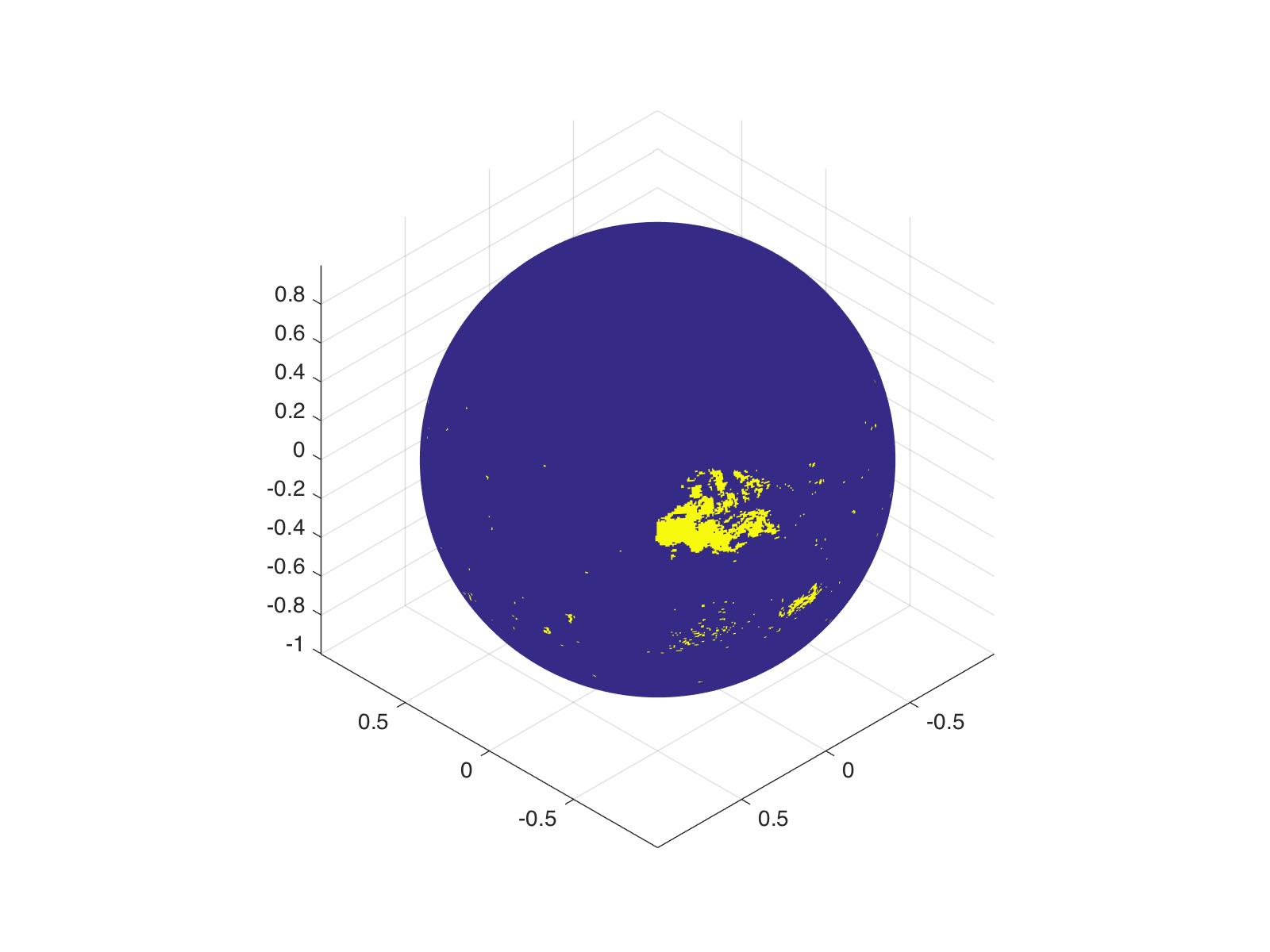} &
        		\includegraphics[trim={{.9\linewidth} {.32\linewidth} {.8\linewidth} {.6\linewidth}}, clip, width=0.2\linewidth]
		{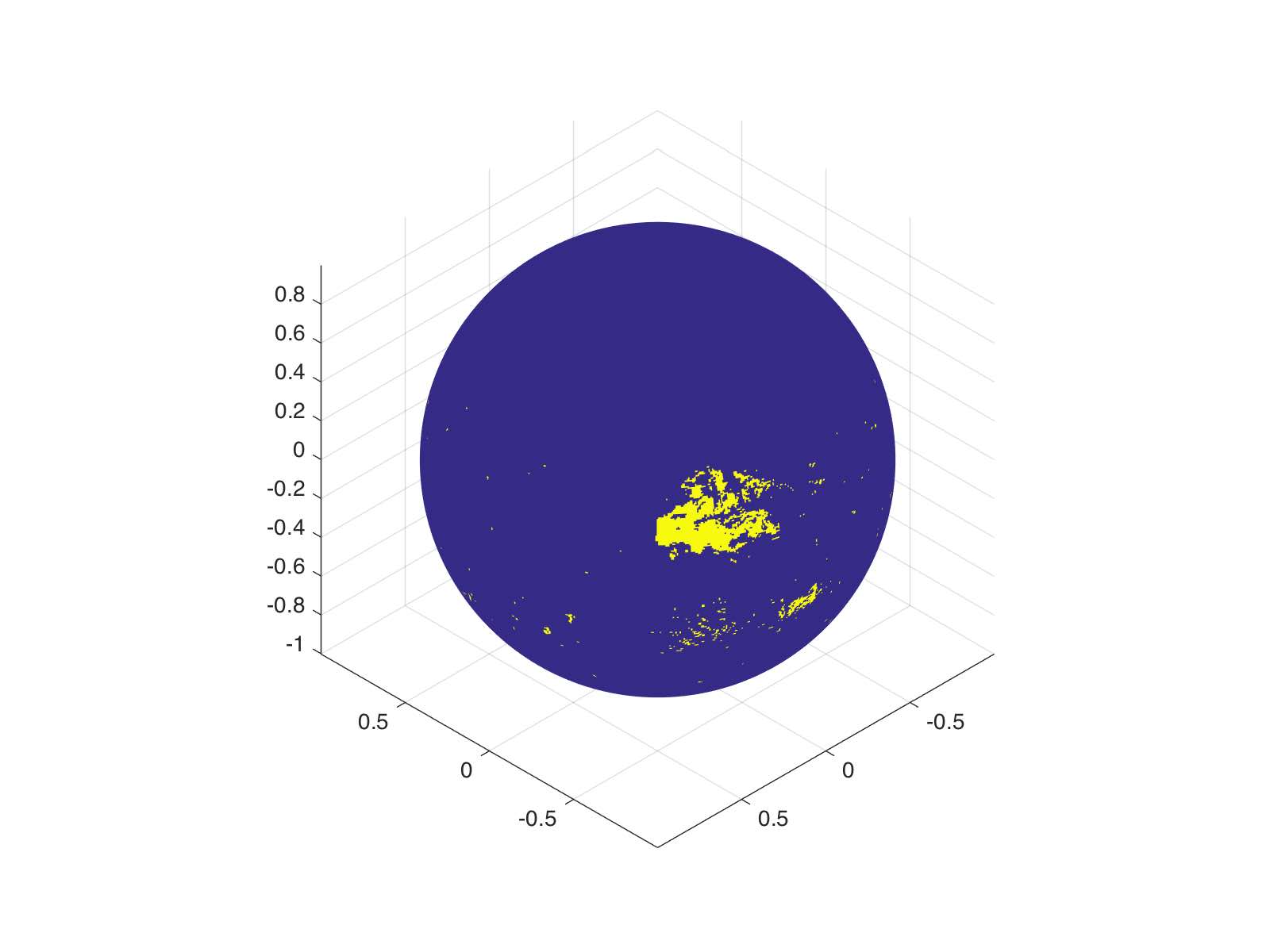} 
		\\
		\includegraphics[trim={{.5\linewidth} {.2\linewidth} {.3\linewidth} {.2\linewidth}}, clip, width=0.2\linewidth]
		{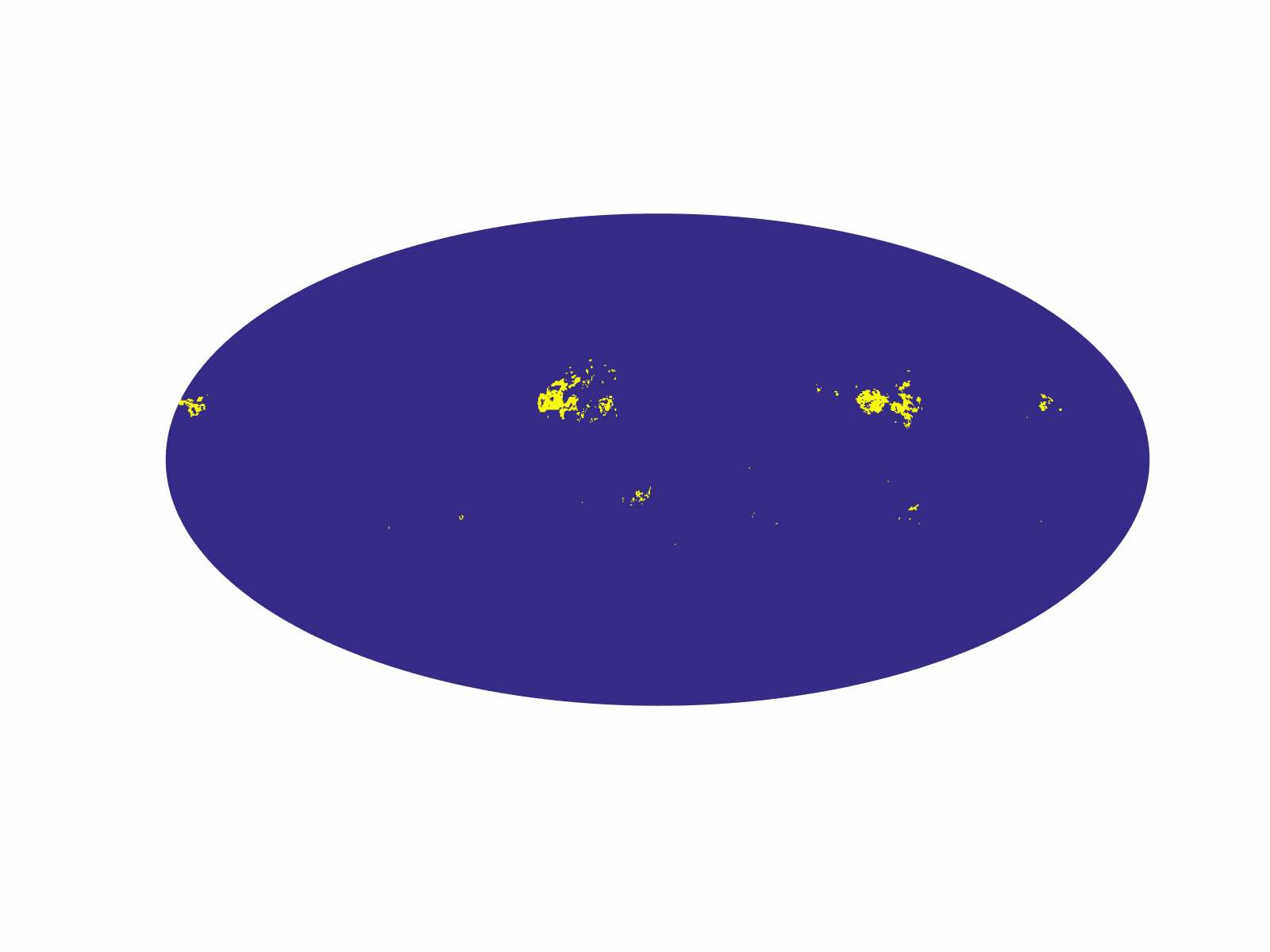}   \put(-25,30){\red{\framebox(10,10){ }}} &
        		\includegraphics[trim={{.5\linewidth} {.2\linewidth} {.3\linewidth} {.2\linewidth}}, clip, width=0.2\linewidth]
		{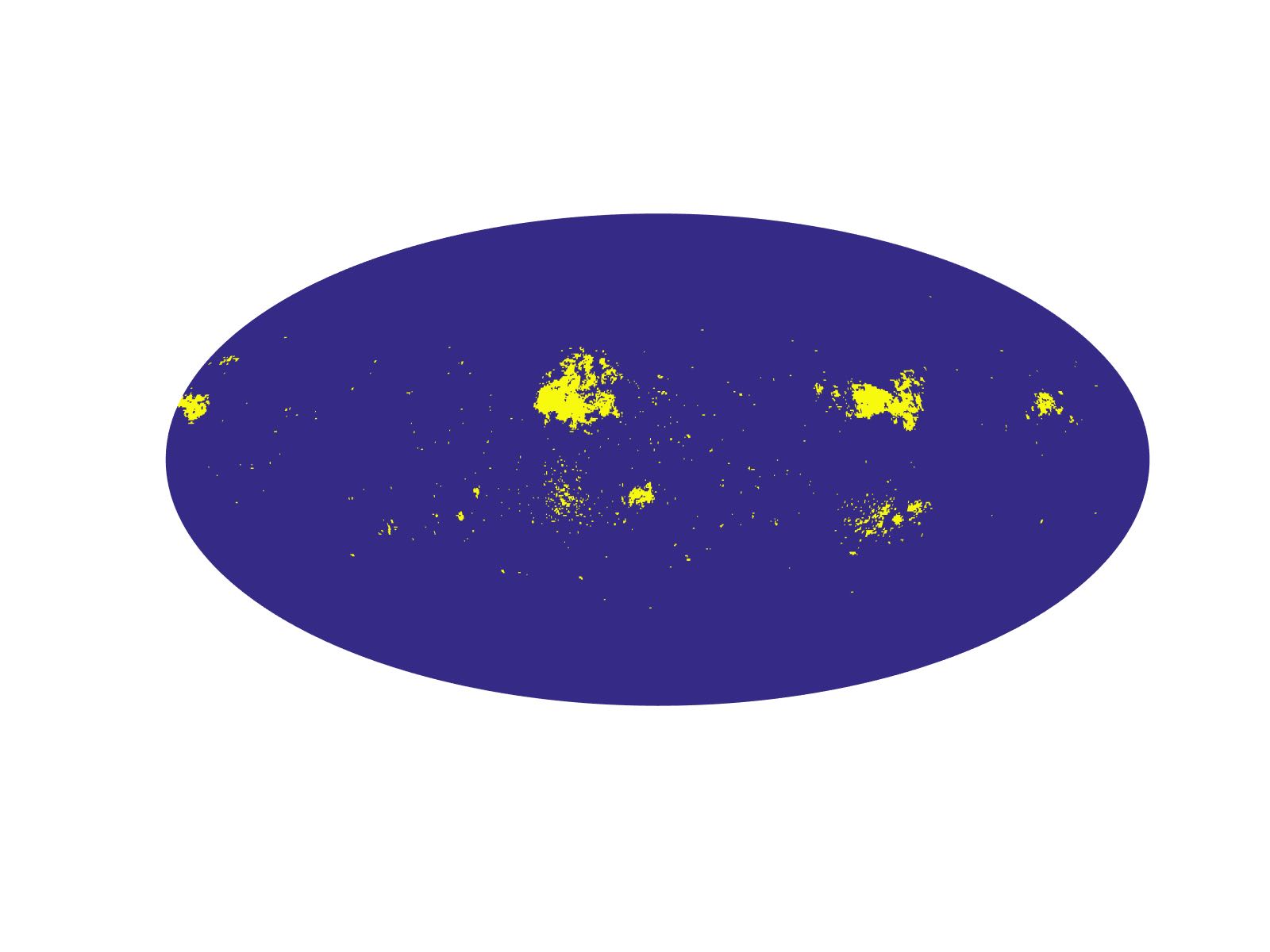}  \put(-25,30){\red{\framebox(10,10){ }}}  &
                 \includegraphics[trim={{.5\linewidth} {.2\linewidth} {.3\linewidth} {.2\linewidth}}, clip, width=0.2\linewidth]
                 {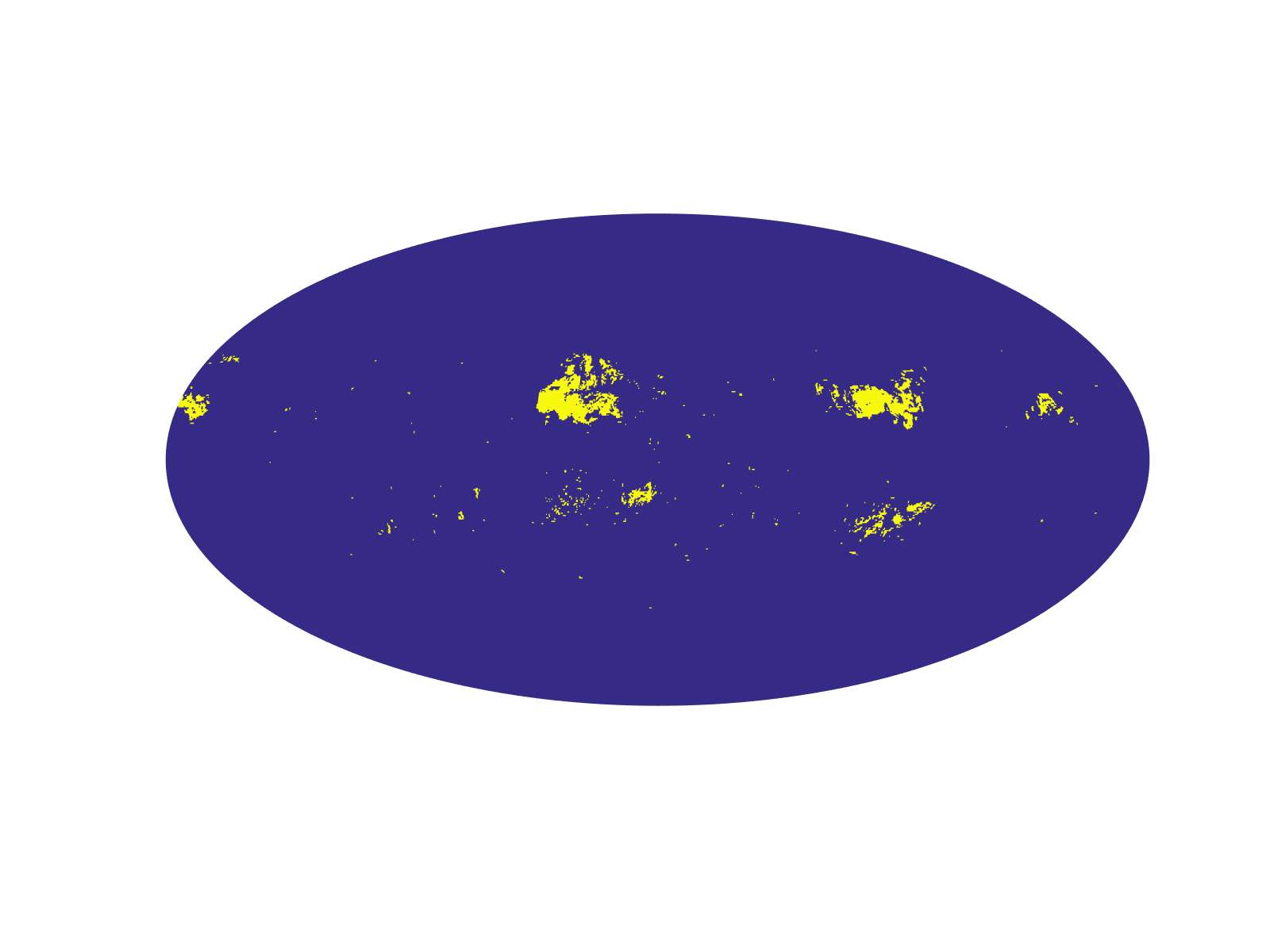} \put(-25,30){\red{\framebox(10,10){ }}}  &  
                 \includegraphics[trim={{.5\linewidth} {.2\linewidth} {.3\linewidth} {.2\linewidth}}, clip, width=0.2\linewidth]
                 {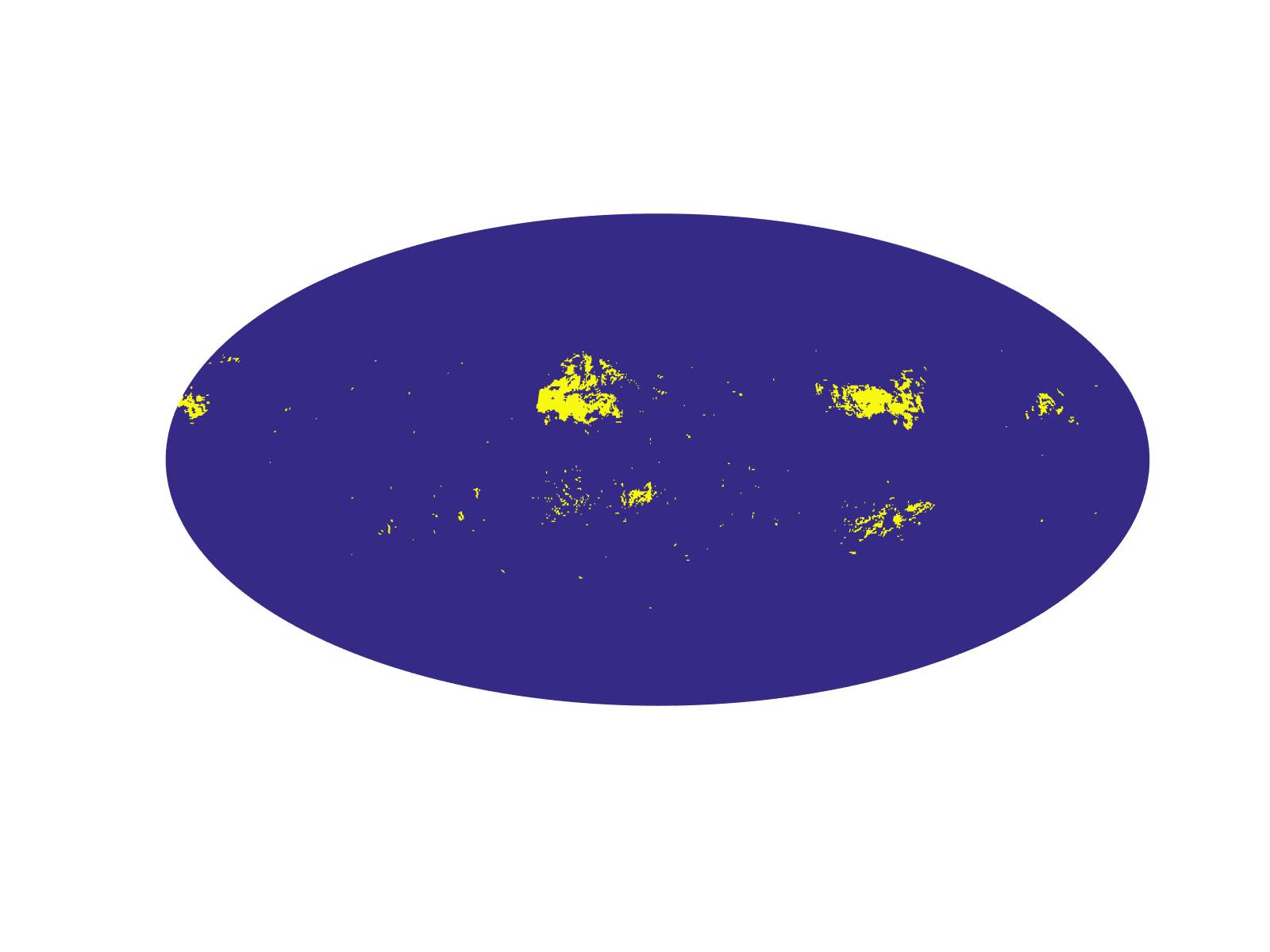} \put(-25,30){\red{\framebox(10,10){ }}}   
                 \\
                 \includegraphics[trim={{2.6\linewidth} {1.6\linewidth} {1.0\linewidth} {1.05\linewidth}}, clip, width=0.2\linewidth]
		{./fig/solar_map_AxisymWav_L_512_Resutls_Th.jpg}  &
                 \includegraphics[trim={{2.6\linewidth} {1.6\linewidth} {1.0\linewidth} {1.05\linewidth}}, clip, width=0.2\linewidth]
		{./fig/solar_map_AxisymWav_L_512_Resutls.jpg}   &
                 \includegraphics[trim={{2.6\linewidth} {1.6\linewidth} {1.0\linewidth} {1.05\linewidth}}, clip, width=0.2\linewidth]
                 {./fig/solar_map_DirectionWav_L_512_N_5_Resutls.jpg} &  
                 \includegraphics[trim={{2.6\linewidth} {1.6\linewidth} {1.0\linewidth} {1.05\linewidth}}, clip, width=0.2\linewidth]
                 {./fig/solar_map_HybridWav_L_512_N_6_Resutls.jpg}  
		\\
		{\small (e) K-means} & {\small (f) WSSA-A} & {\small (g) WSSA-D} & {\small (h) WSSA-H}
        \end{tabular}
	\caption{Results of solar map.  
	First row: noisy image shown on the sphere (a) and in 2D using a mollweide projection (b), and the zoomed-in red rectangle area 
	of the noisy (c) and original images (d), respectively; 
	Second to fourth rows from left to right: results of methods K-means (e), WSSA-A (f), WSSA-D (g) with $N=5$ (odd $N$), 
	and WSSA-H (h), respectively.}
	\label{fig-solarmap}
\end{figure}

\begin{figure}
	\centering
	\begin{tabular}{cccc}
	        \multicolumn{4}{c}{\bf Test data} \vspace{-0.15in}
	        \\
		\includegraphics[trim={{.9\linewidth} {.32\linewidth} {.8\linewidth} {.6\linewidth}}, clip, width=0.2\linewidth]
		{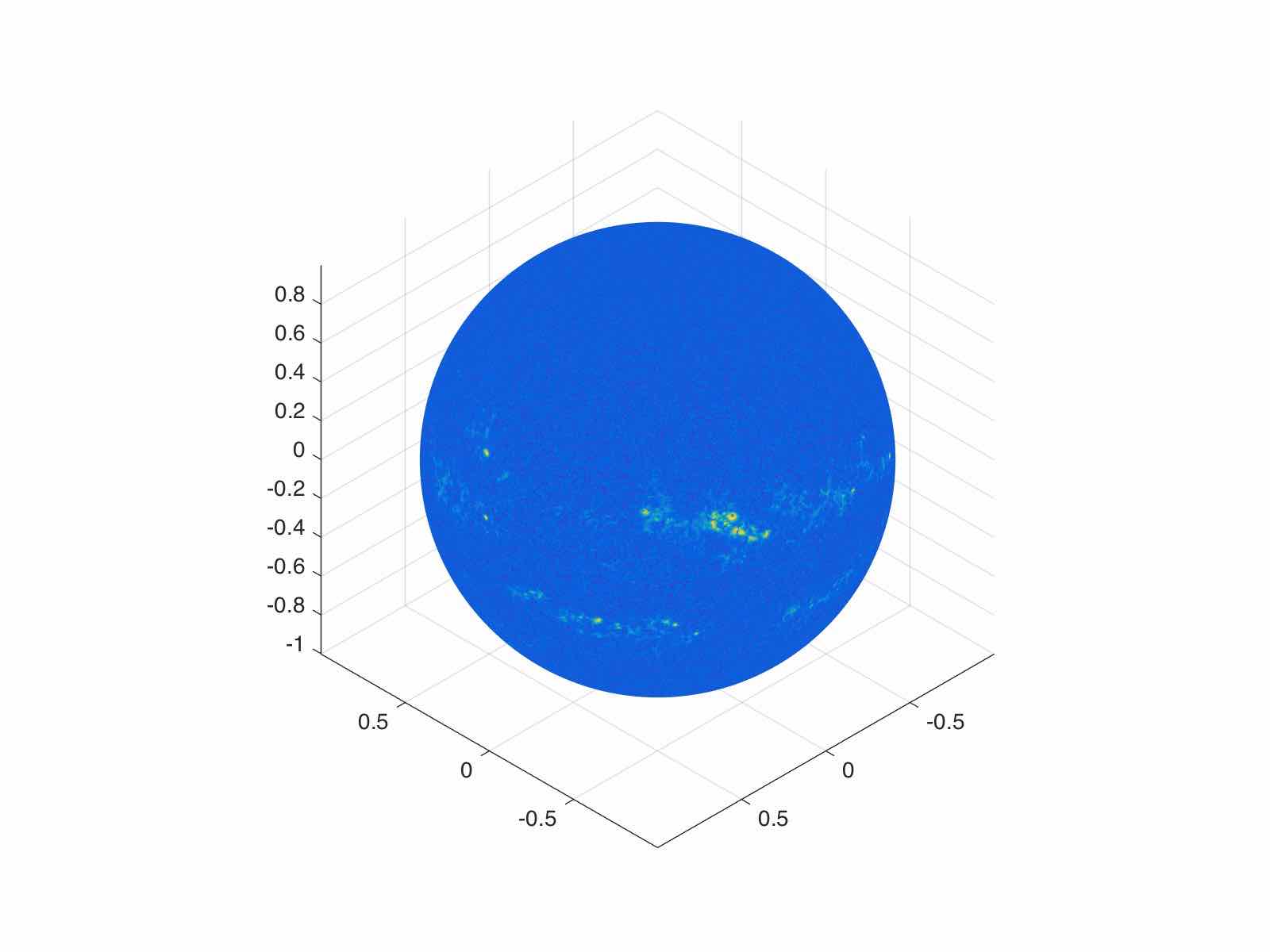} &
		\includegraphics[trim={{.18\linewidth} {.12\linewidth} {.09\linewidth} {.1\linewidth}}, clip, width=0.2\linewidth]
		{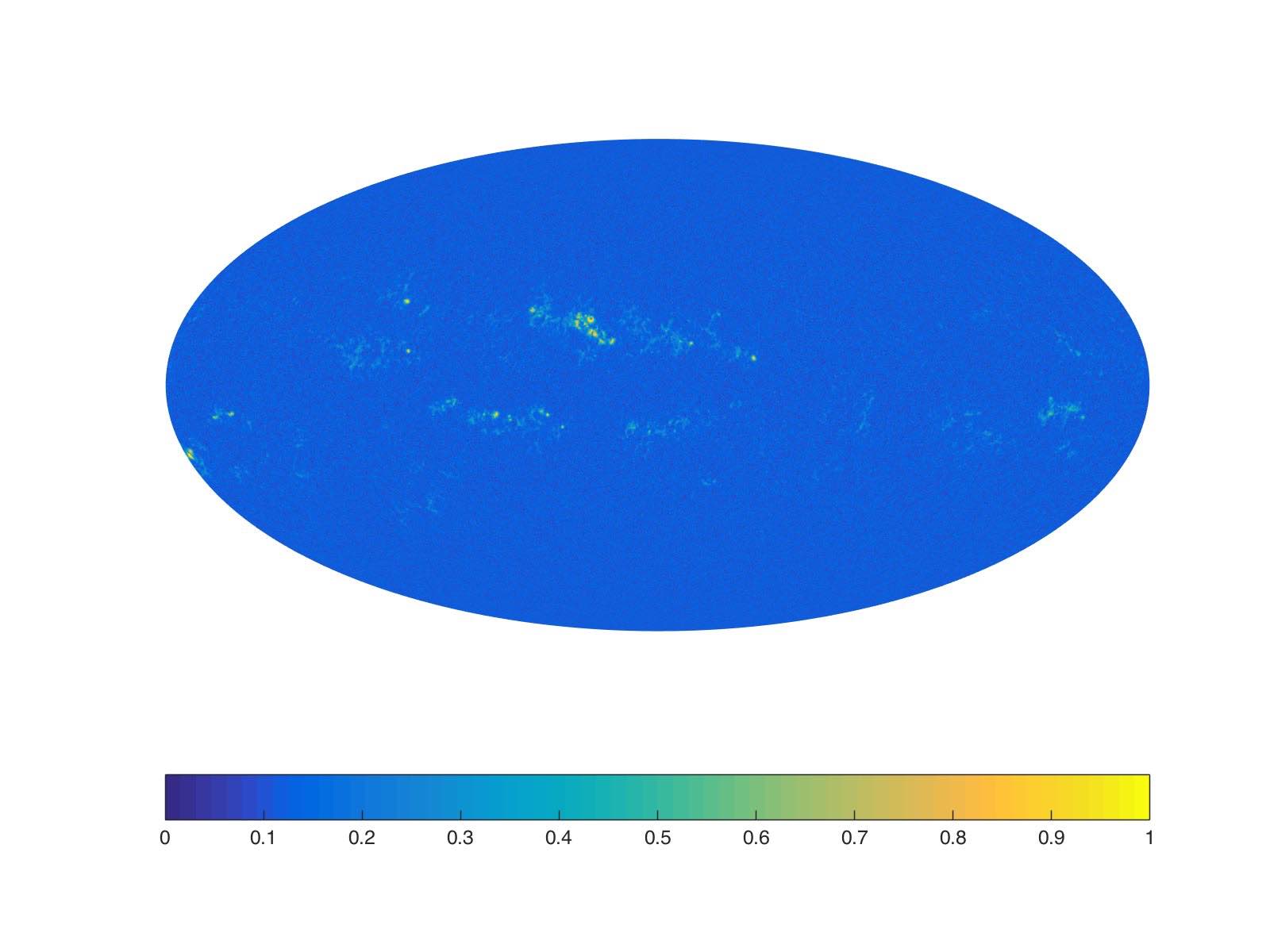} \put(-55,26){\red{\framebox(27,18){ }}}  &
		\includegraphics[trim={{1.55\linewidth} {1.5\linewidth} {1.65\linewidth} {0.95\linewidth}}, clip, width=0.2\linewidth]
		{./fig/solar_map_031301_AxisymWav_L_512_Noisy_image.jpg} & 
		\includegraphics[trim={{1.55\linewidth} {1.5\linewidth} {1.65\linewidth} {0.95\linewidth}}, clip, width=0.2\linewidth]
		{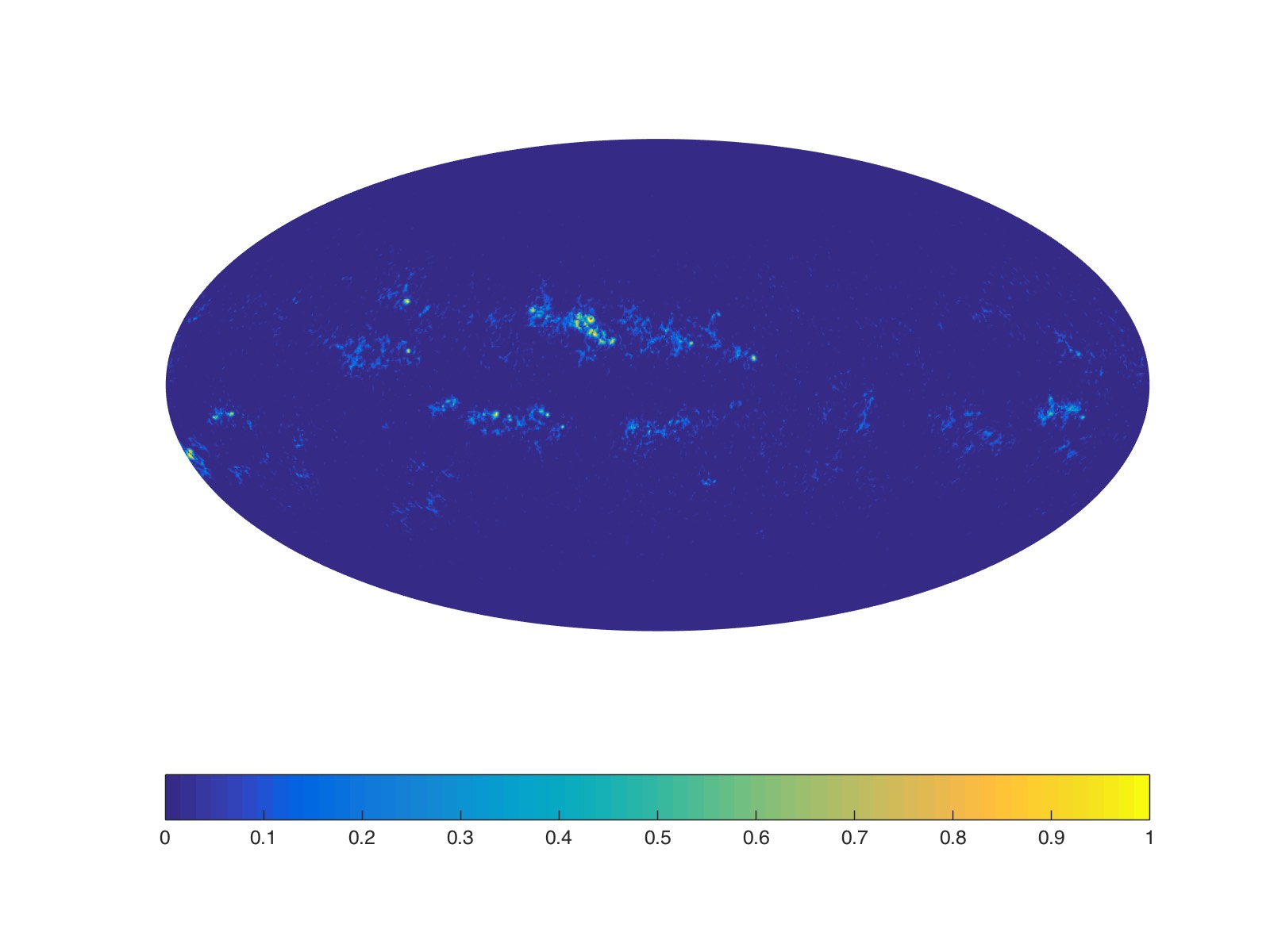}
		\\
		{\small (a) noisy image} & {\small (b) noisy image} & {\small (c) noisy image} & {\small (d) original image}  \vspace{0.15in}
		\\
		 \multicolumn{4}{c}{\bf Segmentation results}  \vspace{-0.02in}
	        \\
	        \includegraphics[trim={{.9\linewidth} {.32\linewidth} {.8\linewidth} {.6\linewidth}}, clip, width=0.2\linewidth]
		{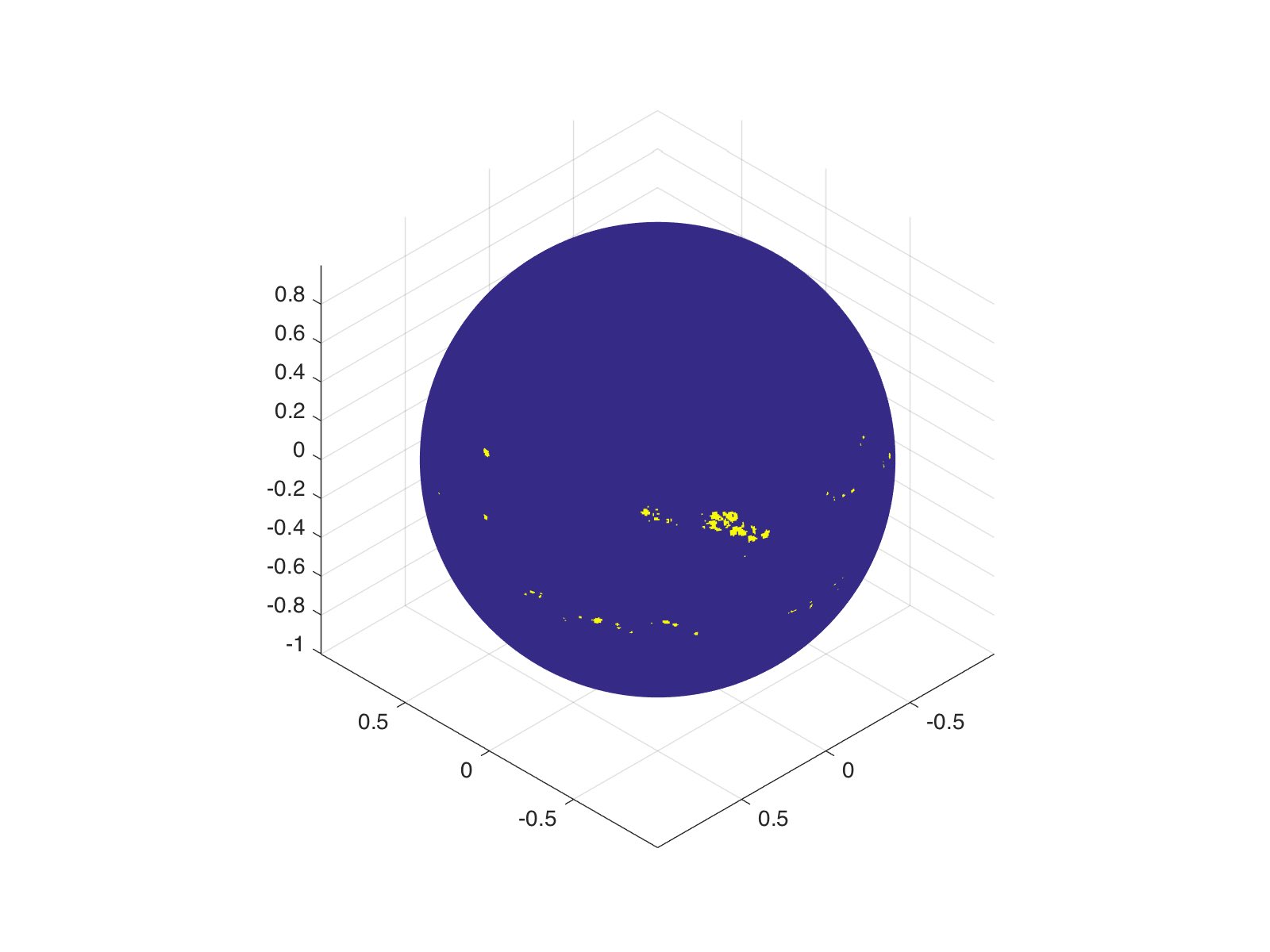} &
        		\includegraphics[trim={{.9\linewidth} {.32\linewidth} {.8\linewidth} {.6\linewidth}}, clip, width=0.2\linewidth]
		{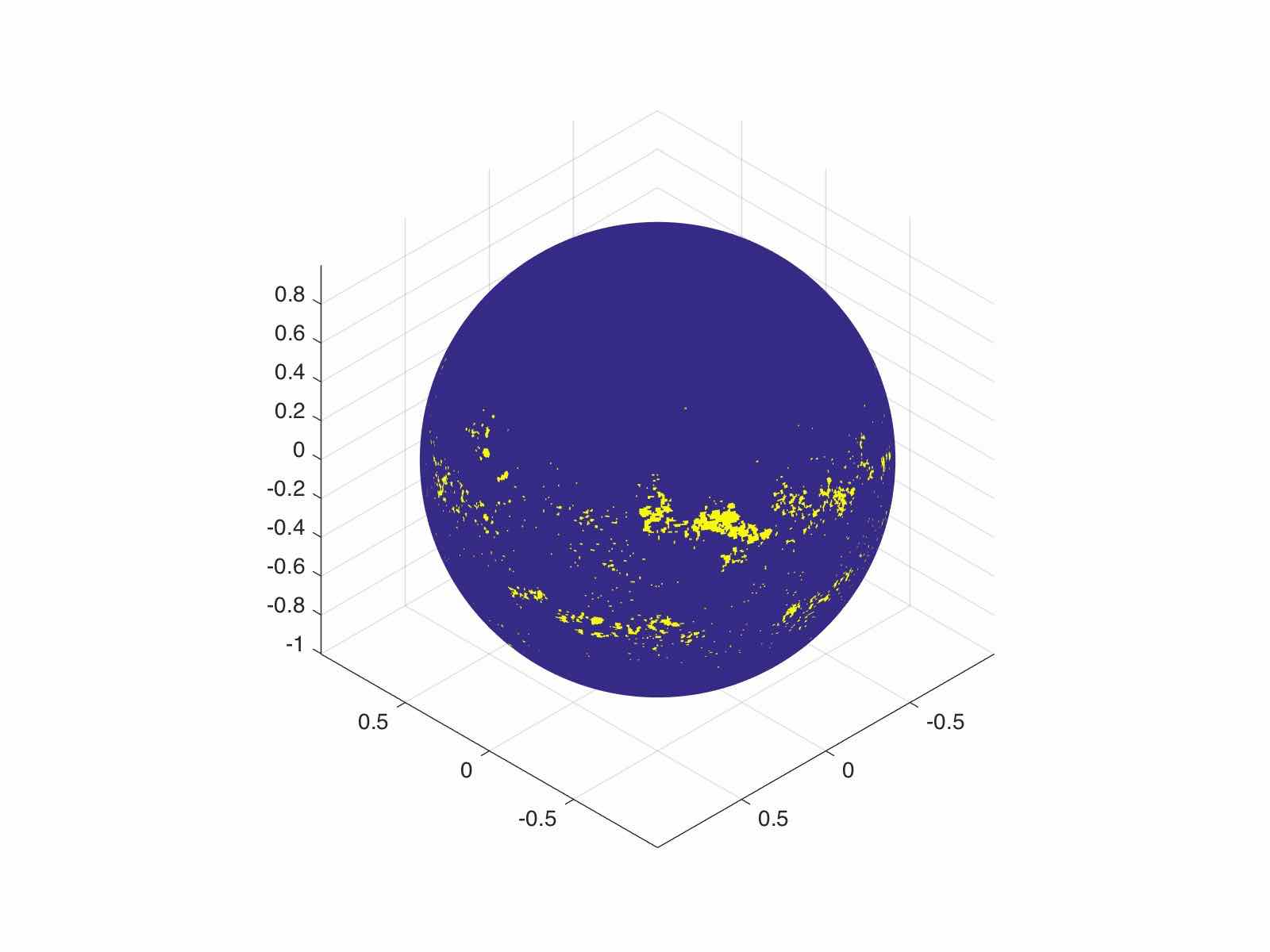} &
        		\includegraphics[trim={{.9\linewidth} {.32\linewidth} {.8\linewidth} {.6\linewidth}}, clip, width=0.2\linewidth]
		{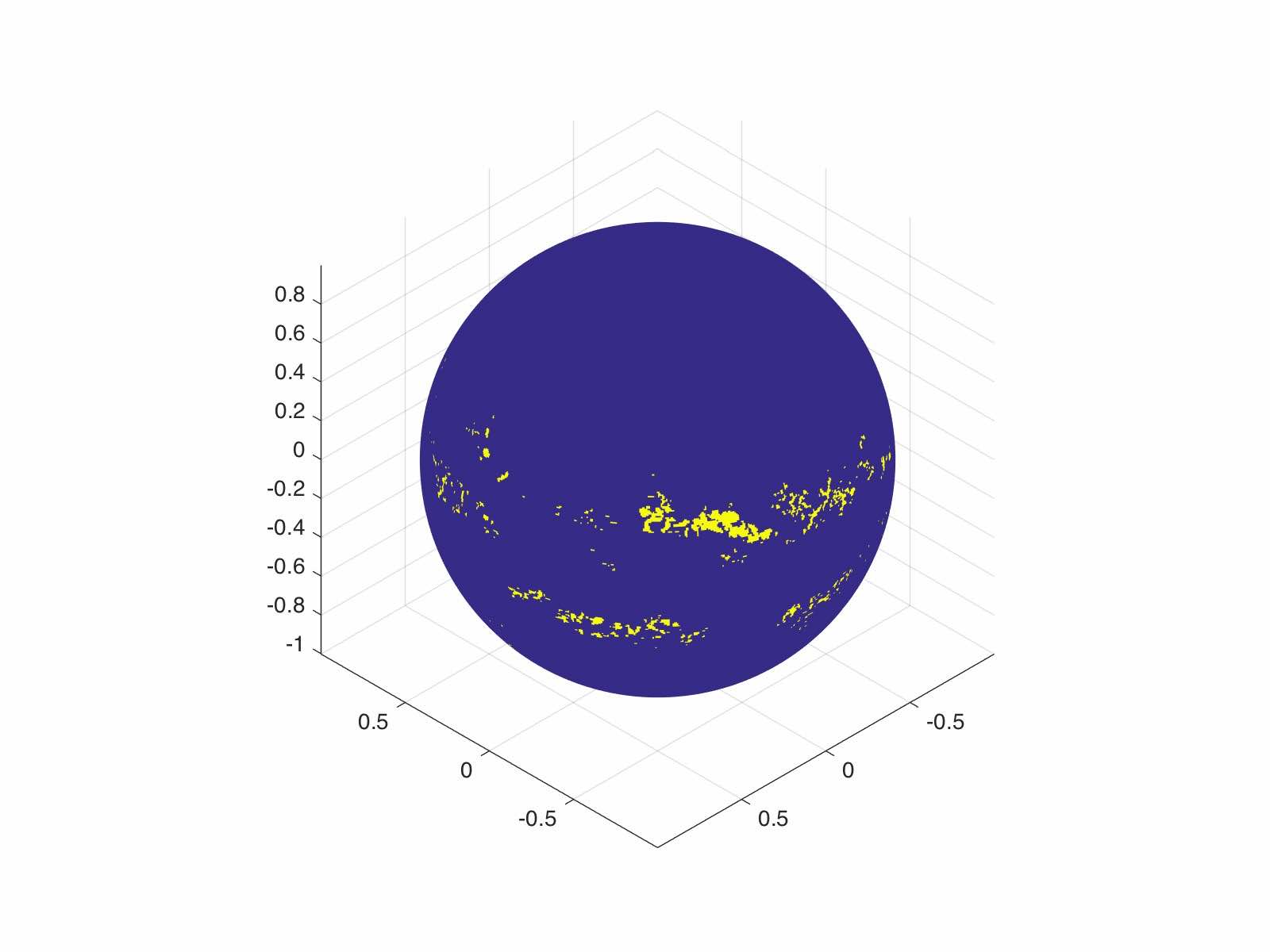} &
        		\includegraphics[trim={{.9\linewidth} {.32\linewidth} {.8\linewidth} {.6\linewidth}}, clip, width=0.2\linewidth]
		{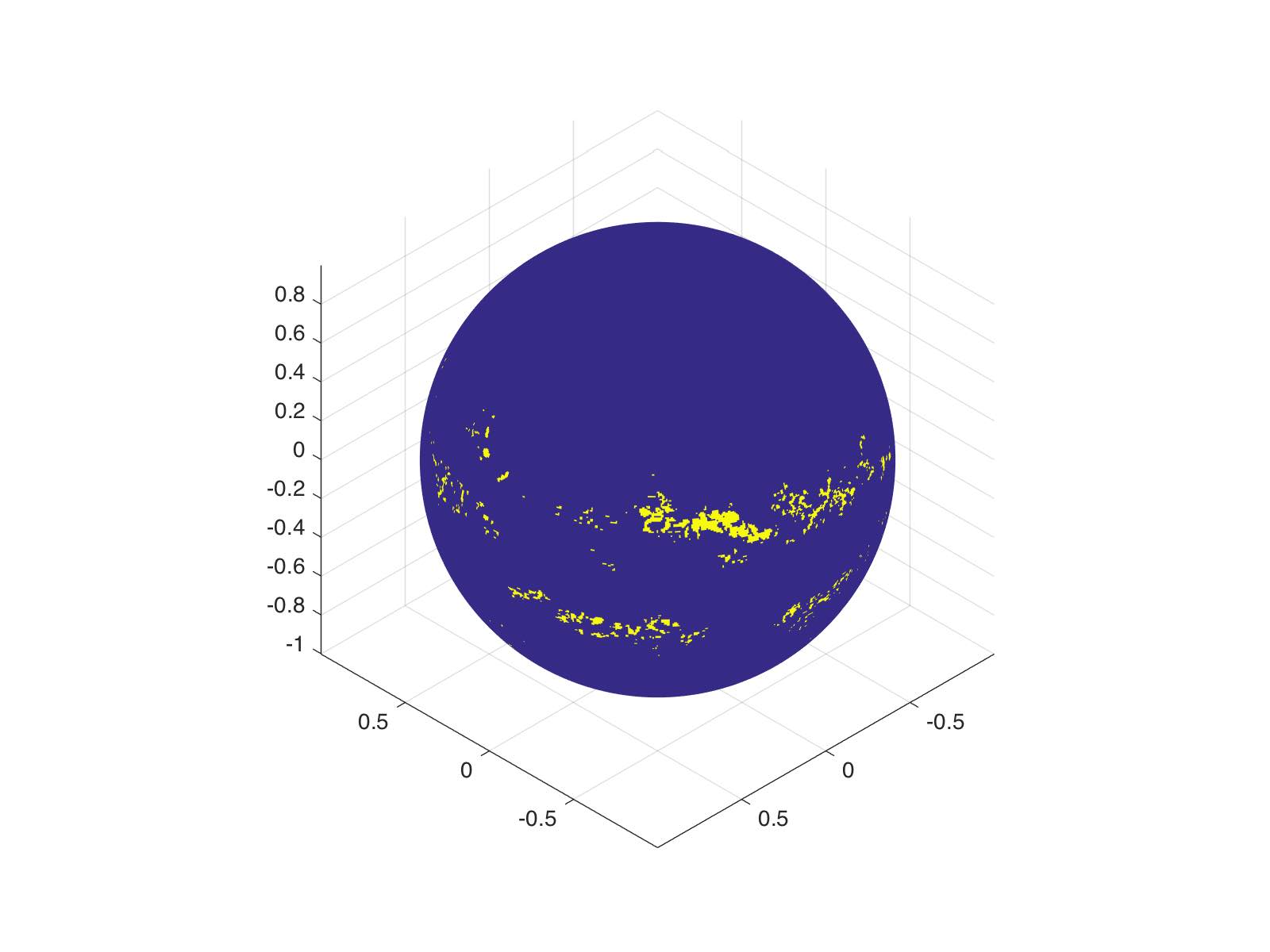} 
		\\
		\includegraphics[trim={{.5\linewidth} {.2\linewidth} {.3\linewidth} {.2\linewidth}}, clip, width=0.2\linewidth]
		{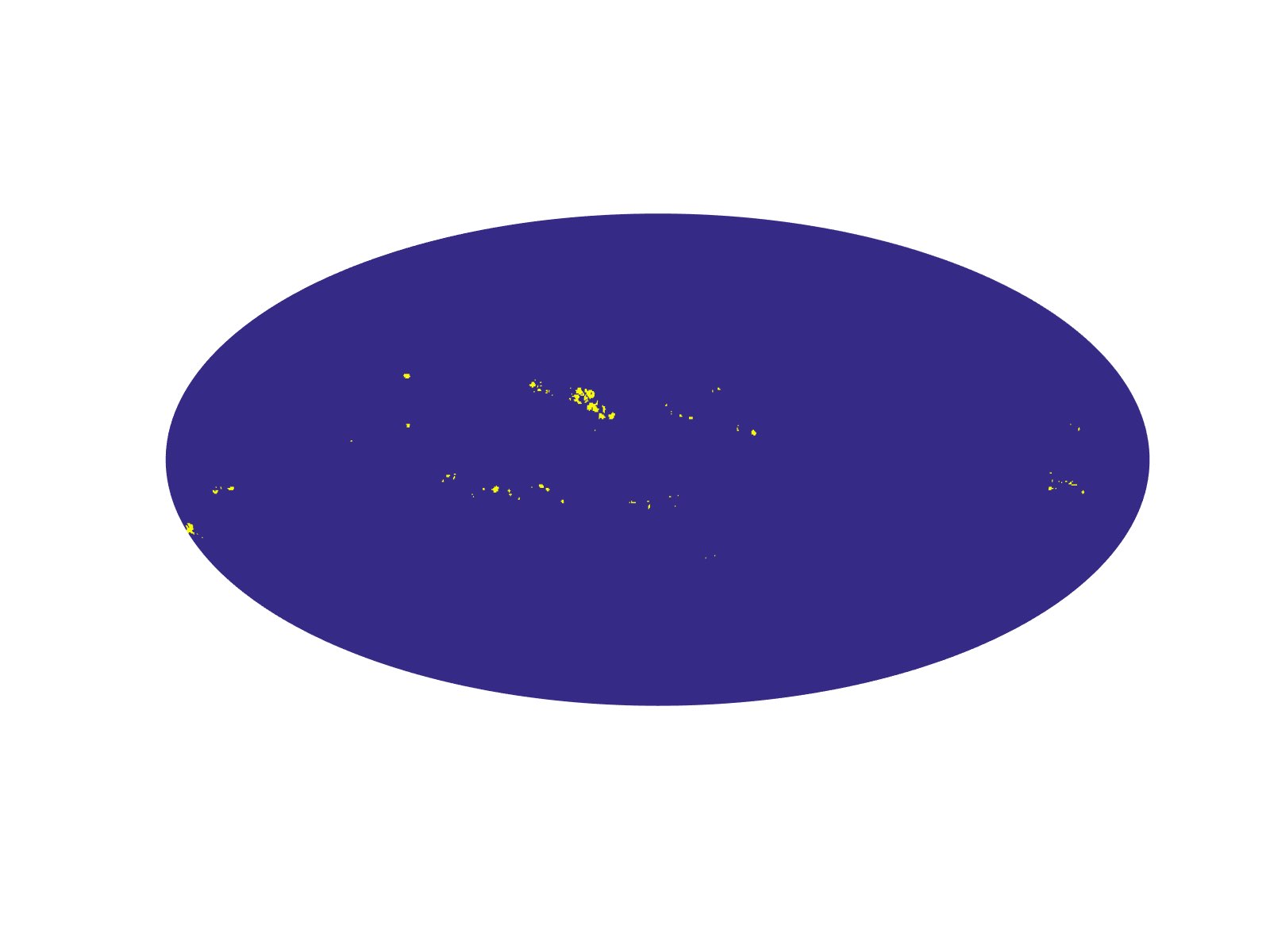}  \put(-55,23){\red{\framebox(27,18){ }}}   &
        		\includegraphics[trim={{.5\linewidth} {.2\linewidth} {.3\linewidth} {.2\linewidth}}, clip, width=0.2\linewidth]
		{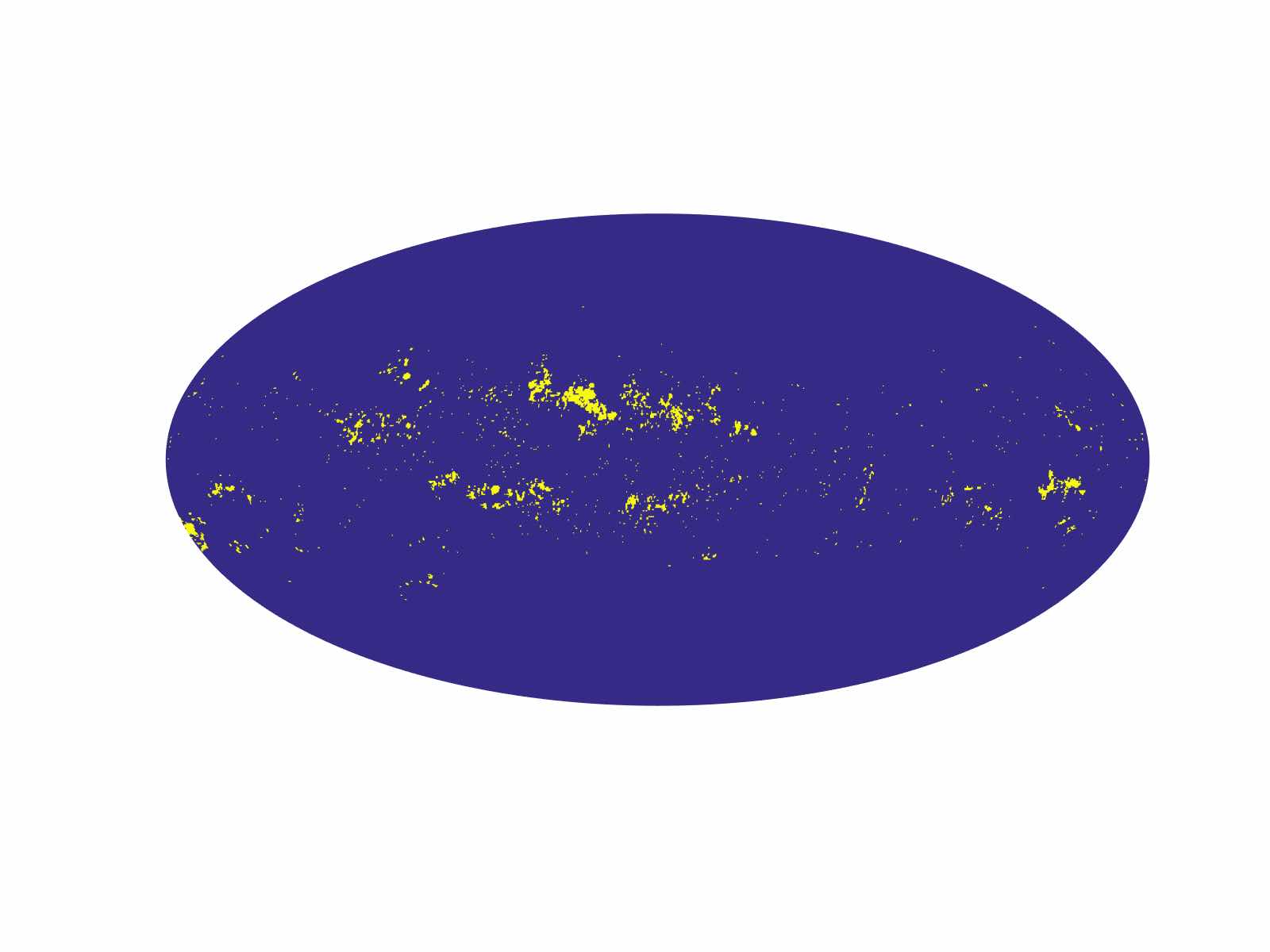}  \put(-55,23){\red{\framebox(27,18){ }}}  &
                 \includegraphics[trim={{.5\linewidth} {.2\linewidth} {.3\linewidth} {.2\linewidth}}, clip, width=0.2\linewidth]
                 {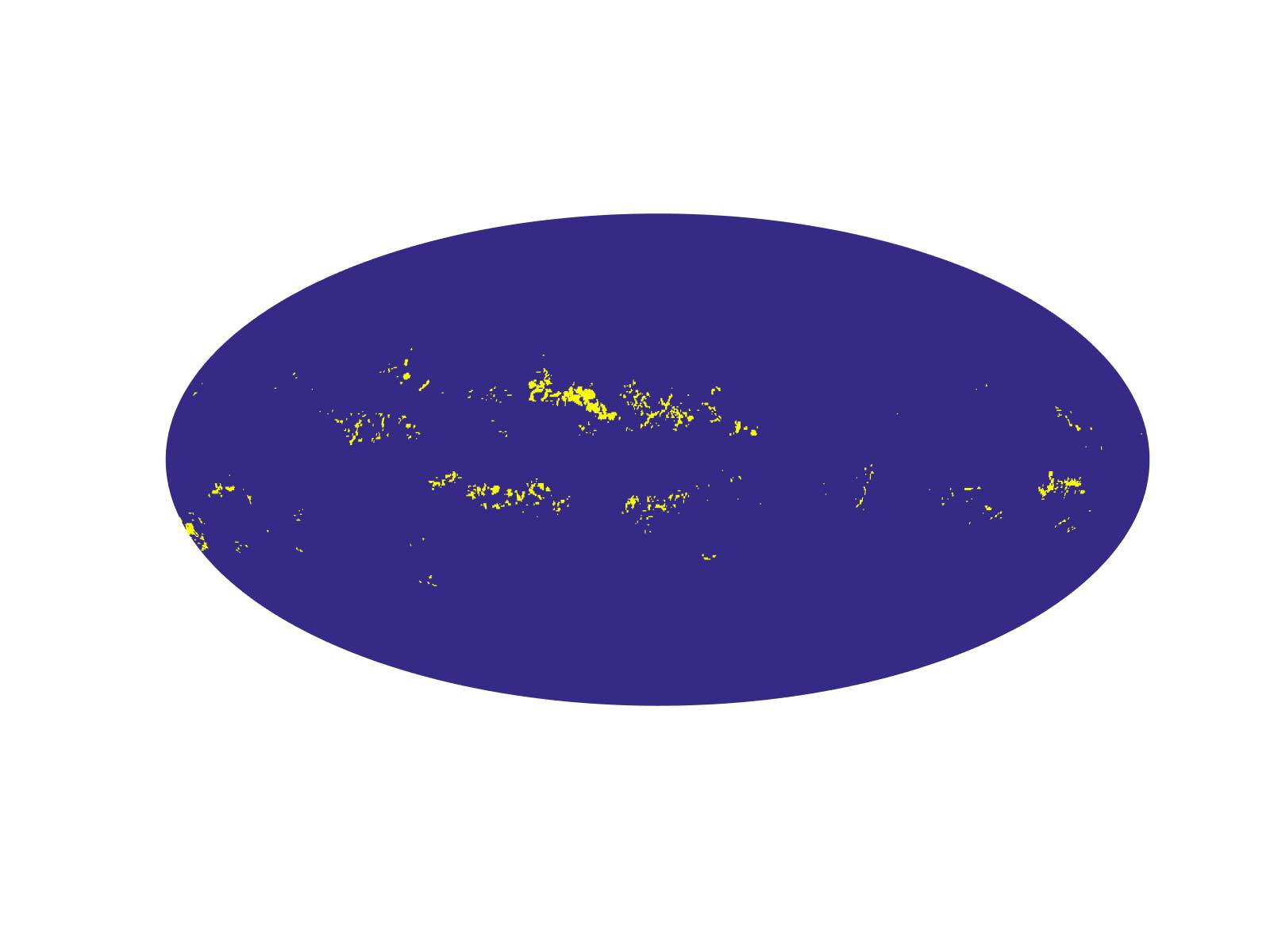} \put(-55,23){\red{\framebox(27,18){ }}}  &  
                 \includegraphics[trim={{.5\linewidth} {.2\linewidth} {.3\linewidth} {.2\linewidth}}, clip, width=0.2\linewidth]
                 {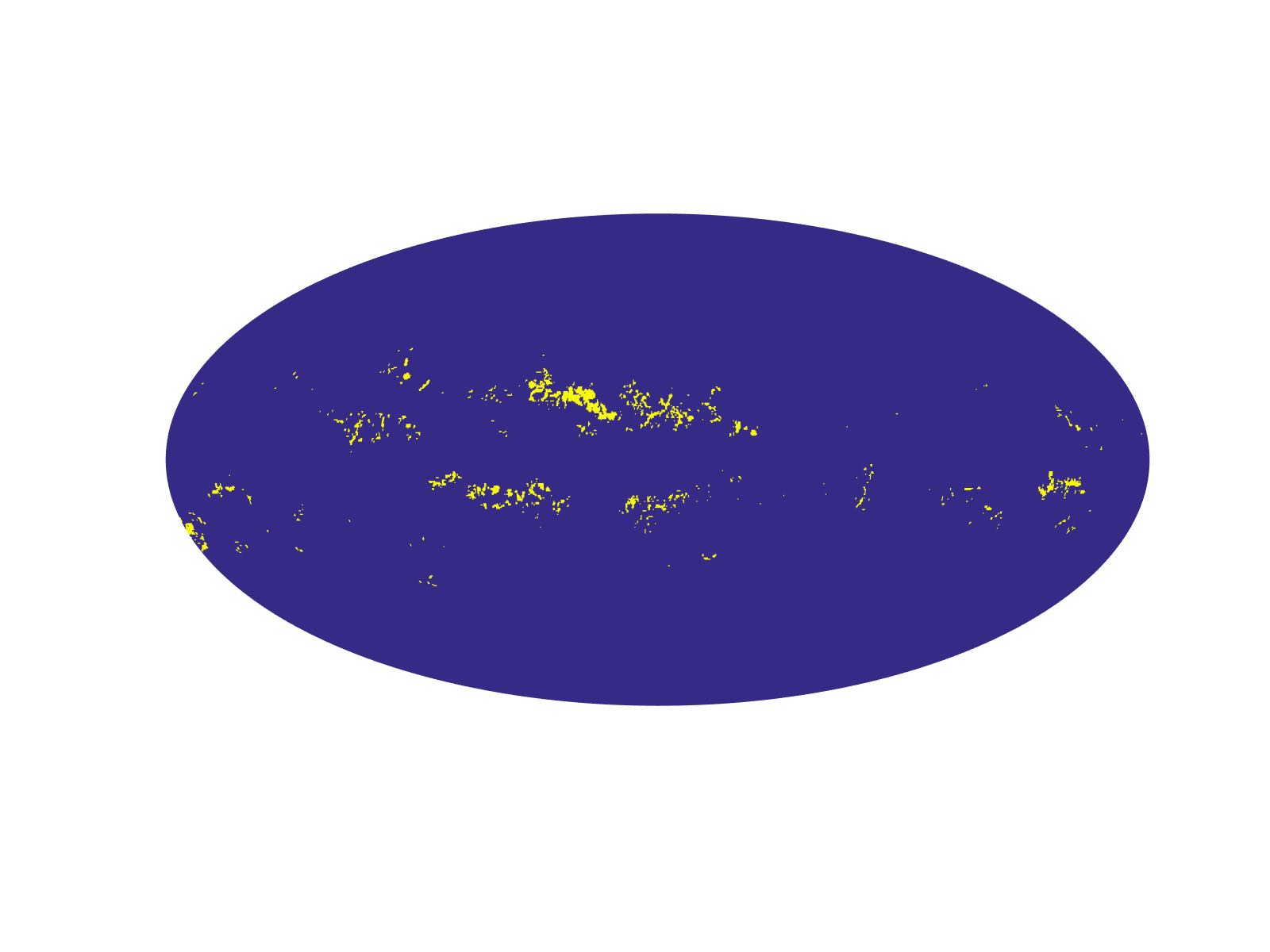} \put(-55,23){\red{\framebox(27,18){ }}}  
                 \\
        		\includegraphics[trim={{1.55\linewidth} {1.25\linewidth} {1.65\linewidth} {1.2\linewidth}}, clip, width=0.2\linewidth]
		{./fig/solar_map_031301_AxisymWav_L_512_Resutls_Th.jpg}  &
        		\includegraphics[trim={{1.55\linewidth} {1.25\linewidth} {1.65\linewidth} {1.2\linewidth}}, clip, width=0.2\linewidth]
		{./fig/solar_map_031301_AxisymWav_L_512_Resutls.jpg}   &
        		\includegraphics[trim={{1.55\linewidth} {1.25\linewidth} {1.65\linewidth} {1.2\linewidth}}, clip, width=0.2\linewidth]
                 {./fig/solar_map_031301_DirectionWav_L_512_N_6_Resutls.jpg} &  
        		\includegraphics[trim={{1.55\linewidth} {1.25\linewidth} {1.65\linewidth} {1.2\linewidth}}, clip, width=0.2\linewidth]
                 {./fig/solar_map_031301_HybridWav_L_512_N_6_Resutls.jpg}  
                 \\
		{\small (e) K-means} & {\small (f) WSSA-A} & {\small (g) WSSA-D} & {\small (h) WSSA-H}
        \end{tabular}
	\caption{Results of solar map.  
	First row: noisy image shown on the sphere (a) and in 2D using a mollweide projection (b), and the zoomed-in red rectangle area 
	of the noisy (c) and original images (d), respectively; 
	Second to fourth rows from left to right: results of methods K-means (e), WSSA-A (f), WSSA-D (g) with $N=6$ (even $N$), 
	and WSSA-H (h), respectively.}
	\label{fig-solarmap-031301}
\end{figure}

\begin{table}[h] 
\begin{center}
\caption{Solar map in Fig.\ \ref{fig-solarmap}: Number of unclassified points at each iteration and computation time in seconds.
	$^\ast$The fourth and fifth columns represent the results of WSSA-D with $N = 5$ and $6$, respectively.} \label{tab:solar}
 \vspace{-0.05in}
\begin{tabular}{|c||c|c|c|c|c|}
\hline  
  & {K-means} & {WSSA-A} & WSSA-D$^\ast$ & WSSA-D$^\ast$ & WSSA-H
\\ \hline \hline
\raisebox{-.15ex}{$|\bar{\mathbb{S}}^2|$} &  $523776$ & $523776$ & $523776$  & $523776$  & $523776$ 
\\ \hline
 $|\Lambda^{(0)}|$ & -  & 17324 &  14422   &  13960  & 13904
\\ \hline
$|\Lambda^{(1)}|$ &   -  & 14480   & 10827   &  10681  &  11326
\\ \hline
$|\Lambda^{(2)}|$ &  -  & 3468 &  2471  &   2484  &  2674
\\ \hline
$|\Lambda^{(3)}|$ &  -  & 884   & 644   &  649   &  714
\\ \hline
$|\Lambda^{(4)}|$ &  -  & 220   &  166  &  174  & 173  
\\ \hline
$|\Lambda^{(5)}|$ &  -  & 60    & 45  &  43  &  36
\\ \hline 
$|\Lambda^{(6)}|$ &  -  & 12   &  10  & 9 &  7
\\ \hline
$|\Lambda^{(7)}|$ &  -   &  0  & 0  & 1  & 1
\\ \hline
$|\Lambda^{(8)}|$ &  -   &  -  & -  &  0 & 0
\\ \hline \hline
Time & $<$ 1 s   &  34.1 s  & 124.3 s & 151.8 s & 682.2 s
\\ \hline
\end{tabular}
\end{center}
\end{table}

The third example is the application of the segmentation algorithms to two different solar data-sets. 
Solar maps are very informative about solar activity, which have direct and indirect impacts on our activities on Earth. 
For illustrative purposes, we apply our method to segment two solar maps which show different solar features. 

The first solar data, presented in Fig.\ \ref{fig-solarmap}, are obtained by synthetically sewing three spacecraft measurements 
taken on $8^{\rm th}$~July~2012 at wavelength 30.4~nm. The three instruments are SDO/AIA\footnote{\url{http://sdo.gsfc.nasa.gov/}}, 
STEREO-A/SECCHI and STEREO-B/SECCHI\footnote{\url{http://www.stereo.rl.ac.uk/}}. 
The three spacecraft orbit around the Sun and together 
they give a full snapshot of the Sun in 360 degrees. The sewing procedures account for the angles each spacecraft instrument covered. 
Then by removing the overlapped observed regions between each pair of instruments and stitching the maps together, 
a complete snapshot of the Sun is obtained. Regions at solar latitudes affected by edge effects and the tilting of the observations 
have their intensity set to zero. A Gaussian filter was applied to smooth the resulting map. 
More sophisticated methods are required to properly combine and interpret the data from different solar instruments but this is beyond 
the scope of this article. Here we focus on segmenting the features seen in a snapshot of the Sun at the wavelength sensitive to solar flares. 
Parameter $\epsilon= 0.04$ is used in our WSSA method. From Fig.\ \ref{fig-solarmap}, one can see that WSSA-A, WSSA-D, 
and WSSA-H methods performed much better than the K-means method, preserving the directional features of the signal more completely. 
Among the three, as is seen from Fig.\ \ref{fig-solarmap} (g) and (h), the quality of WSSA-H is slightly improved compared to the result from WSSA-D (in terms of the area of the sunspots), although the computation time is longer (see Table \ref{tab:solar}). 

The second solar test data, presented in Fig.\ \ref{fig-solarmap-031301}, is the radial-magnetic-field synoptic image 
of the Sun measured from $13^{\rm th}$~March~2001 to $9^{\rm th}$~April~2001 
(Carrington Rotation 1974\footnote{\url{http://jsoc.stanford.edu/cgi-bin/hmisynop.pl?cr=1974&instrument=HMI&mag=Mag}}), 
during which the solar activity peaked and a high number of sunspots were detected. The image shows the spatial variation of the strength of magnetic fields on the Sun (in our demonstration only absolute values are considered). 
We applied the K-means and the WSSA methods, with $\epsilon= 0.05$, to segment active magnetic regions on the Sun. As seen in 
Fig.\ \ref{fig-solarmap-031301} (e)-(h), the K-means method is able to pull out some of the sunspots (i.e.\ most magnetic-active regions) 
but it fails to capture the more diffusive and patchy features within the image. The WSSA-A, WSSA-D, and WSSA-H methods 
give quite similar results, which is mainly because
the solar data themselves do not contain textures with strong directional information. 
At the same time, it can be seen that WSSA-D, and WSSA-H are more immune to noise.

\subsection{Retina images on the sphere}
\begin{figure}
	\centering
		\begin{tabular}{cccc}
		\multicolumn{4}{c}{\bf Original retina images - 2D}  \vspace{-0.015in} 
		\\
  		\includegraphics[trim={{.25\linewidth} {0.1\linewidth} {.25\linewidth} {0.1\linewidth}}, clip, width=0.25\linewidth]
		{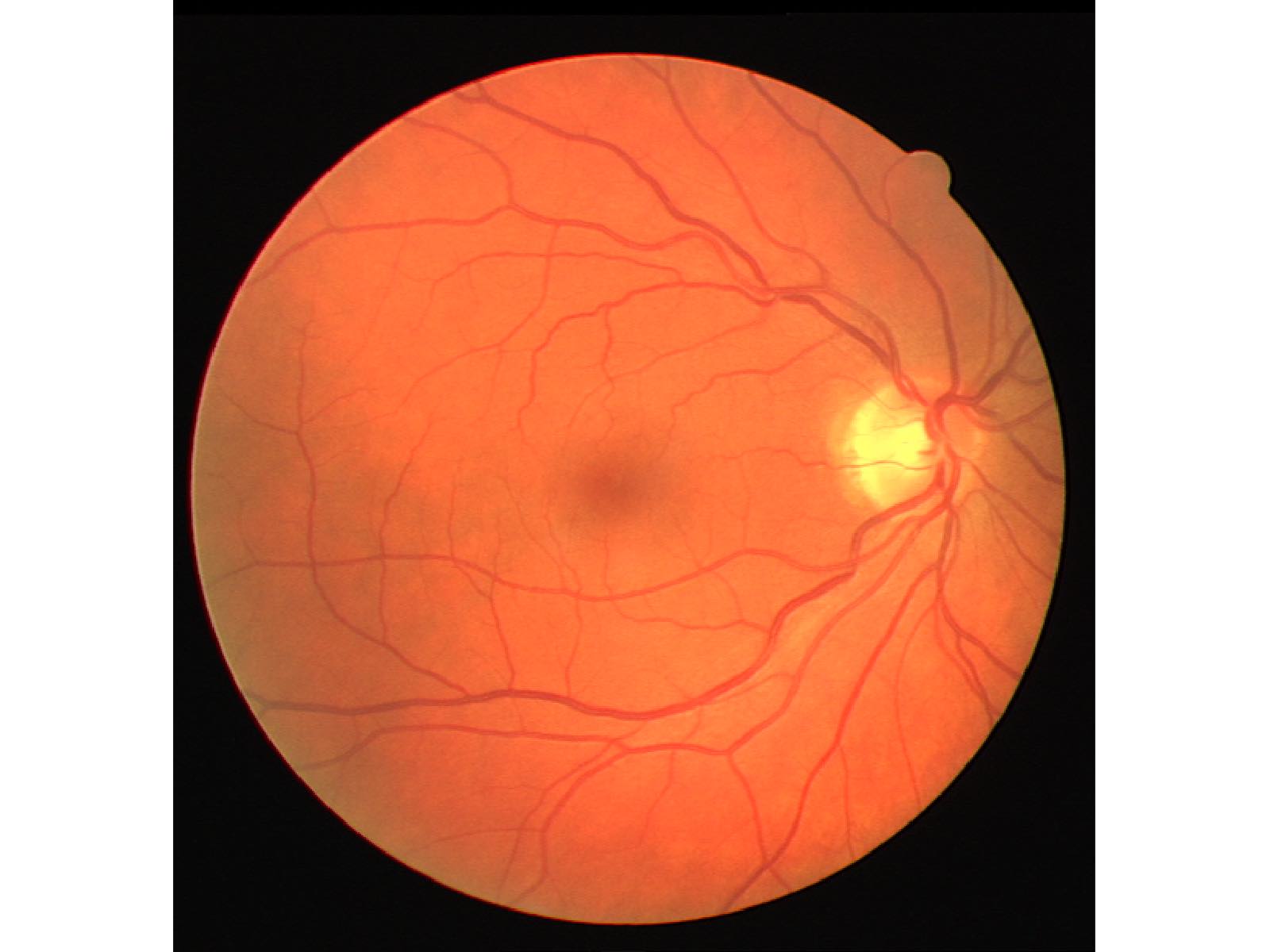} &
		\includegraphics[trim={{.25\linewidth} {0.1\linewidth} {.25\linewidth} {0.1\linewidth}}, clip, width=0.25\linewidth]
		{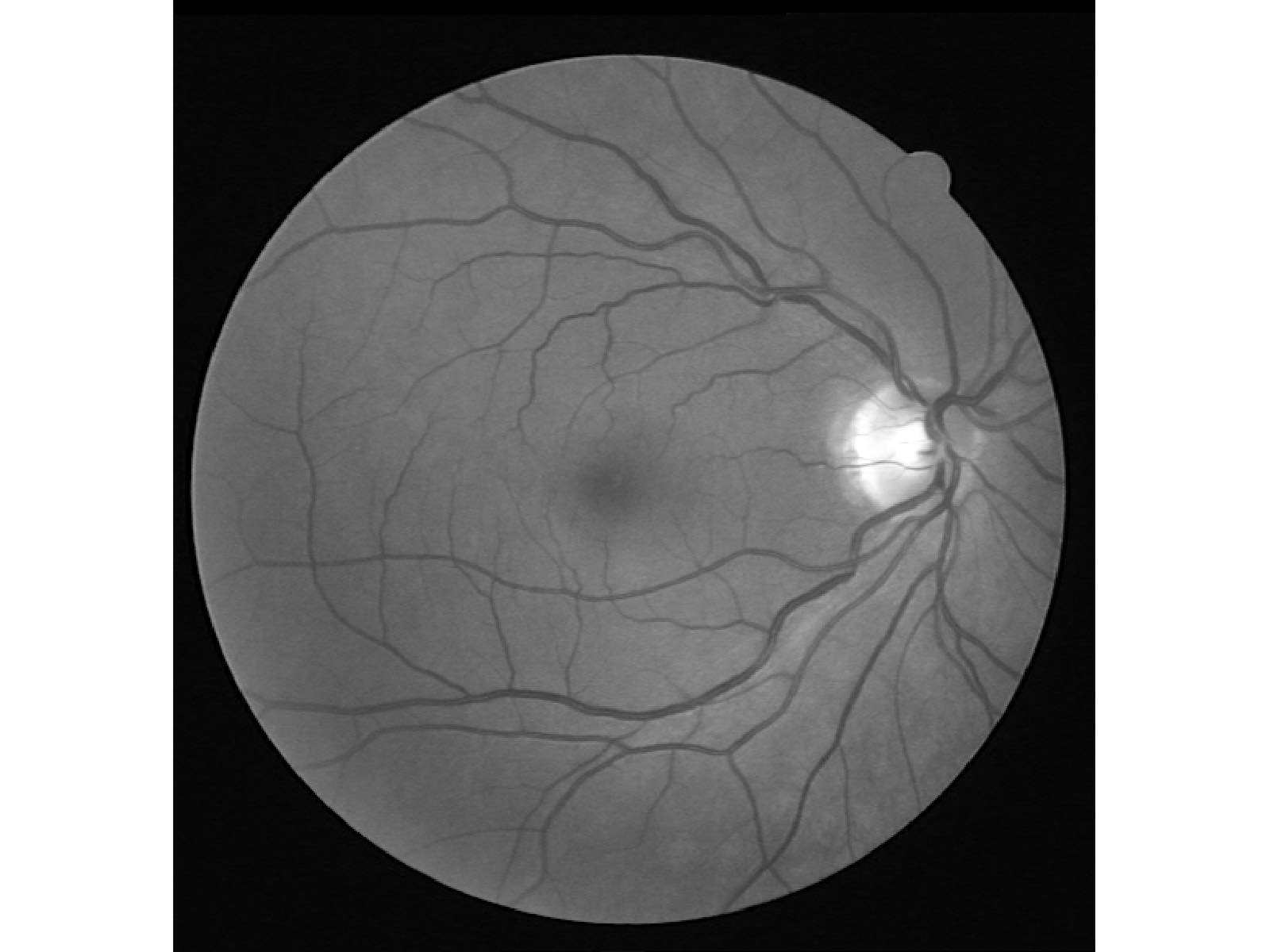} &
	        \includegraphics[trim={{.25\linewidth} {0.1\linewidth} {.25\linewidth} {0.1\linewidth}}, clip, width=0.25\linewidth]
		{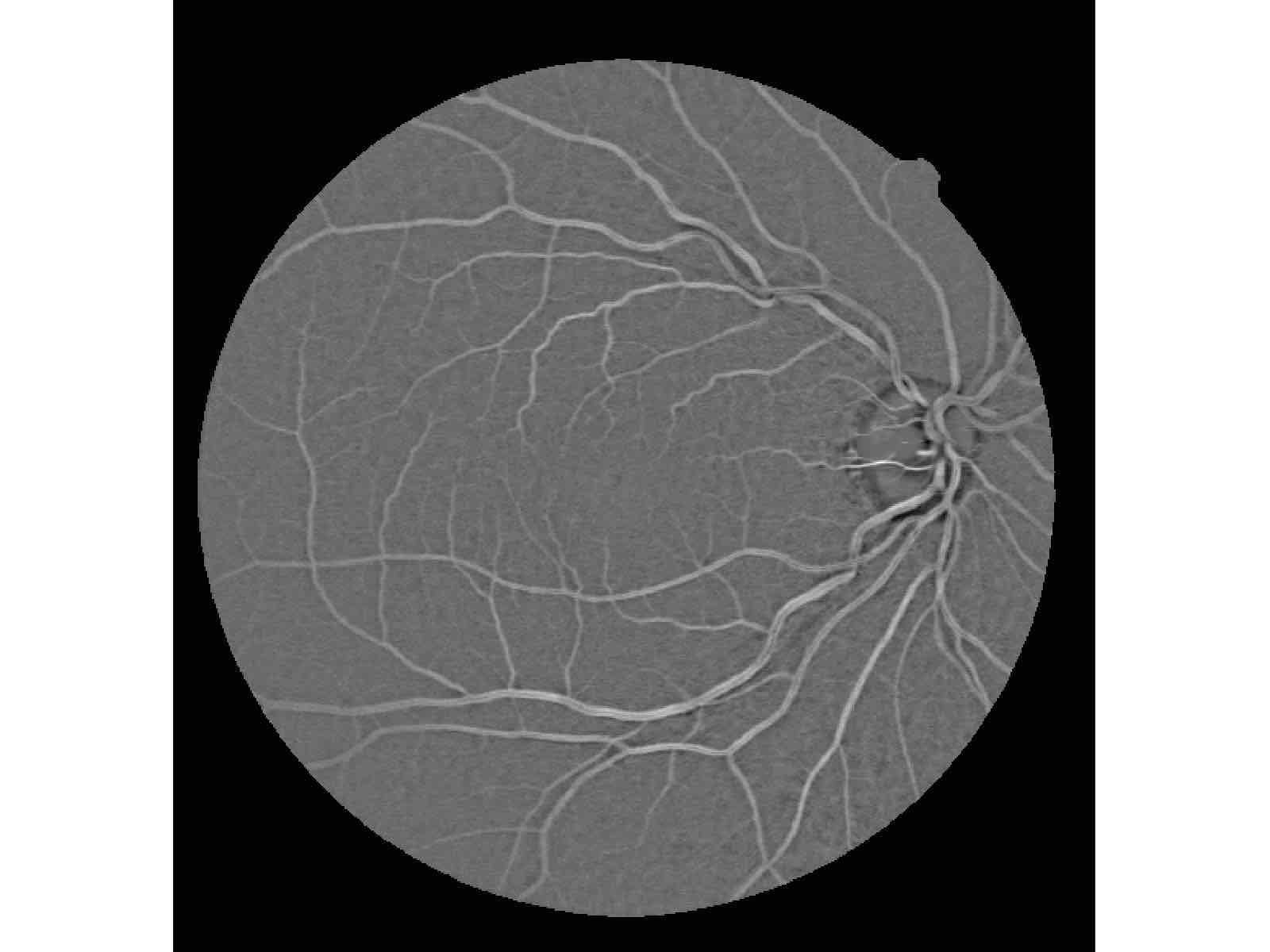} &
		 \includegraphics[trim={{.25\linewidth} {0.1\linewidth} {.25\linewidth} {0.1\linewidth}}, clip, width=0.25\linewidth]
		{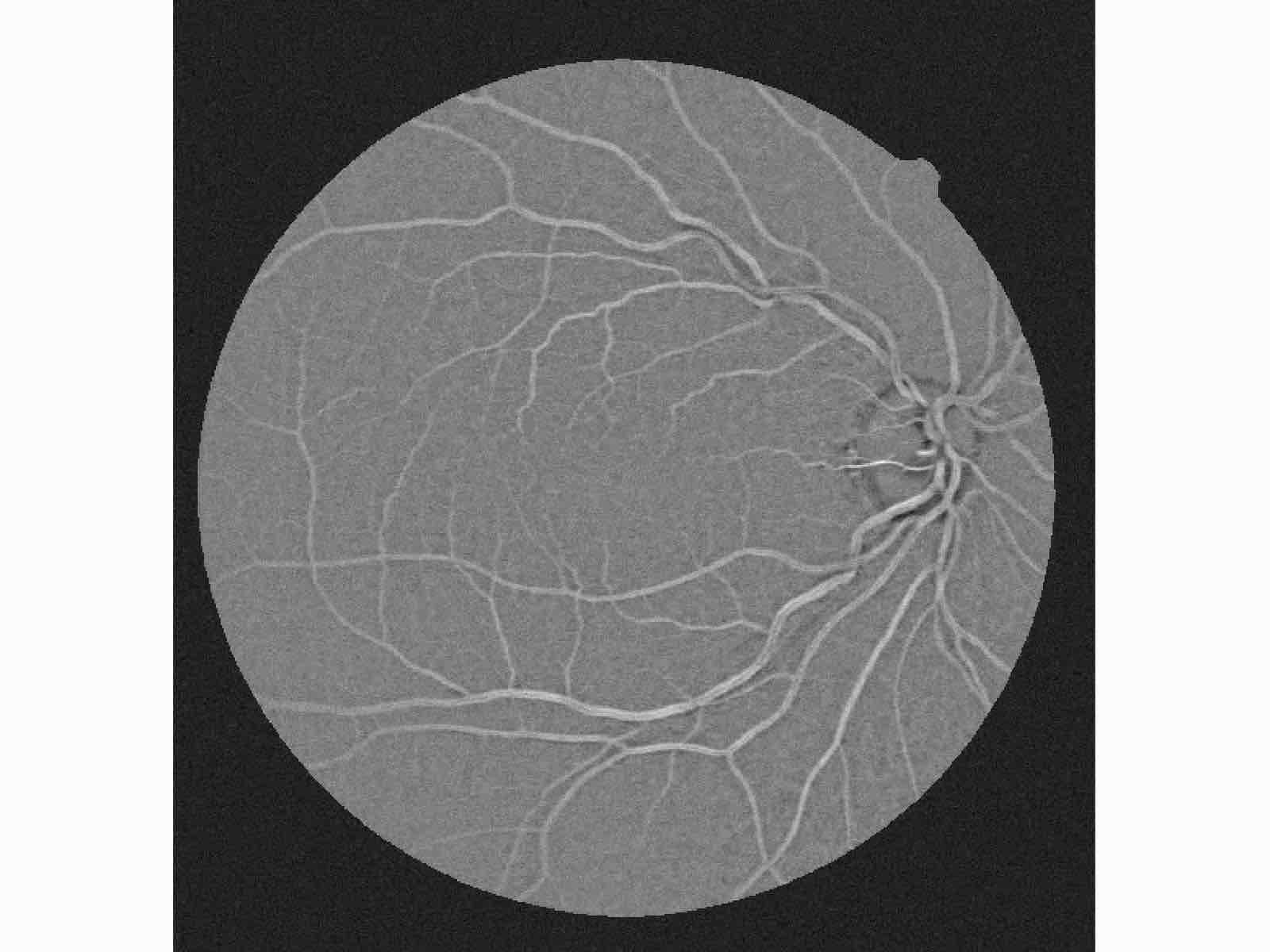}   \vspace{-0.05in}
		\\
	       {\footnotesize (a1) colour image} & {\footnotesize (b1) green channel} & {\footnotesize (c1) tidy background} & {\footnotesize (d1) noisy image}  \vspace{0.15in}
		\\
	        \multicolumn{4}{c}{\bf Test data} \vspace{-0.0in}
	        \\
		\includegraphics[trim={{.9\linewidth} {.32\linewidth} {.8\linewidth} {.6\linewidth}}, clip, width=0.2\linewidth]
		{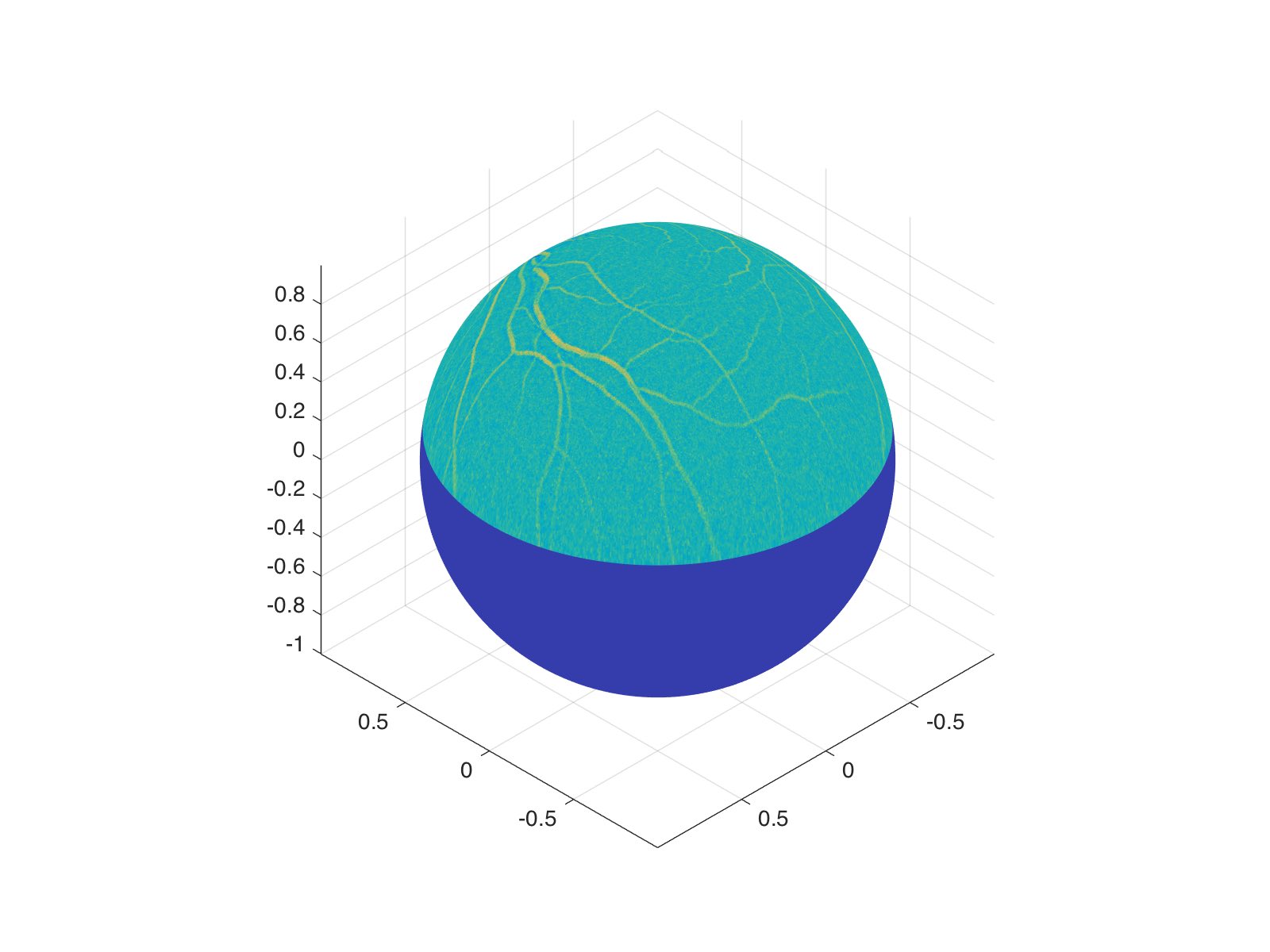} &
		\includegraphics[trim={{.28\linewidth} {.22\linewidth} {.19\linewidth} {.2\linewidth}}, clip, width=0.2\linewidth]
		{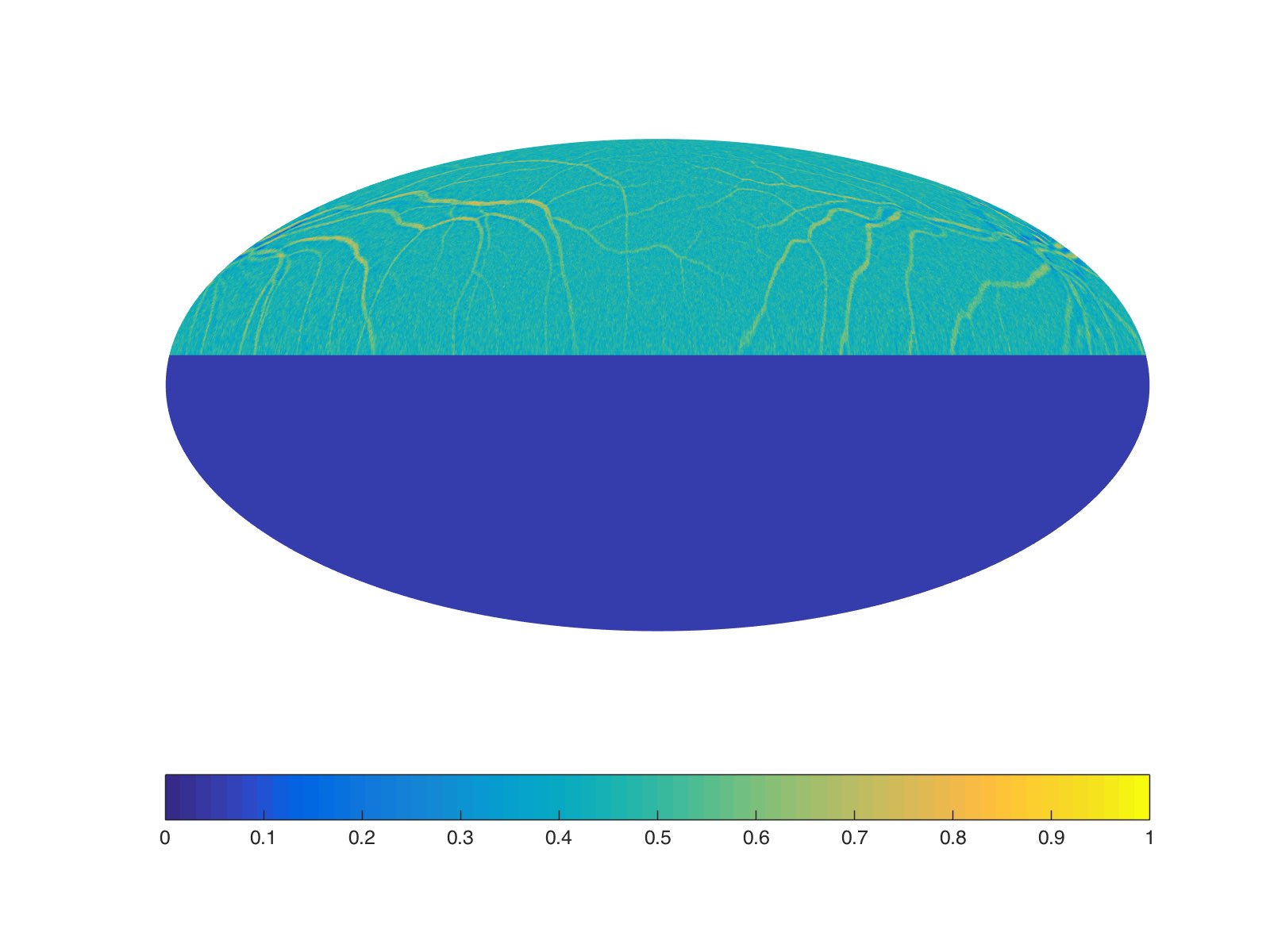}  \put(-56,38){\red{\framebox(10,8){ }}} &
		\includegraphics[trim={{1.4\linewidth} {2.3\linewidth} {2.5\linewidth} {0.63\linewidth}}, clip, width=0.2\linewidth]
		{./fig/retina02_AxisymWav_L_512_Noisy_image.jpg} & 
		\includegraphics[trim={{1.4\linewidth} {2.3\linewidth} {2.5\linewidth} {0.63\linewidth}}, clip, width=0.2\linewidth]
		{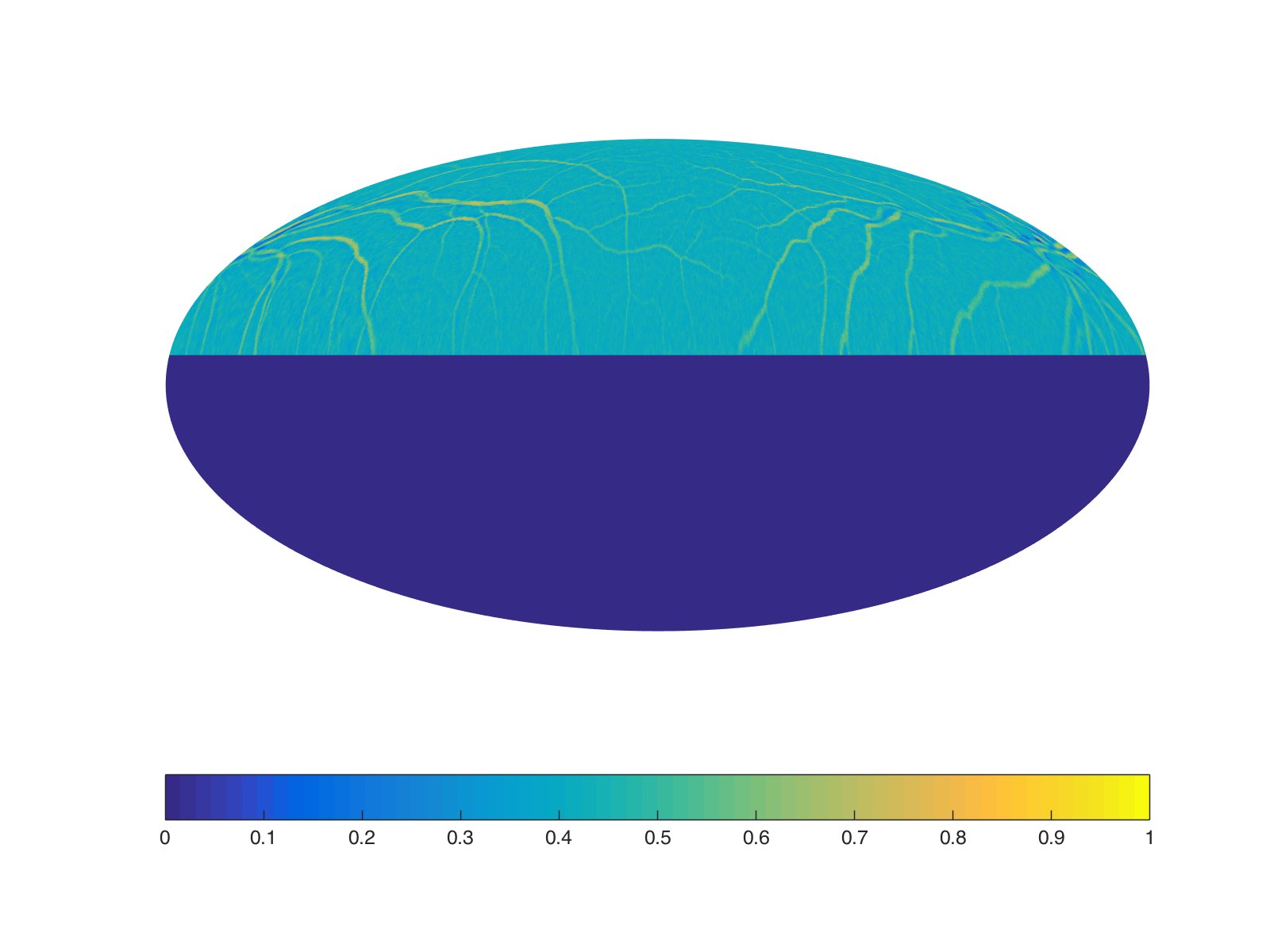}
		\\
		{\small (a) noisy image} & {\small (b) noisy image} & {\small (c) noisy image} & {\small (d) original image}  \vspace{0.15in}
		\\
		 \multicolumn{4}{c}{\bf Segmentation results}  \vspace{-0.02in}
	        \\
	        \includegraphics[trim={{.9\linewidth} {.32\linewidth} {.8\linewidth} {.6\linewidth}}, clip, width=0.2\linewidth]
		{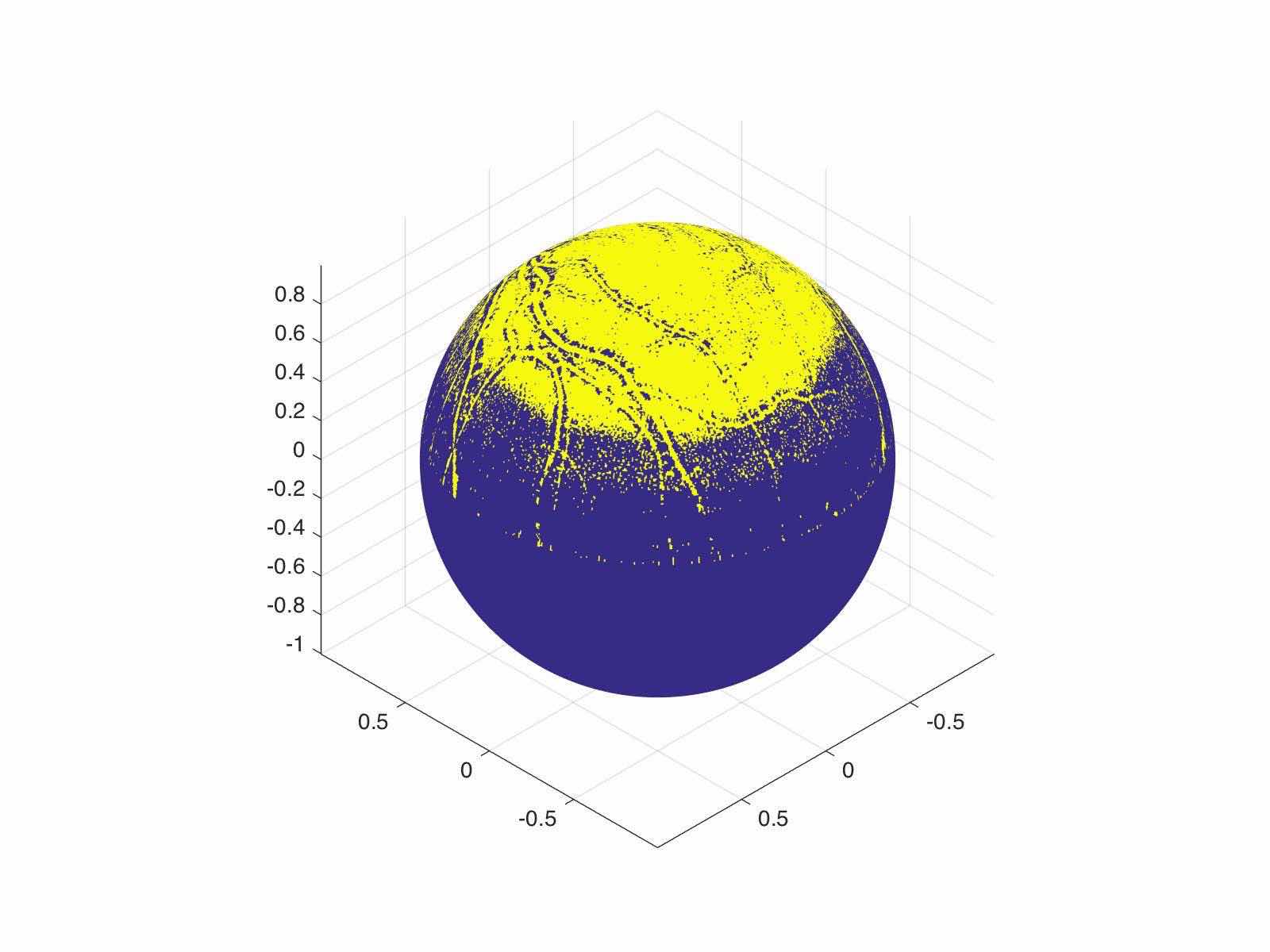} &
        		\includegraphics[trim={{.9\linewidth} {.32\linewidth} {.8\linewidth} {.6\linewidth}}, clip, width=0.2\linewidth]
		{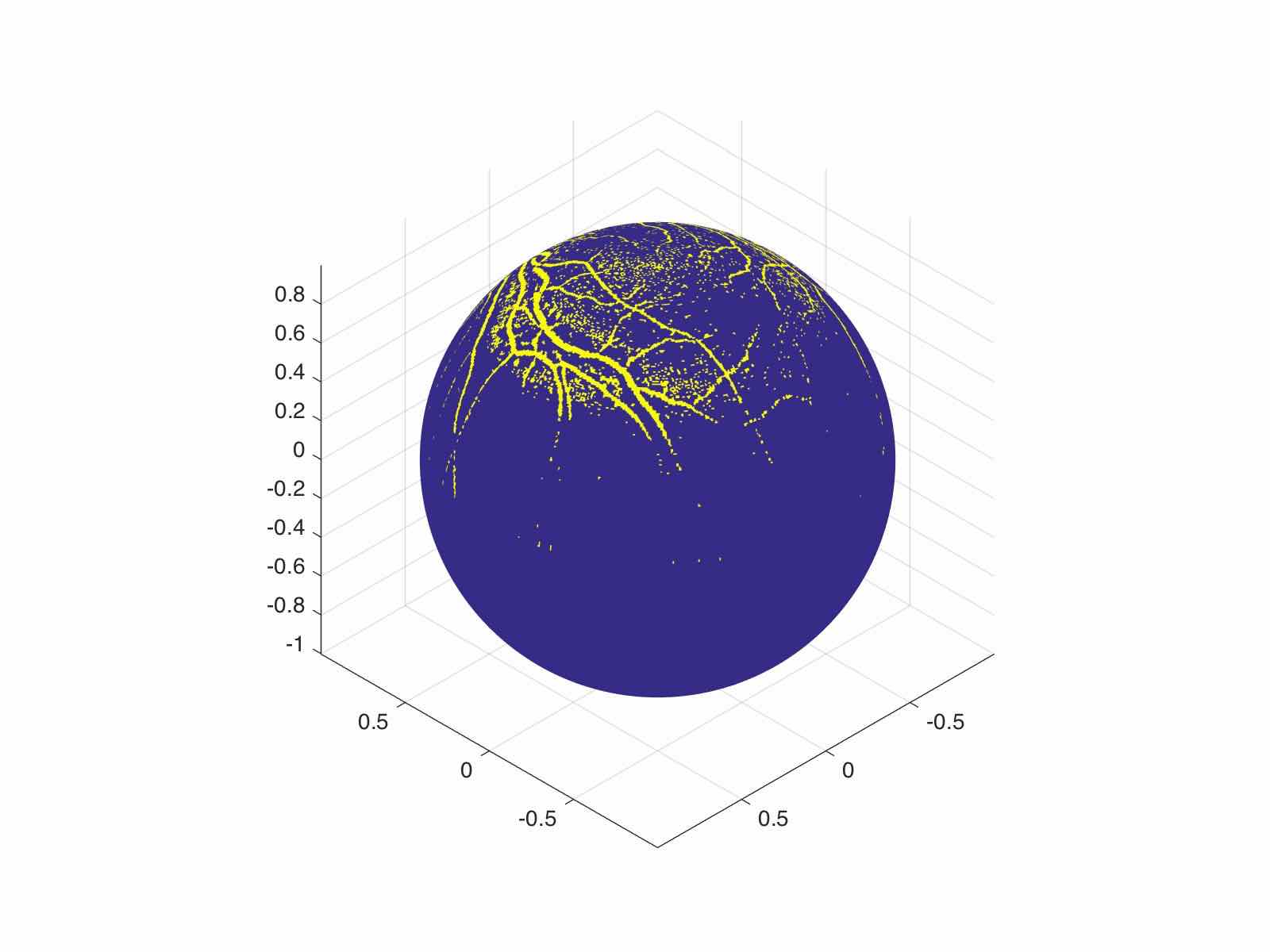} &
        		\includegraphics[trim={{.9\linewidth} {.32\linewidth} {.8\linewidth} {.6\linewidth}}, clip, width=0.2\linewidth]
		{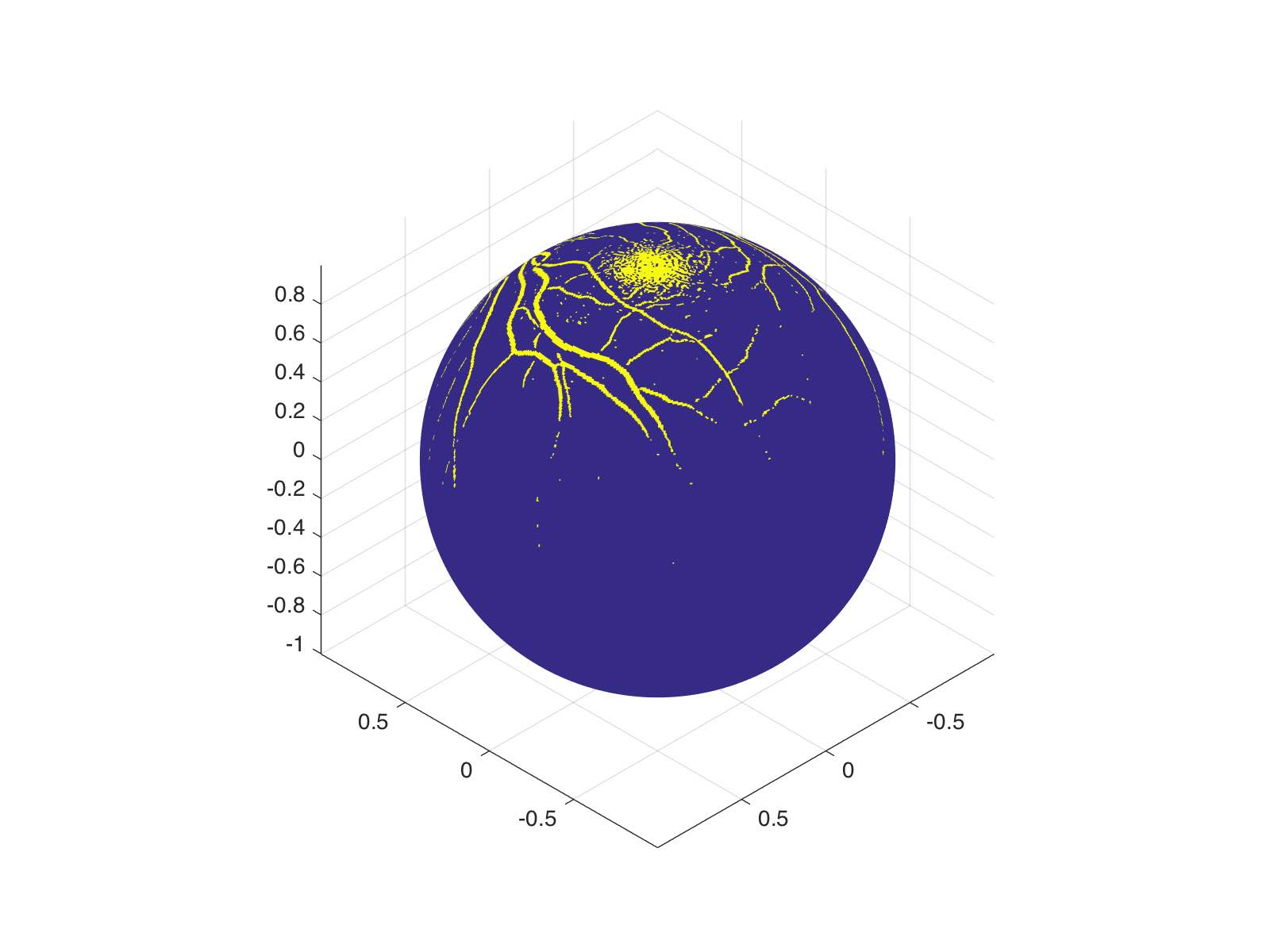} &
        		\includegraphics[trim={{.9\linewidth} {.32\linewidth} {.8\linewidth} {.6\linewidth}}, clip, width=0.2\linewidth]
		{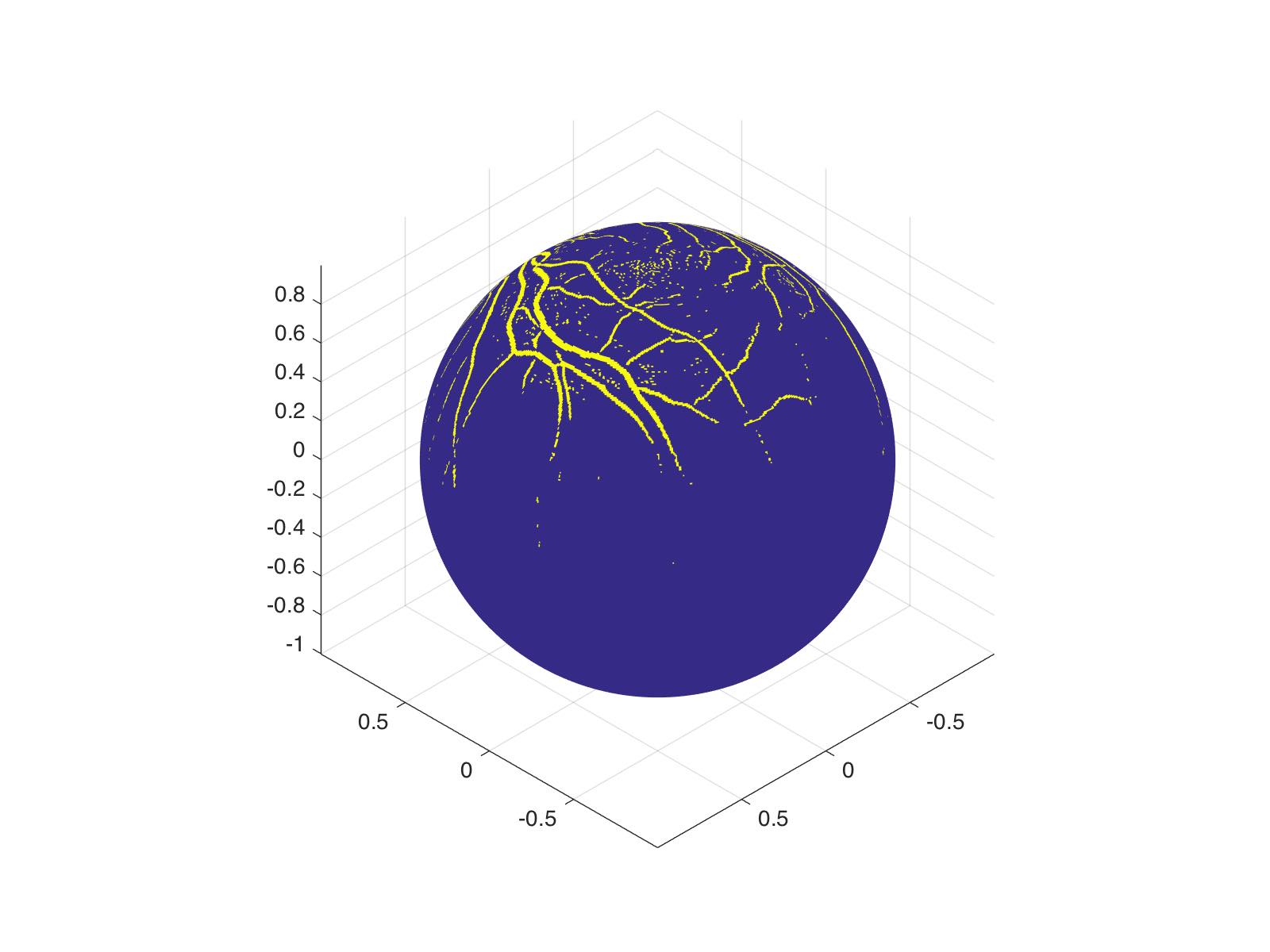} 
		\\
		\includegraphics[trim={{.4\linewidth} {.2\linewidth} {.35\linewidth} {.2\linewidth}}, clip, width=0.2\linewidth]
		{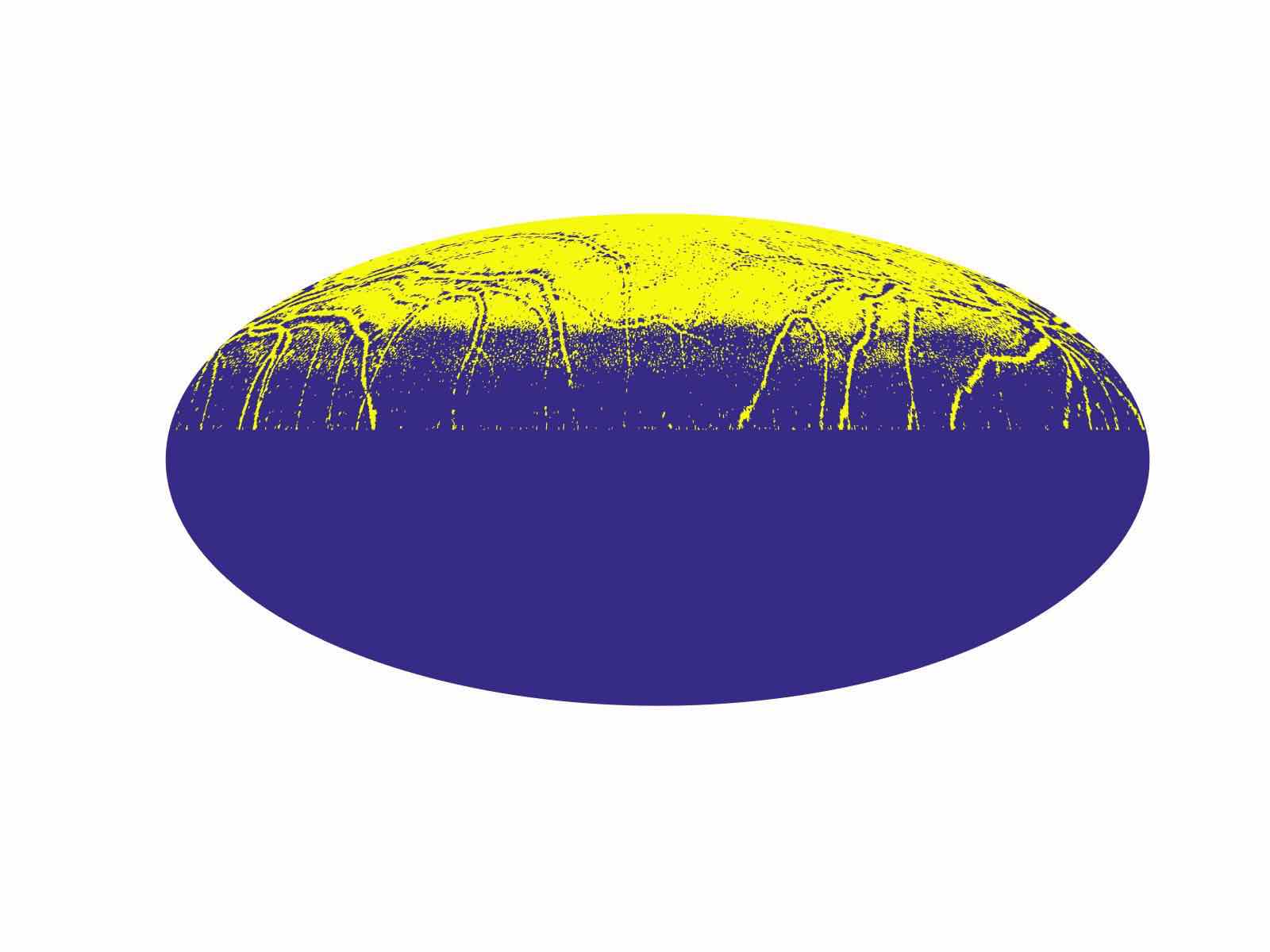}  \put(-56,38){\red{\framebox(10,8){ }}}  &
        		\includegraphics[trim={{.4\linewidth} {.2\linewidth} {.35\linewidth} {.2\linewidth}}, clip, width=0.2\linewidth]
		{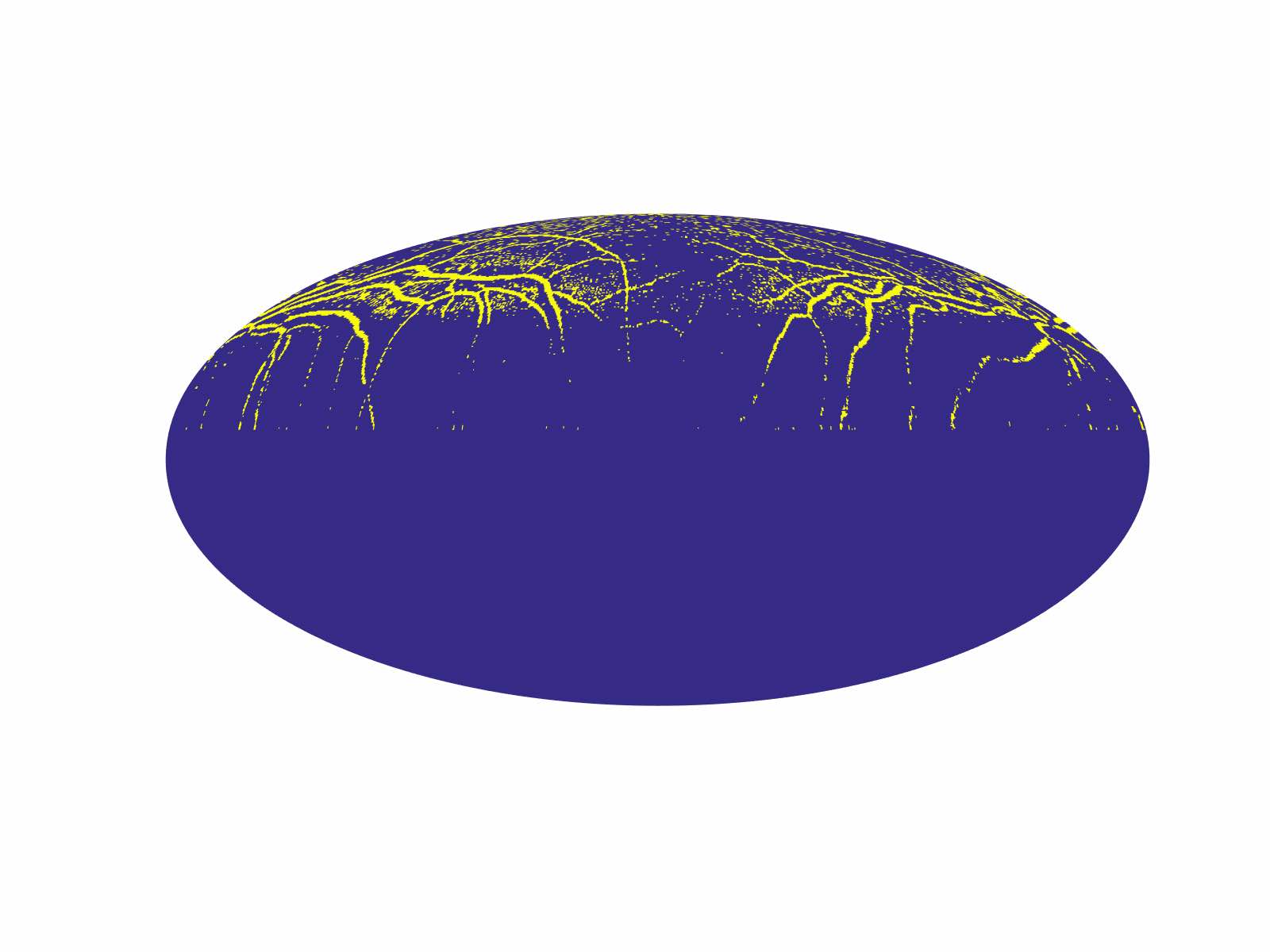}  \put(-56,38){\red{\framebox(10,8){ }}}  &
                 \includegraphics[trim={{.4\linewidth} {.2\linewidth} {.35\linewidth} {.2\linewidth}}, clip, width=0.2\linewidth]
                 {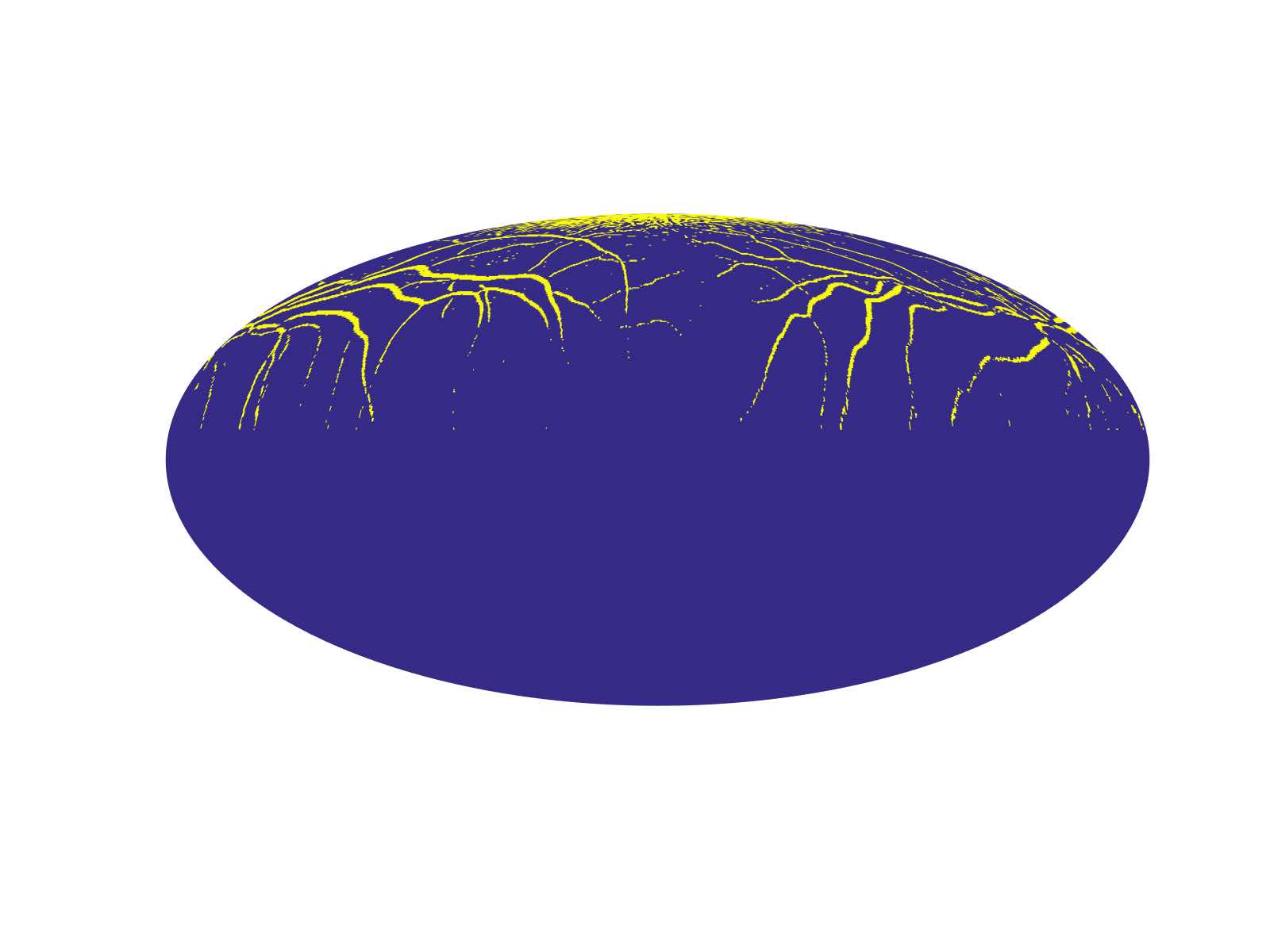}   \put(-56,38){\red{\framebox(10,8){ }}}  &  
                 \includegraphics[trim={{.4\linewidth} {.2\linewidth} {.35\linewidth} {.2\linewidth}}, clip, width=0.2\linewidth]
                 {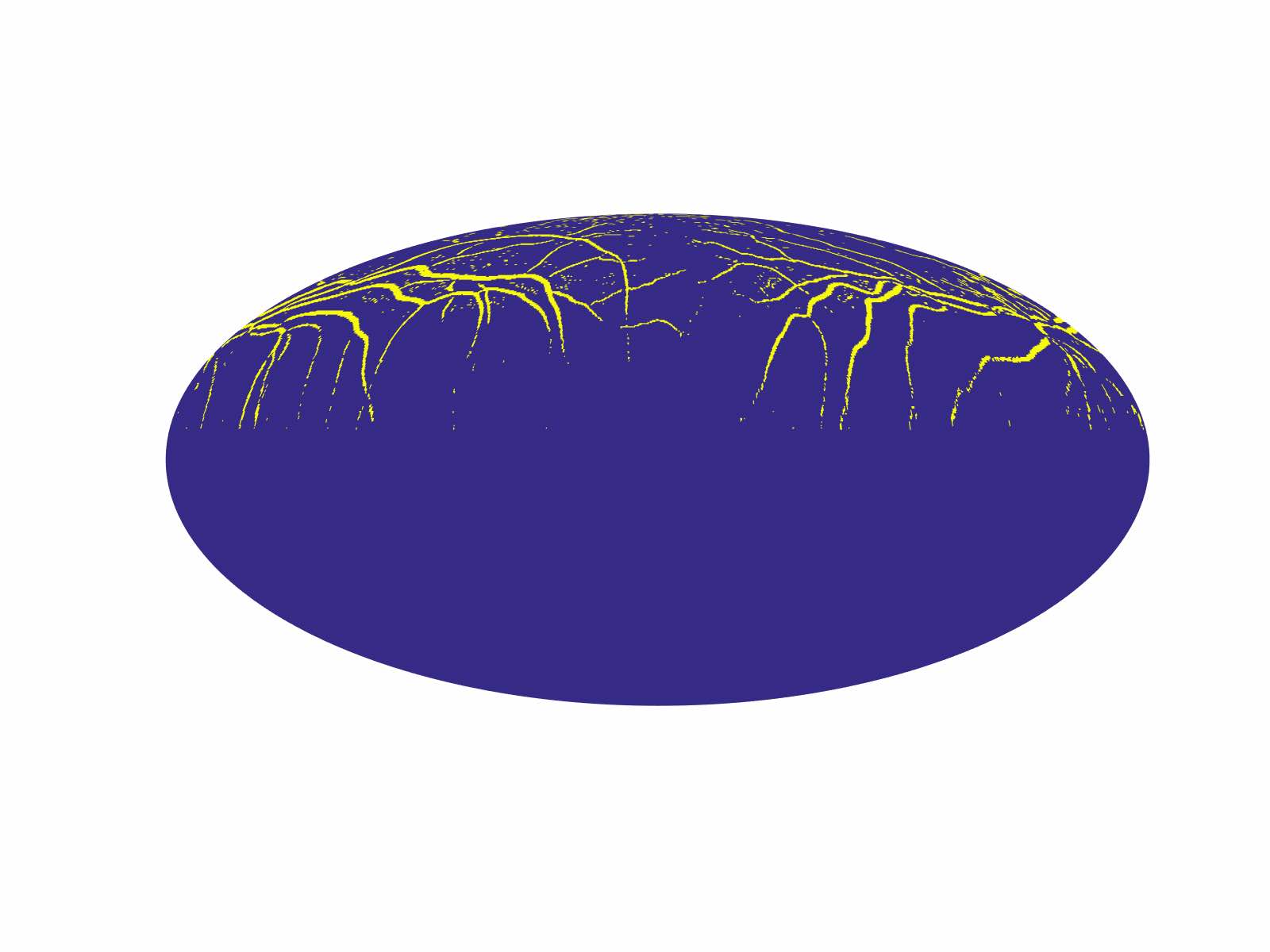}   \put(-56,38){\red{\framebox(10,8){ }}} 
                 \\
        		\includegraphics[trim={{1.4\linewidth} {2.0\linewidth} {2.5\linewidth} {0.90\linewidth}}, clip, width=0.2\linewidth]
		{./fig/retina02_AxisymWav_L_512_Resutls_Th.jpg}  &
        		\includegraphics[trim={{1.4\linewidth} {2.0\linewidth} {2.5\linewidth} {0.90\linewidth}}, clip, width=0.2\linewidth]
		{./fig/retina02_AxisymWav_L_512_Resutls.jpg}   &
        		\includegraphics[trim={{1.4\linewidth} {2.0\linewidth} {2.5\linewidth} {0.90\linewidth}}, clip, width=0.2\linewidth]
                 {./fig/retina02_DirectionWav_L_512_N_5_Resutls.jpg} &  
        		\includegraphics[trim={{1.4\linewidth} {2.0\linewidth} {2.5\linewidth} {0.90\linewidth}}, clip, width=0.2\linewidth]
                 {./fig/retina02_HybridWav_L_512_N_6_Resutls_64.jpg}  
                 \\
		{\small (e) K-means} & {\small (f) WSSA-A} & {\small (g) WSSA-D} & {\small (h) WSSA-H}
        \end{tabular}
        	\caption{Results of retina image.  
	First row: original retina image (a1), its green channel (b1), the image after removing its background (c1) and the noisy image (d1) respectively; 
	Second row: projected retina image shown on the sphere (a) and in 2D using a mollweide projection (b), and the zoomed-in red rectangle area 
	of the noisy (c) and original images (d), respectively; 
	Third row to fifth row: results of methods K-means (e), WSSA-A (f), WSSA-D (g) with $N=5$ (odd $N$), 
	WSSA-H (h) with $L_{\rm trans}=64$ for curvelets, respectively.}	
	\label{fig-retina02}
\end{figure}

\begin{figure}
	\centering
		\begin{tabular}{cccc}
		\multicolumn{4}{c}{\bf Original retina images - 2D}  \vspace{-0.015in} 
		\\
  		\includegraphics[trim={{.2\linewidth} {0\linewidth} {.2\linewidth} {0\linewidth}}, clip, width=0.242\linewidth]
		{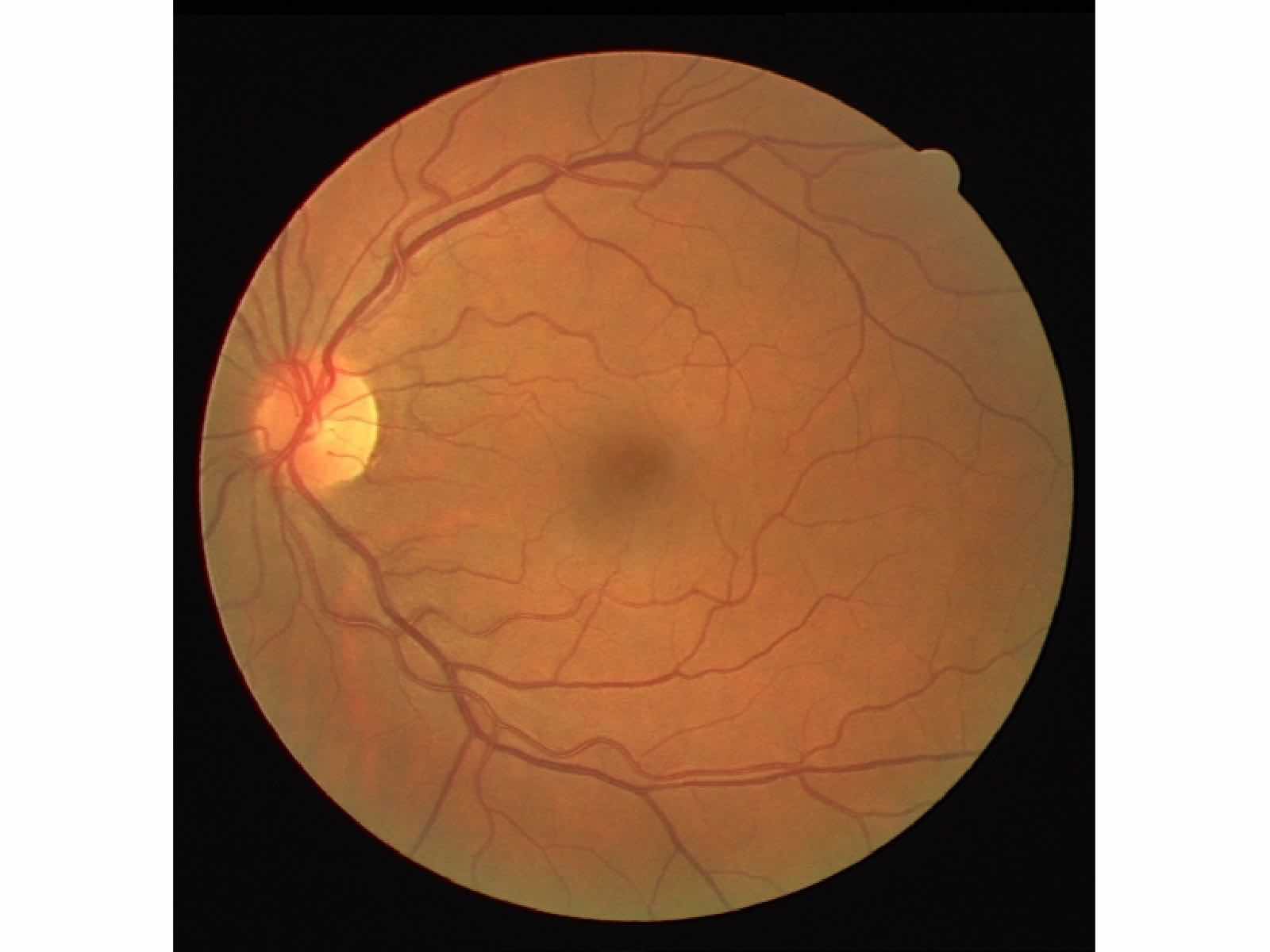} &
		\includegraphics[trim={{.2\linewidth} {0\linewidth} {.2\linewidth} {0\linewidth}}, clip, width=0.242\linewidth]
		{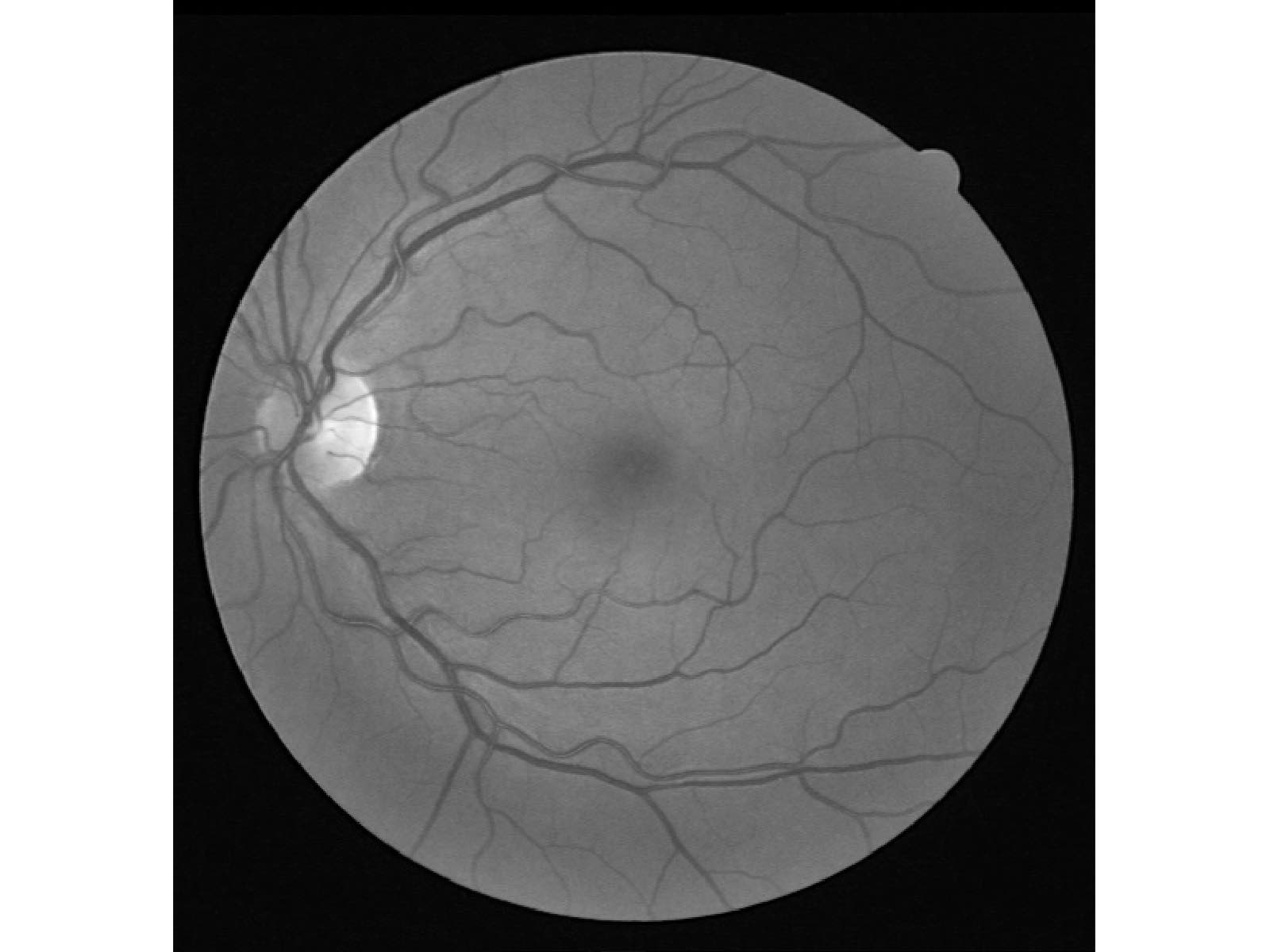} &
	        \includegraphics[trim={{.2\linewidth} {0\linewidth} {.2\linewidth} {0\linewidth}}, clip, width=0.242\linewidth]
		{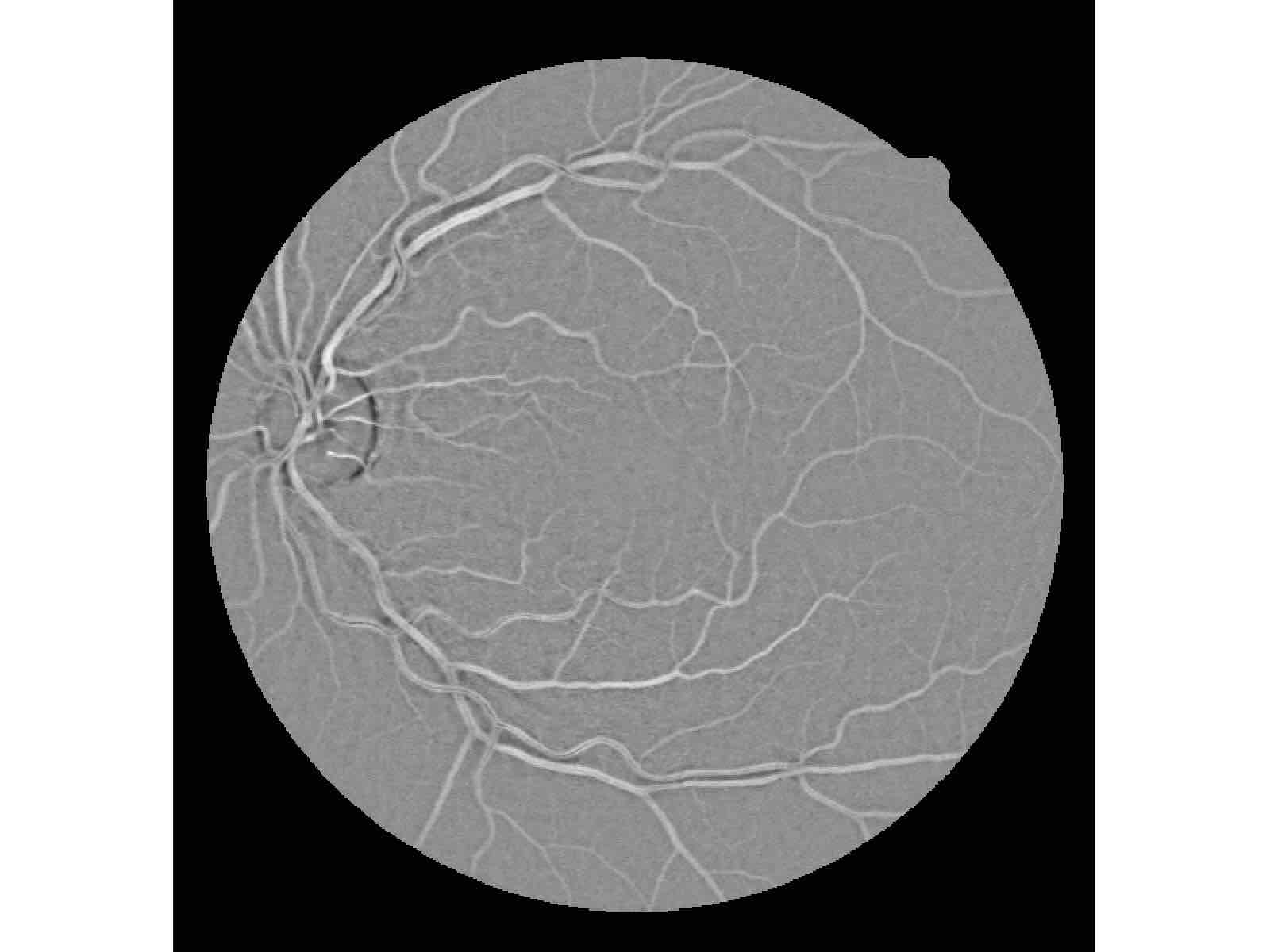} &
		 \includegraphics[trim={{.2\linewidth} {0\linewidth} {.2\linewidth} {0\linewidth}}, clip, width=0.242\linewidth]
		{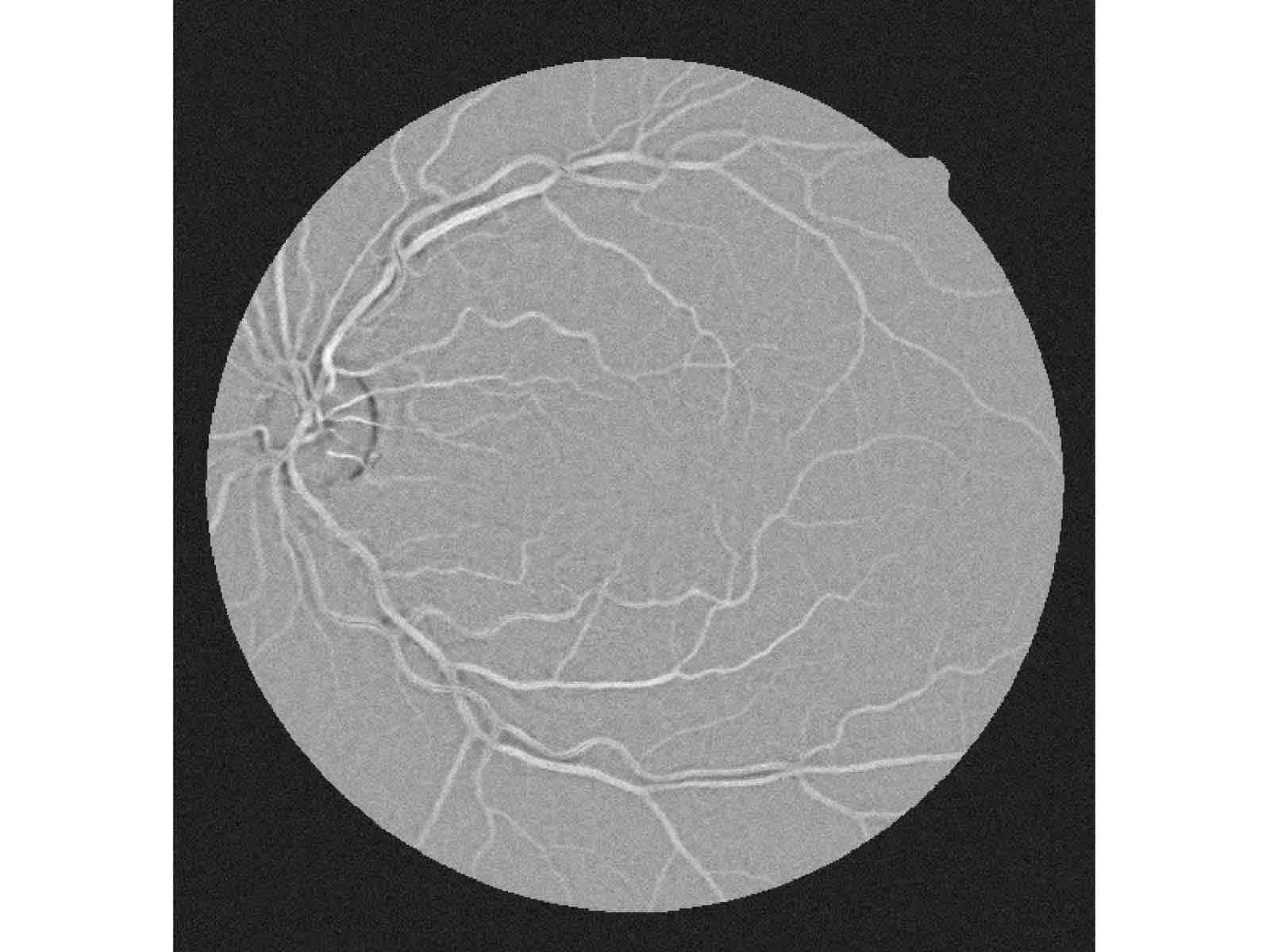}   \vspace{-0.05in}
		\\
	       {\footnotesize (a1) colour image} & {\footnotesize (b1) green channel} & {\footnotesize (c1) tidy background} & {\footnotesize (d1) noisy image}  \vspace{0.15in}
		\\
	        \multicolumn{4}{c}{\bf Test data} \vspace{-0.1in}
	        \\
		\includegraphics[trim={{.9\linewidth} {.32\linewidth} {.8\linewidth} {.6\linewidth}}, clip, width=0.2\linewidth]
		{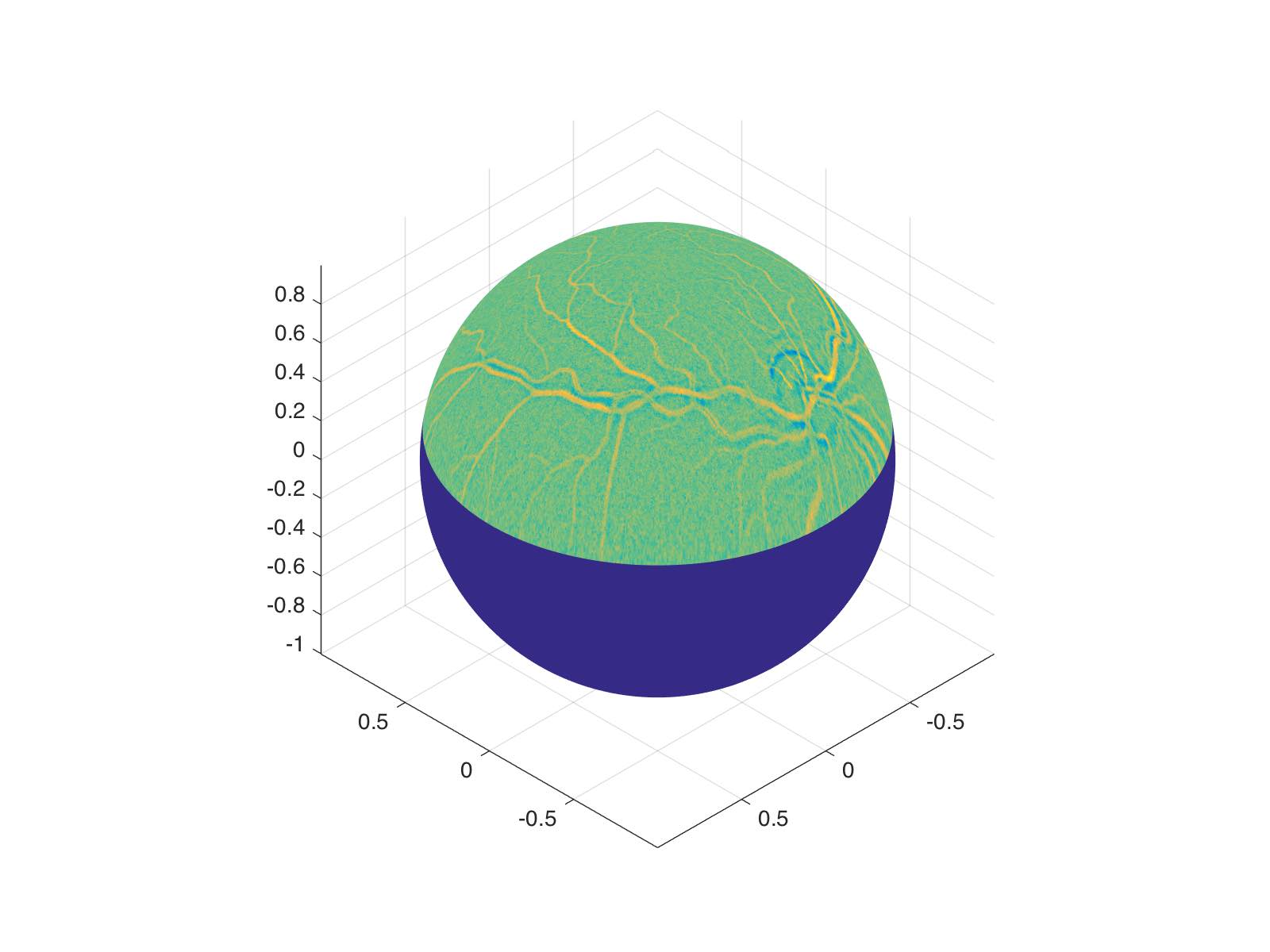} &
		\includegraphics[trim={{.28\linewidth} {.22\linewidth} {.19\linewidth} {.2\linewidth}}, clip, width=0.2\linewidth]
		{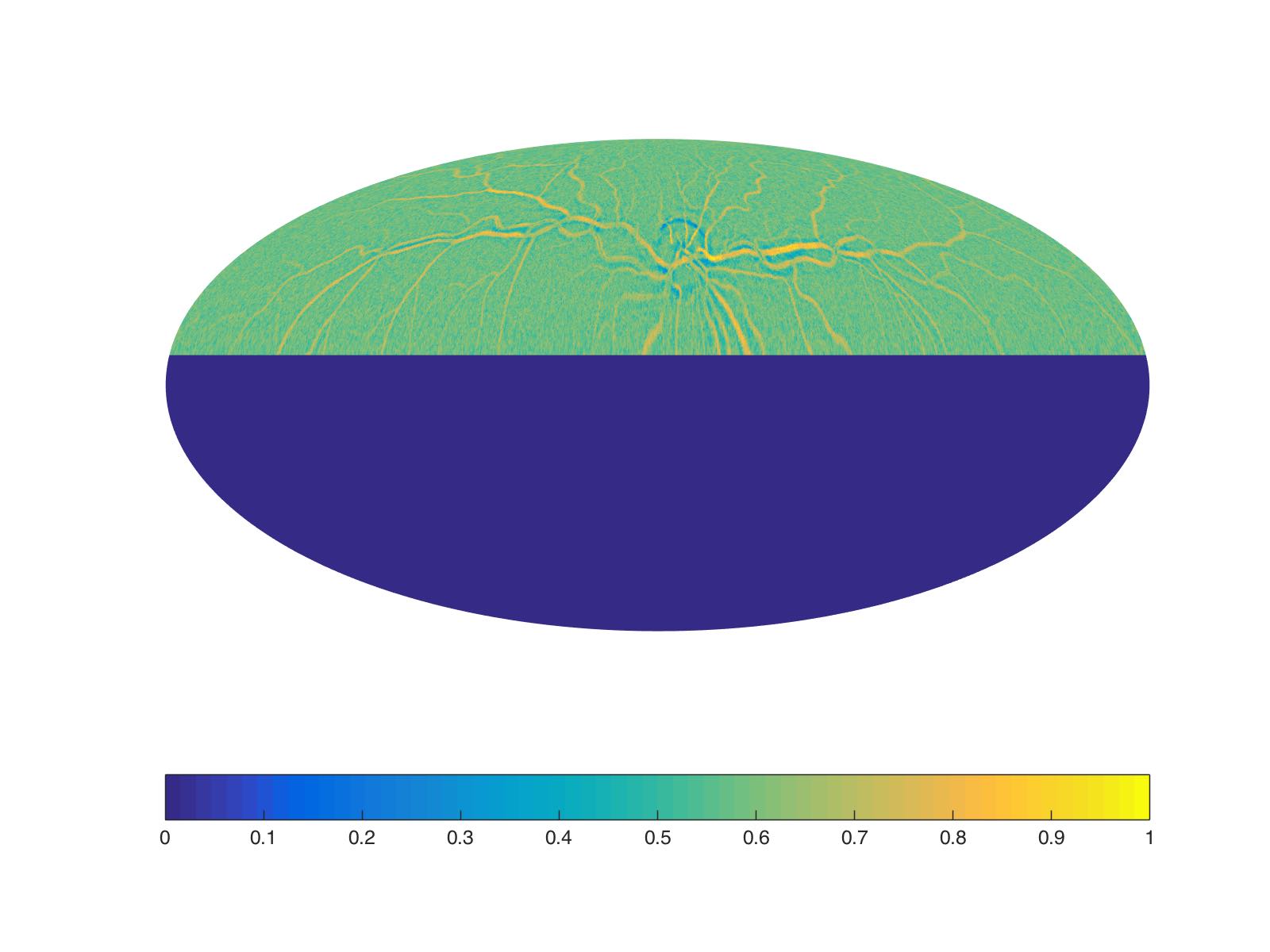}  \put(-56,38){\red{\framebox(10,8){ }}}  &
		\includegraphics[trim={{1.4\linewidth} {2.3\linewidth} {2.5\linewidth} {0.63\linewidth}}, clip, width=0.2\linewidth]
		{./fig/retina01_AxisymWav_L_512_Noisy_image.jpg} & 
		\includegraphics[trim={{1.4\linewidth} {2.3\linewidth} {2.5\linewidth} {0.63\linewidth}}, clip, width=0.2\linewidth]
		{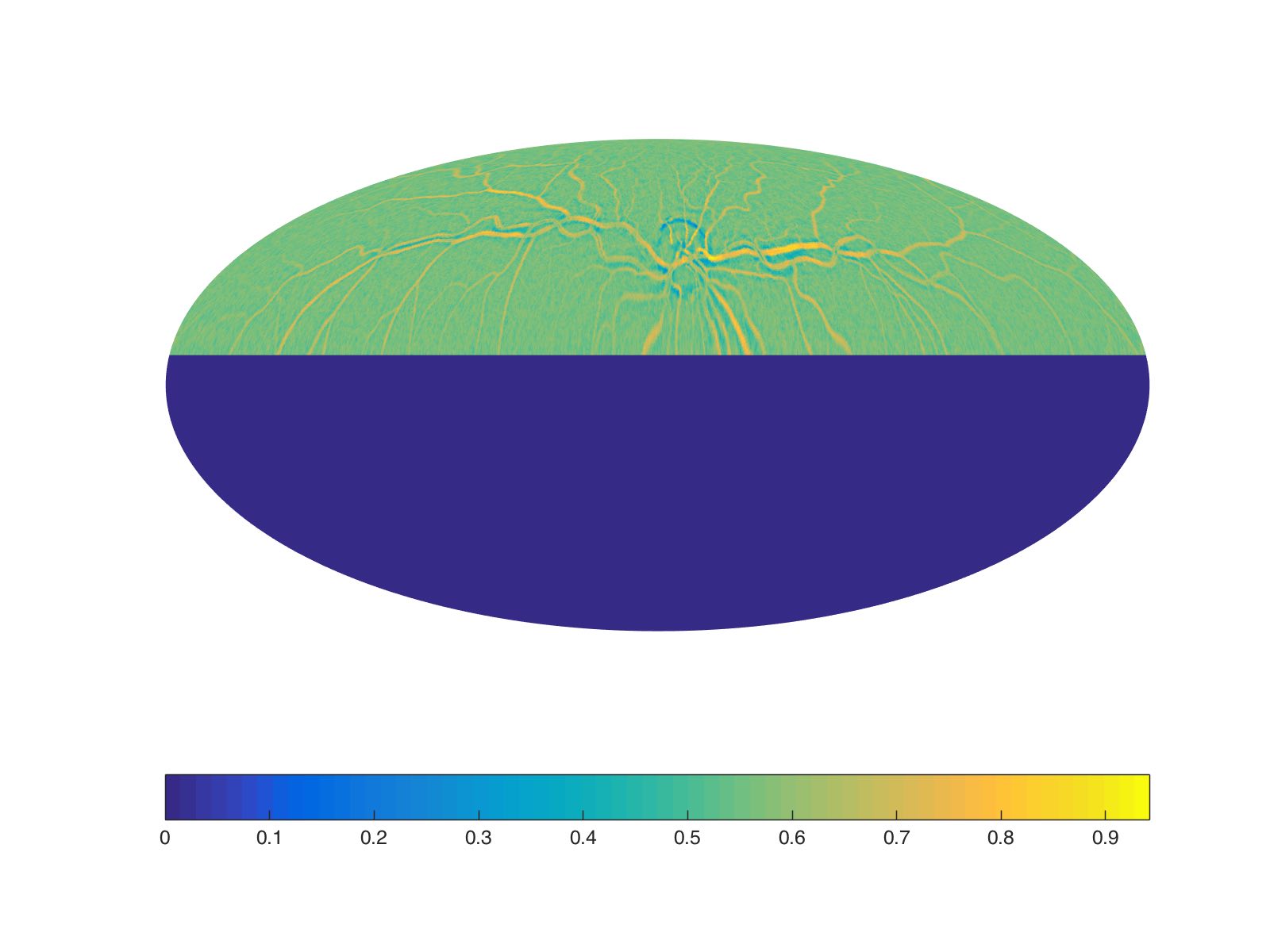}
		\\
		{\small (a) noisy image} & {\small (b) noisy image} & {\small (c) noisy image} & {\small (d) original image}  \vspace{0.15in}
		\\
		 \multicolumn{4}{c}{\bf Segmentation results}  \vspace{-0.02in}
	        \\
	        \includegraphics[trim={{.9\linewidth} {.32\linewidth} {.8\linewidth} {.6\linewidth}}, clip, width=0.2\linewidth]
		{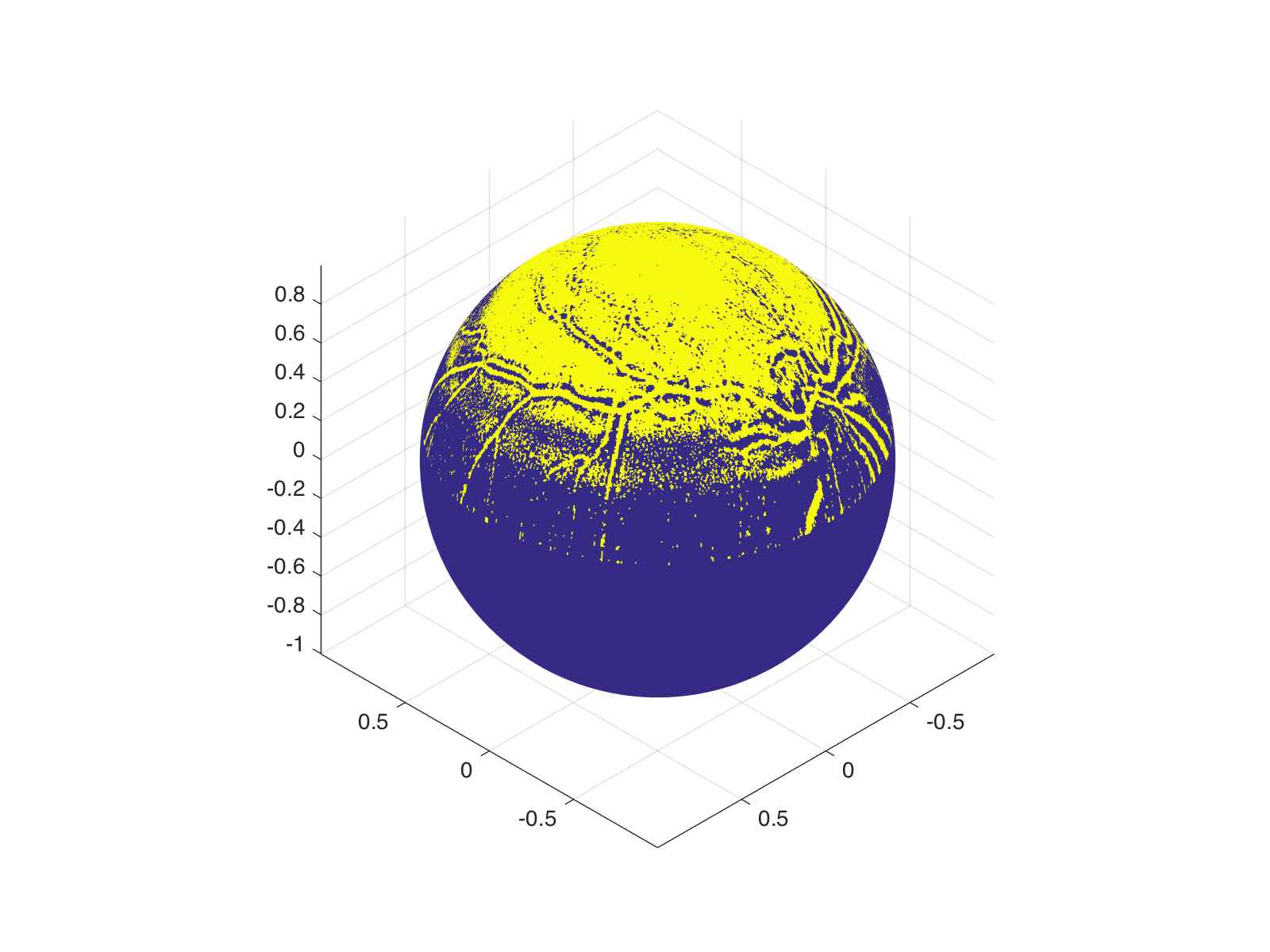} &
        		\includegraphics[trim={{.9\linewidth} {.32\linewidth} {.8\linewidth} {.6\linewidth}}, clip, width=0.2\linewidth]
		{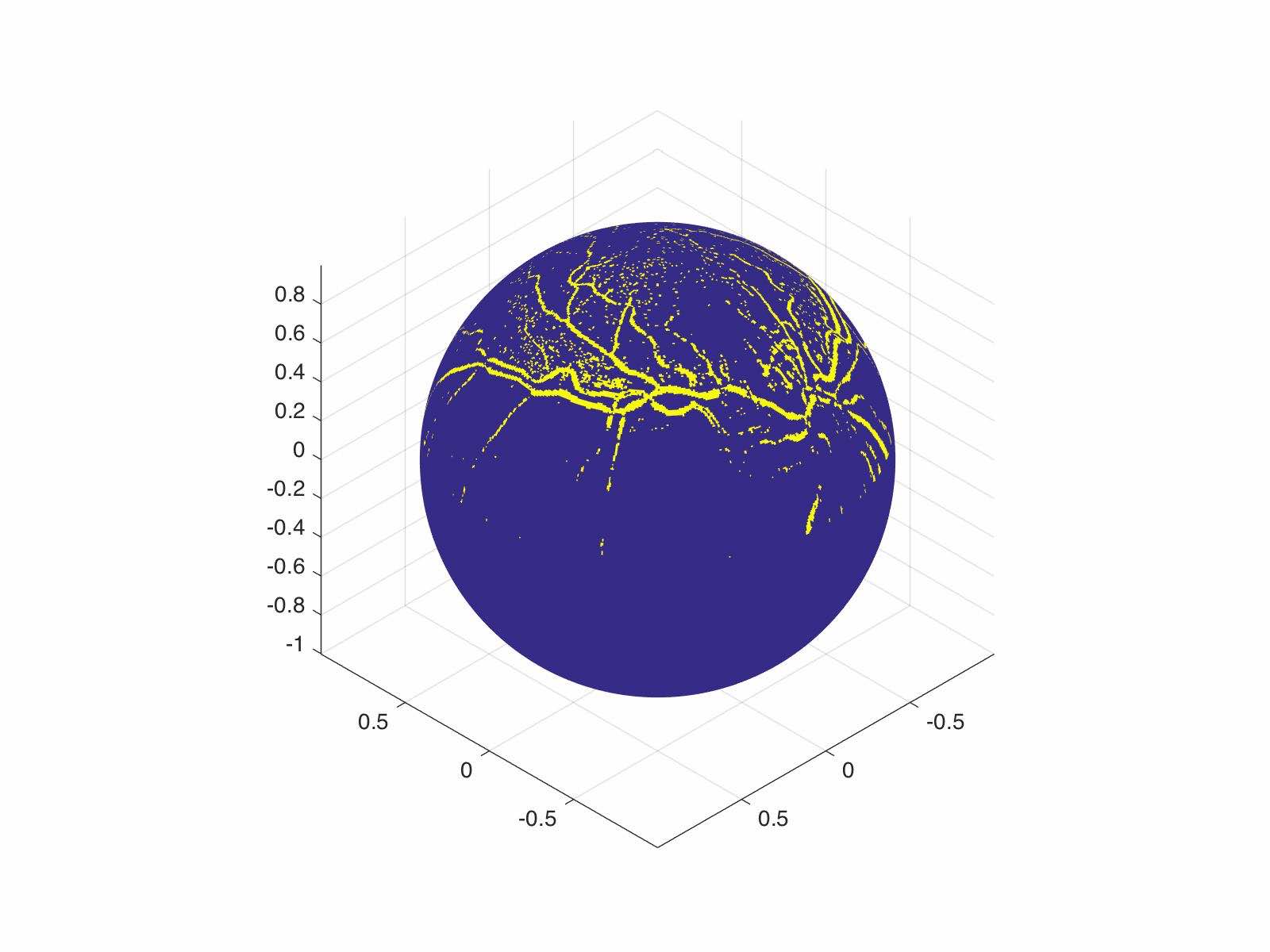} &
        		\includegraphics[trim={{.9\linewidth} {.32\linewidth} {.8\linewidth} {.6\linewidth}}, clip, width=0.2\linewidth]
		{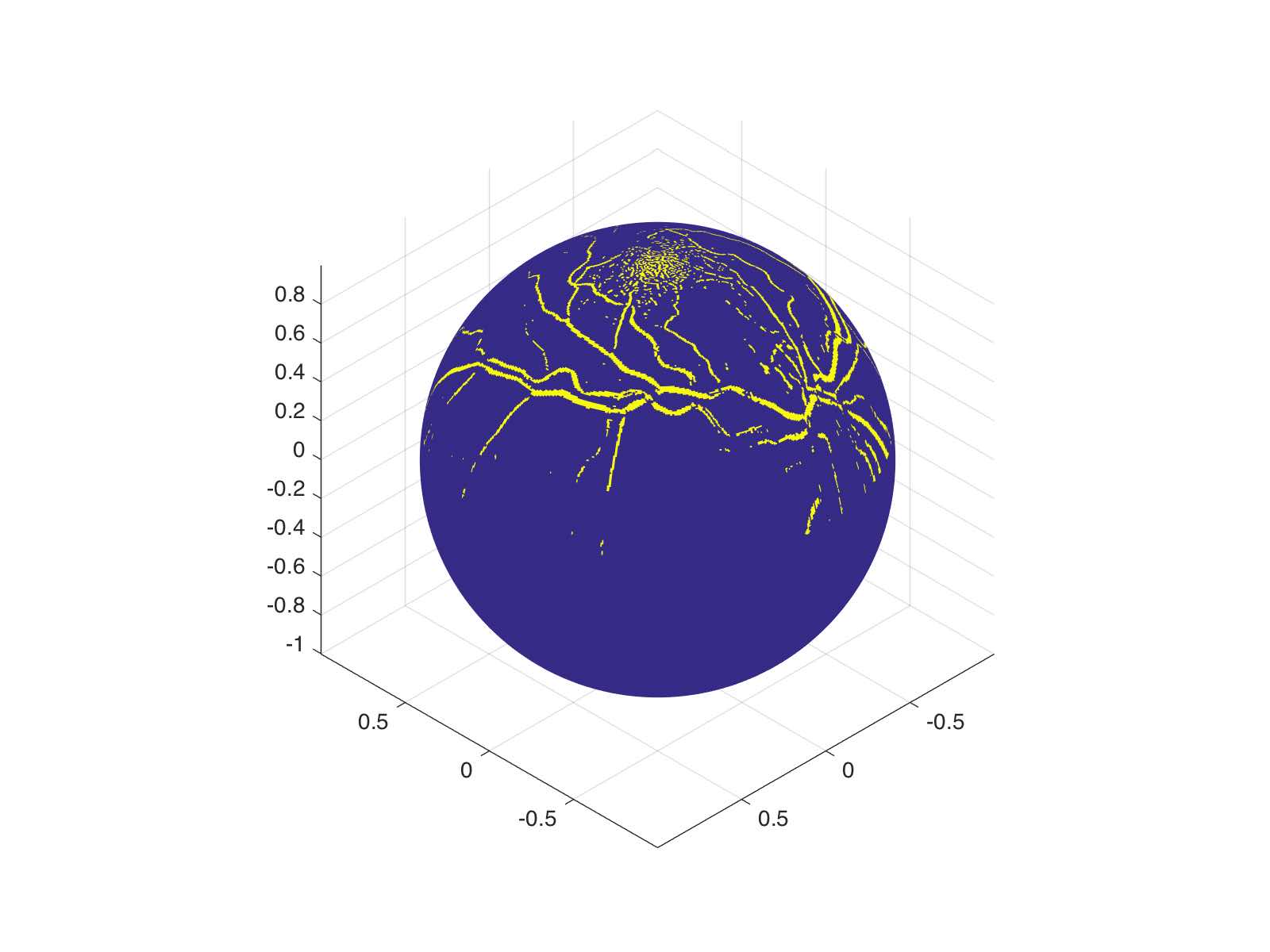} &
        		\includegraphics[trim={{.9\linewidth} {.32\linewidth} {.8\linewidth} {.6\linewidth}}, clip, width=0.2\linewidth]
		{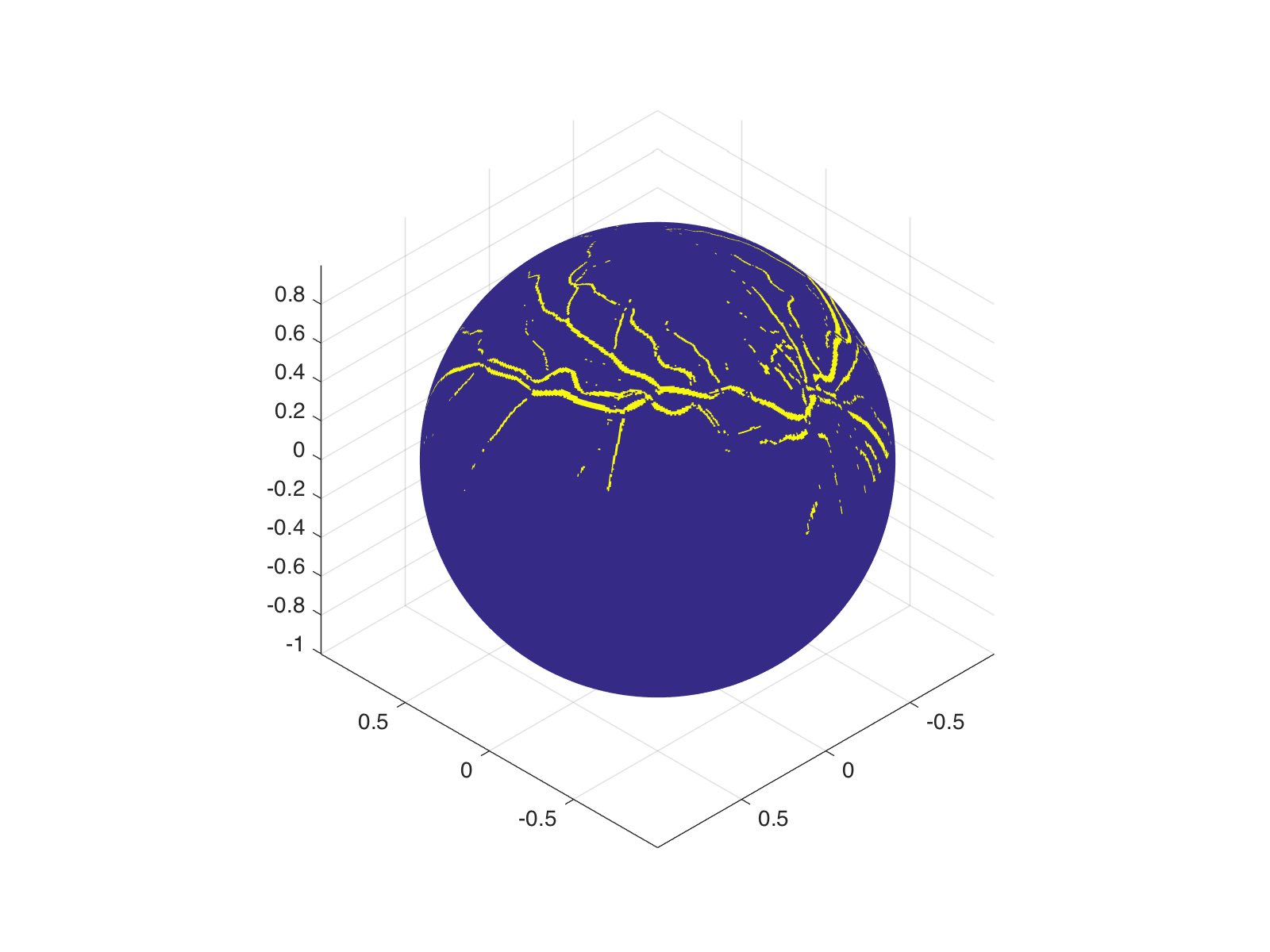} 
		\\
		\includegraphics[trim={{.4\linewidth} {.2\linewidth} {.35\linewidth} {.2\linewidth}}, clip, width=0.2\linewidth]
		{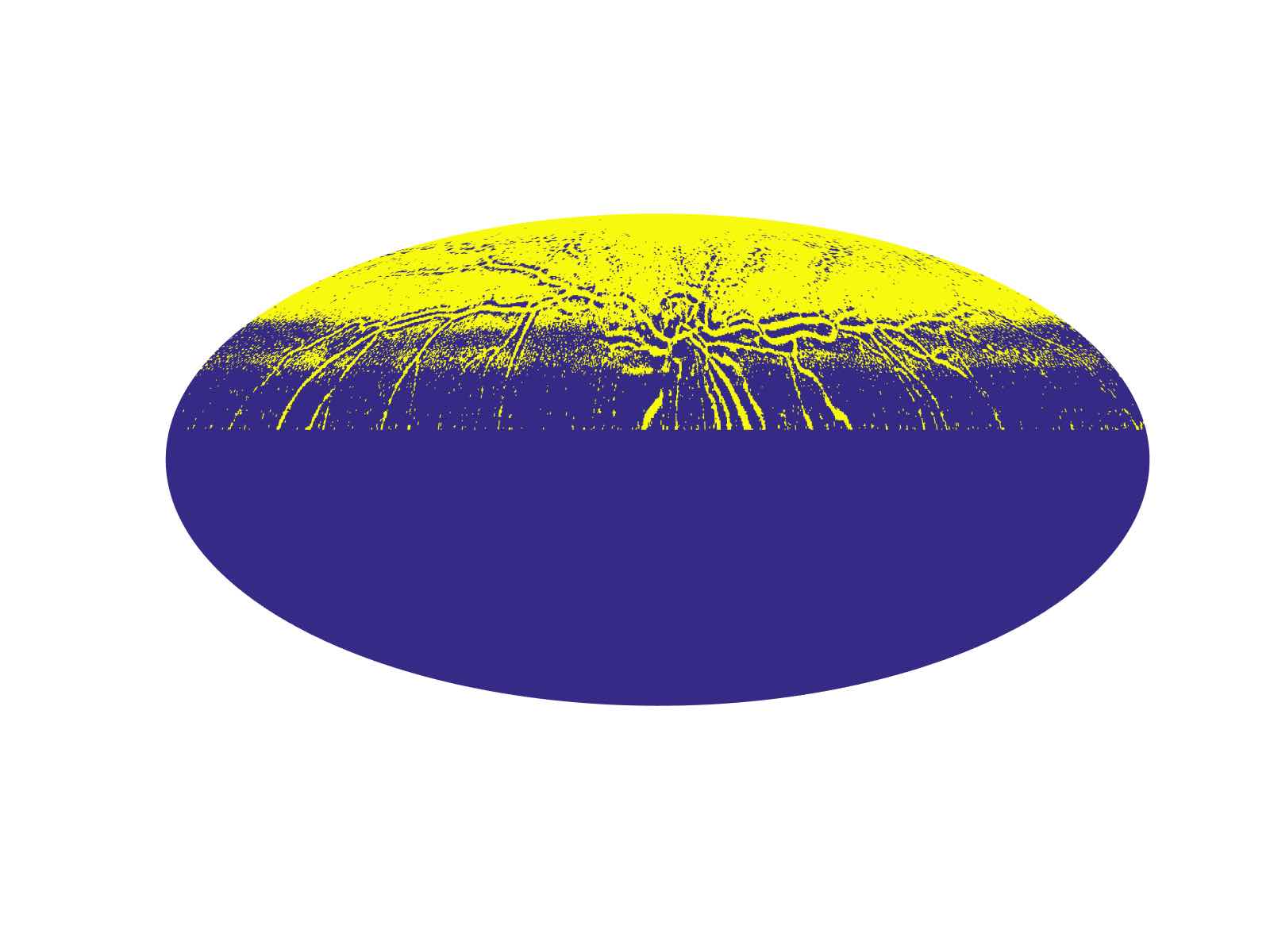}  \put(-56,38){\red{\framebox(10,8){ }}}  &
        		\includegraphics[trim={{.4\linewidth} {.2\linewidth} {.35\linewidth} {.2\linewidth}}, clip, width=0.2\linewidth]
		{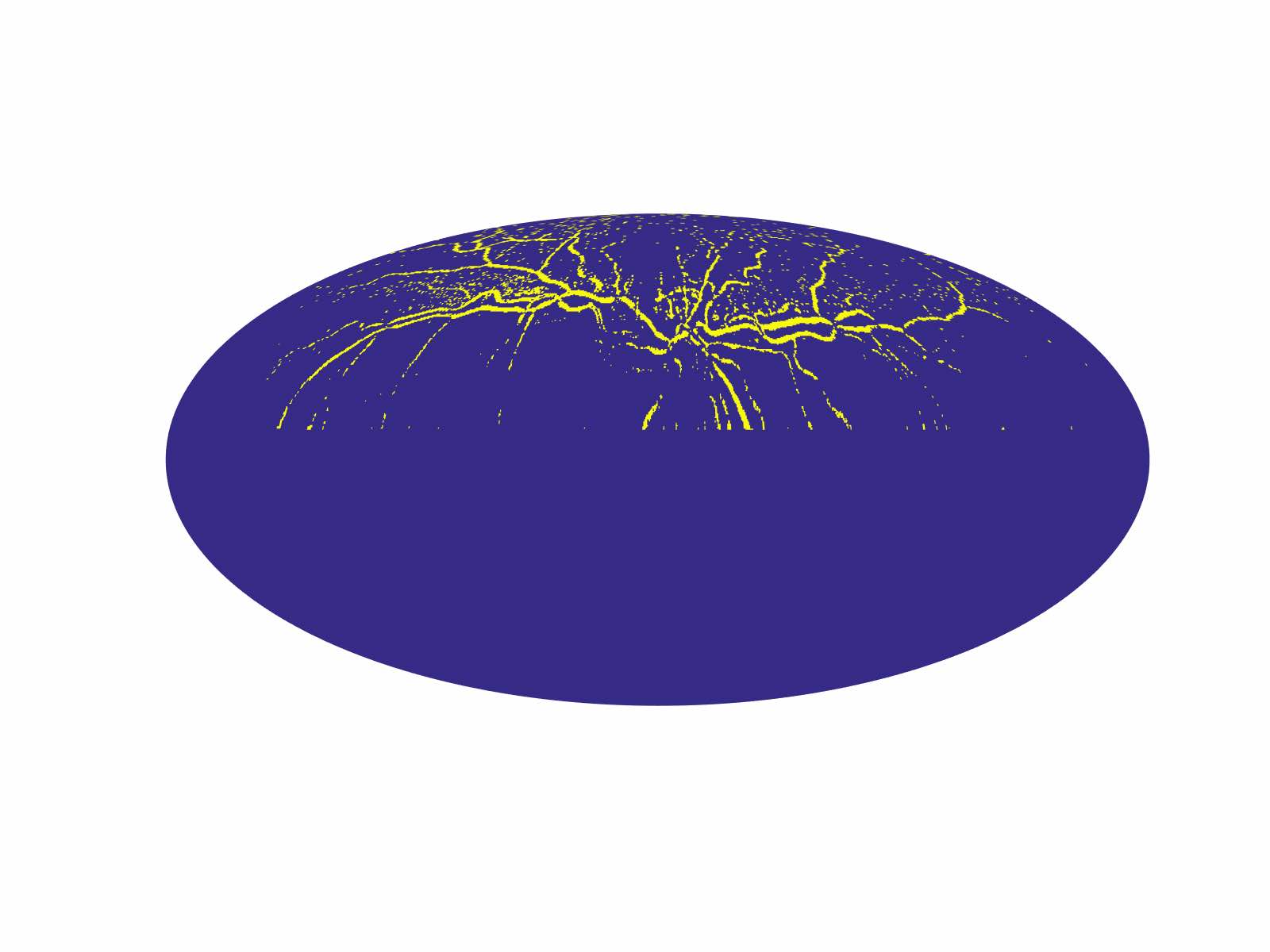}  \put(-56,38){\red{\framebox(10,8){ }}}  &
                 \includegraphics[trim={{.4\linewidth} {.2\linewidth} {.35\linewidth} {.2\linewidth}}, clip, width=0.2\linewidth]
                 {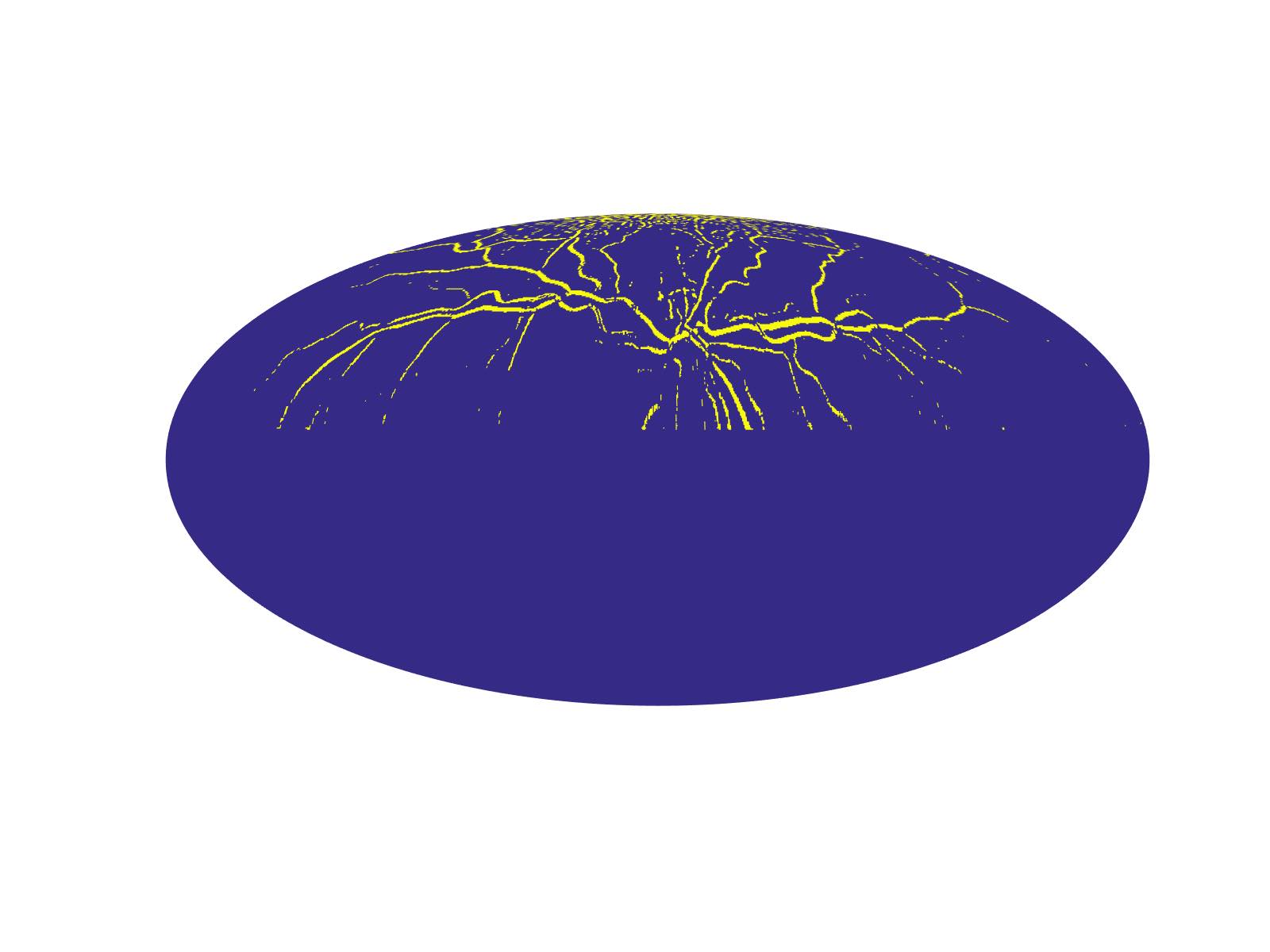}   \put(-56,38){\red{\framebox(10,8){ }}}  &  
                 \includegraphics[trim={{.4\linewidth} {.2\linewidth} {.35\linewidth} {.2\linewidth}}, clip, width=0.2\linewidth]
                 {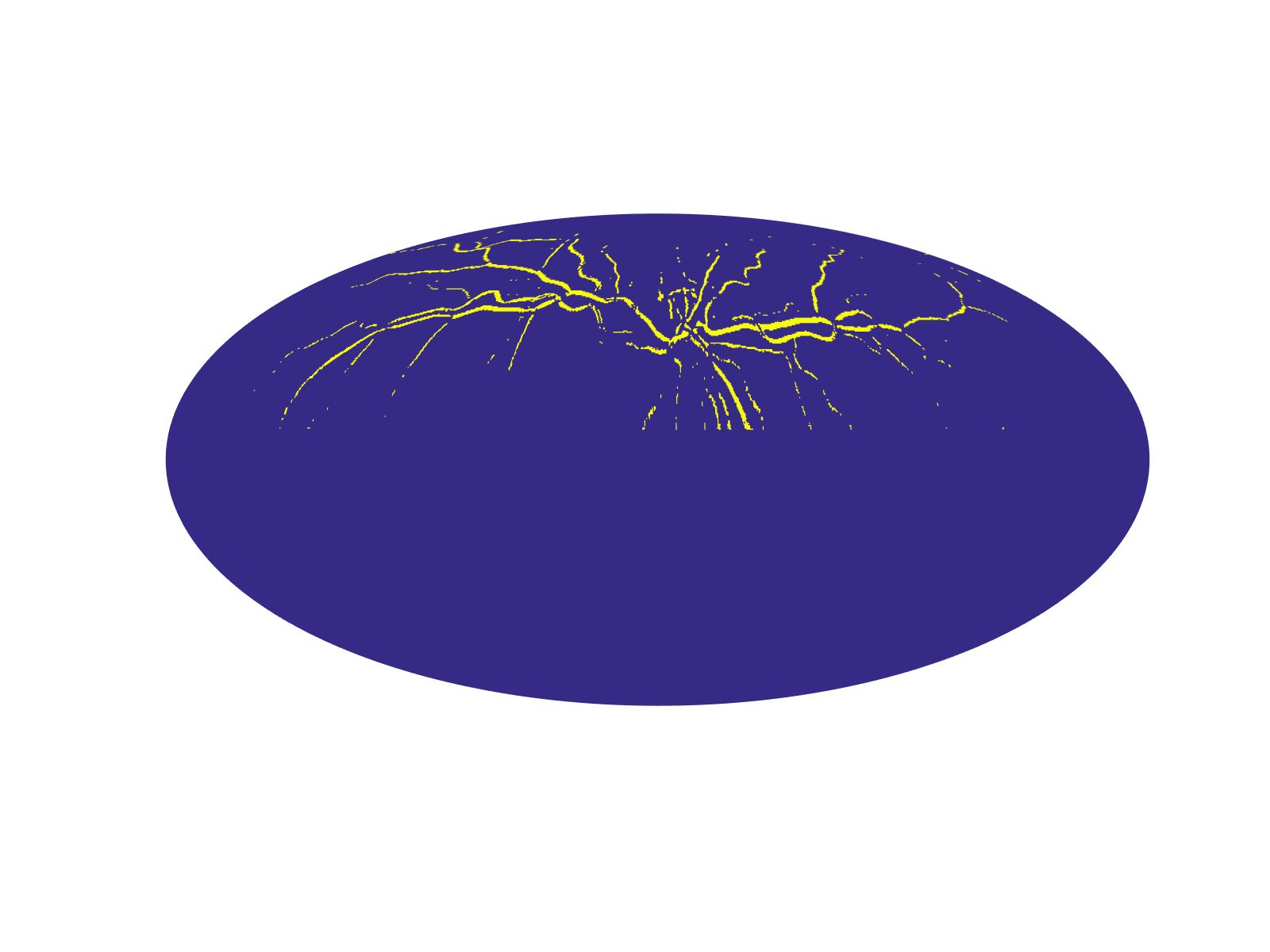}   \put(-56,38){\red{\framebox(10,8){ }}} 
                 \\
        		\includegraphics[trim={{1.4\linewidth} {2.0\linewidth} {2.5\linewidth} {0.90\linewidth}}, clip, width=0.2\linewidth]
		{./fig/retina01_AxisymWav_L_512_Resutls_Th.jpg}  &
        		\includegraphics[trim={{1.4\linewidth} {2.0\linewidth} {2.5\linewidth} {0.90\linewidth}}, clip, width=0.2\linewidth]
		{./fig/retina01_AxisymWav_L_512_Resutls.jpg}   &
        		\includegraphics[trim={{1.4\linewidth} {2.0\linewidth} {2.5\linewidth} {0.90\linewidth}}, clip, width=0.2\linewidth]
                 {./fig/retina01_DirectionWav_L_512_N_6_Resutls.jpg} &  
        		\includegraphics[trim={{1.4\linewidth} {2.0\linewidth} {2.5\linewidth} {0.90\linewidth}}, clip, width=0.2\linewidth]
                 {./fig/retina01_HybridWav_L_512_N_6_Resutls.jpg}  
                 \\
		{\small (e) K-means} & {\small (f) WSSA-A} & {\small (g) WSSA-D} & {\small (h) WSSA-H}
        \end{tabular}
        	\caption{Results of retina image.  
	First row: original retina image (a1), its green channel (b1), the image after removing its background (c1) and the noisy image (d1) respectively; 
	Second row: projected retina image shown on the sphere (a) and in 2D using a mollweide projection (b), and the zoomed-in red rectangle area 
	of the noisy (c) and original images (d), respectively; 
	Third row to fifth row: results of methods K-means (e), WSSA-A (f), WSSA-D (g) with $N=6$ (even $N$), 
	WSSA-H (h) with $L_{\rm trans}=32$ for curvelets, respectively.}
	\label{fig-retina01}
\end{figure}

In this example, we constructed test data containing very strong anisotropic structures from retina images. 
The retina images raised a very challenging segmentation problem because of the well known thin-vessel network
(for example see \cite{AKV07,JCLSCL07}). These kind of thin vessels are excellent for testing the ability of methods 
to tackle highly directional structures.
The retina images tested here, (a) of Figs. \ref{fig-retina02} and \ref{fig-retina01}, are from the DRIVE 
data-set\footnote{\url{http://www.isi.uu.nl/Research/Databases/DRIVE/}} obtained from a diabetic retinopathy screening program in The Netherlands. 
They were acquired using a Canon CR5 non-mydriatic 3CCD 
camera with a 45 degree field of view (FOV). Each image, captured using 8 bits per colour plane at $768 \times 584$ pixels, 
is scaled to \mbox{[0, 1]} in our case. The FOV of each image is circular with a diameter of approximately 540 pixels. 
The images have been cropped around the FOV, and a mask image is provided that delineates the FOV.

To recover spherical retina images, for our tests, from the colour images (e.g.\ Fig.\ \ref{fig-retina02} (a1)),
we transform the original planar images to the sphere in the following manner:
(1) extract the green channel from the colour image, i.e.\ extract Fig.\ \ref{fig-retina02} (b1) from (a1); 
(2) obtain the background of the green channel by implementing the {\sc Matlab} built-in function {\tt medfilt2} then remove it
from Fig.\ \ref{fig-retina02} (b1) to get figure (c1); (3) add Gaussian noise to (c1) to generate the noisy image (d1) in Fig.\ \ref{fig-retina02};
(4) project Fig.\ \ref{fig-retina02} (d1) to the spherical coordinate system to form the 
spherical retina image as our test data, shown in Fig.\ \ref{fig-retina02} (a) and (b).
Fig.\ \ref{fig-retina02} (c) and (d) are the zoomed-in details of the rectangle in figure (b1) and in the
projected image of figure (c1), respectively. 
Fig.\ \ref{fig-retina01} is generated and arranged using the same way as that in  Fig.\ \ref{fig-retina02}.

For the generated spherical retina images, (b) of Figs. \ref{fig-retina02} and \ref{fig-retina01}, we use 
$\epsilon= 0.04$ in our WSSA method. The third to the fifth rows of the figures show the segmentation results.
Clearly, the K-means method failed to identify most of the vessels (see the first column), while our WSSA method detected most of 
the vessels on the sphere (see the second until the fourth columns). After comparing the zoomed-in results, (f) -- (h) of 
Figs. \ref{fig-retina02} and \ref{fig-retina01}, we conclude that the WSSA-D and WSSA-H methods give better results than the
WSSA-A method, the results of which contain more non-vessel artefacts. 
From Fig.\ \ref{fig-retina02}, we see the improvement in the result of \mbox{WSSA-H} compared with the result of WSSA-D in terms of 
suppressing those non-vessel components about the north pole. Table \ref{tab:retina} presents the time performance of each method, 
which is consistent with the conclusions obtained in the previous examples, i.e., the more directional the wavelet transform, 
the longer the computation time required. In particular, to test the property of the WSSA-H method, we consider the hybrid wavelets with 
$L_{\rm trans}= 32$ and  $L_{\rm trans}=64$ in 
Fig.\ \ref{fig-retina01} and Fig.\ \ref{fig-retina02}, respectively, 
meaning that bands $\ell \lesssim 32$ and $\ell \lesssim 64$ are probed by curvelets,
respectively, and the other bands are probed by directional wavelets. Their performance are given in Table \ref{tab:retina}. 
We see that the higher the $L_{\rm trans}$, the longer computation time needed in the WSSA-H method, while the improvement in segmentation is small for these cases.

\begin{table}[h] 
\begin{center}
\caption{Retina image in Fig.\ \ref{fig-retina01}: Number of unclassified points at each iteration and computation time in seconds. 
	$^\ast$The fourth and fifth columns represent the results of WSSA-D with $N = 5$ and $6$, respectively.
	$^\dagger$The last two columns represent the results of WSSA-H with $L_{\rm trans}=32$ and 64 for curvelets, respectively.} \label{tab:retina}
 \vspace{-0.05in}
\begin{tabular}{|c||c|c|c|c|c|c|}
\hline 
  & {K-means} & {WSSA-A} & WSSA-D$^\ast$ & WSSA-D$^\ast$ & WSSA-H$^\dagger$ & WSSA-H$^\dagger$  
\\  \hline  \hline
\raisebox{-.15ex}{$|\bar{\mathbb{S}}^2|$} &  $523776$ & $523776$ & $523776$  & $523776$  & $523776$  & $523776$
\\ \hline
 $|\Lambda^{(0)}|$ & -  & 1640 & 1914   &  1934  &  1934 & 1928
\\ \hline
$|\Lambda^{(1)}|$ &   -  & 46611   &  28376  &  49958  &  12237 &  12088
\\ \hline
$|\Lambda^{(2)}|$ &  -  & 11366 &  6457  & 16346    &  3040  & 2945
\\ \hline
$|\Lambda^{(3)}|$ &  -  & 3095   &  1783  &   5001  &  852 &  832
\\ \hline
$|\Lambda^{(4)}|$ &  -  & 949   &  512  & 1603   &  257 &  258
\\ \hline
$|\Lambda^{(5)}|$ &  -  & 301    &  175 &  538  &  79 &  76
\\ \hline
$|\Lambda^{(6)}|$ &  -  & 95   & 63  & 183  & 19 &  25
\\ \hline
$|\Lambda^{(7)}|$ &  -   &  31  & 20  & 49  & 1 & 5
\\ \hline
$|\Lambda^{(8)}|$ &  -   &  10  & 6  & 18  & 0 & 0
\\ \hline
$|\Lambda^{(9)}|$ &  -   &  4  &  0 &  5 & - & -
\\ \hline
$|\Lambda^{(10)}|$ &  -   &  0  & -  &  0  & - & -
\\ \hline \hline
Time & $<$ 1 s   &  50.66 s  & 160.54 s & 197.0  s & 789.6  s & 4538.9 s
\\ \hline
\end{tabular}
\end{center}
\end{table}

\section{Conclusions}\label{sec:con}
In this paper we proposed a wavelet-based segmentation method (WSSA) for spherical images, which is, to the best of our knowledge, the first method 
performing segmentation directly on the sphere. The method is compatible with any invertible wavelet transform constructed on the sphere
(e.g.\ axisymmetric wavelets, directional wavelets, curvelets, or hybrid wavelets).
Consequently, WSSA is very flexible and can be equipped with spherical wavelets appropriate for the texture property of the given spherical data of interest.
WSSA needs just a few iterations to converge, and the main computation within each iteration is the pair of forward and backward 
wavelet transforms. We applied our WSSA method to several real-world problems, i.e., the Earth topographic map, a light probe image, two Solar maps, 
and two projected spherical retina images. 
The comparisons with the K-means method and different types of wavelets demonstrate that the WSSA method is an
efficient and effective spherical segmentation method and is superior to K-means. One important future work will be focusing on purifying the uncertainty area $\Lambda^{(i)}$ 
at each step in each iteration of WSSA to improve segmentation quality according to specific applications.
{Another important future work will be extending the proposed spherical segmentation method by exploiting the recent developments of spherical neural networks.}

\section*{Acknowledgements}
This work is supported by the UK Engineering and Physical Sciences Research Council (EPSRC) by grant EP/M011852/1 and EP/M011089/1.  We thank Professor Raymond H. F. Chan in City University, Hong Kong, for the very helpful discussion.
We also thank Dr David Peres-Suarez for his help in providing the first set of solar data.
{We would also like to thank the editors and anonymous reviewers for their
valuable comments and suggestions to improve the quality of the paper.}



\end{document}